\documentclass[runningheads]{llncs}
\usepackage{graphicx}

\usepackage{tikz}
\usepackage{comment}
\usepackage{amsmath,amssymb} 
\usepackage{color}
\usepackage{mathrsfs}
\usepackage[export]{adjustbox}
\usepackage{array}

\usepackage[accsupp]{axessibility}  
\usepackage{booktabs}
\usepackage{float}

\definecolor{teal}{rgb}{0,0.5,0.5}
\definecolor{darkgreen}{rgb}{0,0.5,0}
\definecolor{darkblue}{rgb}{0,0,0.7}
\definecolor{revmwcol}{rgb}{0.4,0.5,0.1}
\definecolor{darkred}{rgb}{0.75,0,0}
\newcommand{\todo}[1]{\textcolor{darkred}{TODO: #1}}

\newcommand{\hide}[1]{}

\newcolumntype{P}[1]{>{\centering\arraybackslash}p{#1}}

\usepackage{comment}
\usepackage[normalem]{ulem}

\usepackage{xspace}

\def\etal{et~al.\xspace}

\begin{document}
\pagestyle{headings}
\mainmatter
\def\ECCVSubNumber{}  

\title{Occlusion Fields: An Implicit Representation for Non-Line-of-Sight Surface Reconstruction} 


\titlerunning{Occlusion Fields}
%
\author{Javier Grau\inst{1} \and
Markus Plack\inst{1} \and
Patrick Haehn\inst{1} \and
Michael Weinmann\inst{2} \and
Matthias Hullin\inst{1}}

\authorrunning{Grau et al.}
%
\institute{University of Bonn, Germany \and
Delft University of Technology, Netherlands \\
\email{\{jgraucho, haehn\}@uni-bonn.com}\\
\email{\{mplack, hullin\}@cs.uni-bonn.com} \\
\email{m.weinmann@tudelft.nl}
}
\maketitle

\begin{abstract}	
{Non-line-of-sight reconstruction (NLoS) is a novel indirect imaging modality that aims to recover objects or scene parts outside the field of view from measurements of light that is indirectly scattered off a directly visible, diffuse wall. Despite recent advances in acquisition and reconstruction techniques, the well-posedness of the problem at large, and the recoverability of objects and their shapes in particular, remains an open question. The commonly employed Fermat path criterion is rather conservative with this regard, as it classifies some surfaces as unrecoverable, although they contribute to the signal.
	
In this paper, we use a simpler necessary criterion for an opaque surface patch to be recoverable. Such piece of surface must be directly visible from \emph{some} point on the wall, and it must occlude the space behind itself. Inspired by recent advances in neural implicit representations, we devise a new representation and reconstruction technique for NLoS scenes that unifies the treatment of recoverability with the reconstruction itself. Our approach, which we validate on various synthetic and experimental datasets, exhibits interesting properties. Unlike memory-inefficient volumetric representations, ours allows to infer adaptively tessellated surfaces from time-of-flight measurements of moderate resolution. It can further recover features beyond the Fermat path criterion, and it is robust to significant amounts of self-occlusion. 
We believe that this is the first time that these properties have been achieved in one system that, as an additional benefit, is trainable and hence suited for data-driven approaches.
}

\end{abstract}


\section{Introduction}

\begin{figure}[t]
	\centering
	\includegraphics[trim=11mm 8mm 5mm 3mm,clip,width = 0.37\columnwidth]{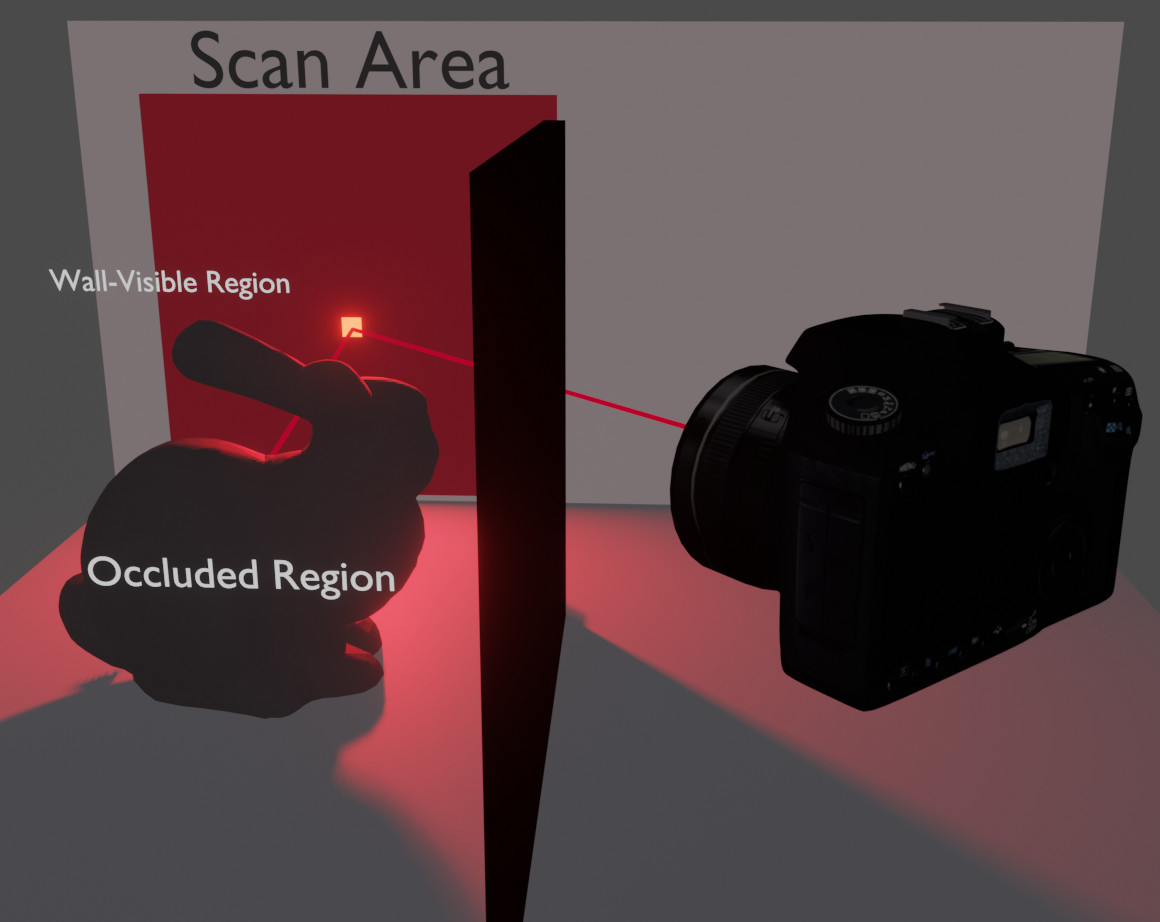}\hspace{5mm}
	\includegraphics[trim=1.5cm 15cm 1.5cm 1cm,clip,width = 0.51\columnwidth]{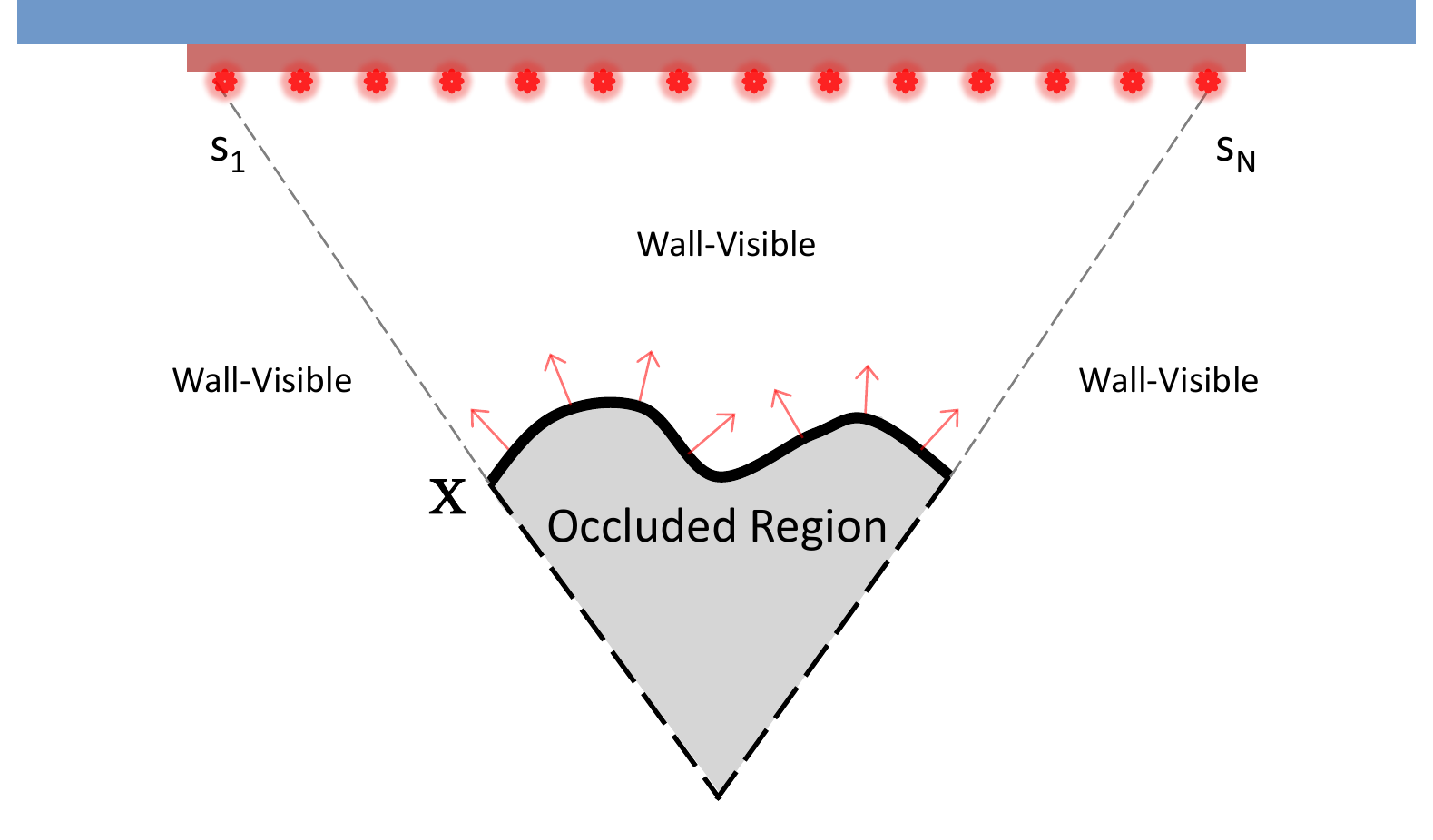}
	%
	\caption{\textbf{Left:} \textbf{Non-line-of-sight scene:} The hidden object is indirectly scanned by sending light pulses from the sensor to the visible parts of a diffuse wall. \textbf{Right:} \textbf{Occlusion Model:} Only the front-face of the hidden scene X is scanned by the virtual sensors $\{s_1, ..., s_N\}$. The NLoS surface can be recovered by sampling the interface between the wall-visible and occluded regions}
	\label{fig:nlosScene+mainIdea}
	\vspace{-4mm}
\end{figure}

The development of time-of-flight sensors has seeded numerous technologies during the last decades. One modality that has caught attention within recent years is non-line-of-sight imaging (NLoS), which seeks to retrieve scene information beyond the camera's field of view. By casting light pulses onto directly-visible diffusers and analyzing the returning reflections, this technique allows to \textit{see} 2D and 3D structure of otherwise unreachable regions of the scene (see Figure \ref{fig:nlosScene+mainIdea}). Endorsing systems with such capabilities could enable new and diverse applications in scale and scope, such as self-driving cars, exploration and rescue missions as well as medical endoscopy.

At the very core of this ill-posed task resides the challenge of inferring the geometric structure of the hidden scene. A large number of works have traditionally modeled the unknown scene as a three-dimensional tensor of voxel-shape activations, a trend likely rooted in its mathematical simplicity. Despite its convenience, the volumetric albedo representation poses obvious limitations for general NLoS applications: it imposes large demands regarding memory and lacks adaptivity for describing complex curvature changes. Moreover, in the case of some state-of-the-art methods \cite{otoole2018confocal,lindell2019wave,liu2019Phasor}, the dimensionality of the reconstructed volume grows further with the number of collected scans and the temporal resolution of the sensor. Generally, an ideal representation should be both computationally and memory efficient, while allowing for arbitrary resolution of the underlying geometry. This makes point clouds and tessellated meshes attractive choices for the task of NLoS reconstruction. However, to our knowledge only three works have attempted non-volumetric descriptions of the problem amid the vast literature. Tsai \etal\cite{tsai2019beyond} and Iseringhausen \etal\cite{iseringhausen2020non} successfully reconstructed tessellated surfaces from time-of-flight measurements, but the former requires strong prior knowledge of the unknown target and the latter is highly sensitive to noisy inputs. Xin \etal\cite{xin2019theory} efficiently retrieved NLoS targets as oriented point clouds, but this method produces sparse reconstructions of the scene and is not suitable for the reconstruction task based on low-resolution inputs. Surface-oriented and point-based formulations of the light transport operator remain still theoretically difficult.

Another issue that hinders NLoS reconstruction frameworks relates the recovery of certain geometric structures with hard physical constraints of the acquisition setup. Previous investigations~\cite{xin2019theory,liu2019analysis} provide theoretical and experimental evidence that although present in the measurement, surface elements pointing outside the relay's scan area are unlikely to be retrieved. This issue, which we refer to as the \textit{specular Fermat property}, imposes fundamental limits to what current NLoS systems can reconstruct, thus suggesting that future efforts may be as well oriented towards devising more optimal scanning procedures or intelligent priors that enable denser shape recovery.

In this paper, we introduce a novel representation for {non-line-of-sight} scenes, {\emph{Occlusion Fields}}, a closed volume of recoverable surfaces that enables both efficient and descriptive recovery from time-of-flight measurements of moderate resolution. Specifically, we model the NLoS surface as the decision boundary of a neural network that discriminates between points that are \emph{wall-visible} and those that are occluded behind the hidden target. Beyond its flexibility, our model is able to recover Fermat and non-Fermat geometry, offers robustness to significant amounts of self-occlusion and allows for end-to-end training on clean or noisy inputs.

In summary, the major contributions of our work are as follows:

\begin{itemize}
	\setlength\itemsep{0.5em}
	
	\item We introduce a novel implicit representation for non-line-of-sight scenes that can be learned by existing architectures for the task of efficient 3D mesh reconstruction from time-resolved measurements.
	
	\item We demonstrate the potential of our representation in the scope of several experiments where we show its capability to also reconstruct scene parts beyond Fermat path limits and under the presence of strong self-occlusion.
	
	\item We will provide the used datasets as well as our code for computing our representation and training procedure upon the acceptance of the submission.
	
\end{itemize}

\section{Related work}


%
%
\hide{
In contrast to conventional sensor systems that capture scene entities in the direct line of sight of the sensor, non-line-of-sight imaging is focused on the recovery of occluded objects from their indirect reflections or shadows on directly observable surfaces.
Thereby, NLoS imaging explicitly forms an essential prerequisite towards scene understanding beyond the visible scene parts.
}
In contrast to conventional sensor systems that capture scene entities in the direct line of sight of the sensor, non-line-of-sight imaging is focused on the recovery of occluded objects outside the visible scene parts from their indirect reflections or shadows on directly observable surfaces.
\hide{
Despite its theoretical introduction in 1990~\cite{freund1990looking} and respective theoretically focused follow-up works~\cite{kirmani2009looking}, a proof of concept of non-line-of sight imaging was only provided in 2012 by Velten et al.~\cite{velten2013femto} at the example of static objects.
}
While early work primarily focused on theoretical considerations~\cite{freund1990looking,kirmani2009looking}, a proof of concept of non-line-of sight imaging was first provided by Velten et al.~\cite{velten2013femto} at the example of static objects.
Major challenges of NLoS arising from the underlying measurement principle include the lacking angular resolution resulting from the consideration of diffuse indirect reflections at diffuse surfaces within the visible domain and the low intensity signal received due to the quadratic decay of the intensity of the indirectly reflected light with respect to the distance along their path to the observed relay surface.
\hide{Increasing the angular resolution in terms of \todo{REFORMULATE AS DIRECTLY TAKEN FROM ELSEWHERE: temporally probing the light-transport in the scene, thereby unmixing light path contributions by their optical path length}~\cite{abramson1978light,kirmani2009looking,naik2011single,pandharkar2011estimating} comes at the cost of the requirement of a high effective temporal resolution of the captured images of light transport in the order of picoseconds.}
To increase the angular resolution, several approaches ~\cite{abramson1978light,kirmani2009looking,naik2011single,pandharkar2011estimating} temporally probe the light-transport in the scene to separate the light path contributions by their optical path length, which relies on a high effective temporal resolution of the captured images of light transport in the order of picoseconds.
A particular progress has been achieved based on directly measuring the temporal echoes of laser pulses sent into the scene~\cite{velten2012recovering,pandharkar2011estimating,gupta2012reconstruction,buttafava2015non,tsai2017geometry,arellano2017fast,otoole2018confocal}.
This approach, however, does not address the low signal of the indirect illumination and the respectively resulting noise sensitivity.
Solutions based on the assumption of retro-reflective materials in the scene~\cite{otoole2018confocal,chen2019steady,lindell2019wave} typically only apply to few surfaces within common scenes and alternative adjustments of the illumination power~\cite{lindell2019wave} require a level significantly beyond safe operation without eye-safety considerations.

Hence, recent techniques focused on incorporating accurate forward models to allow the generalization to different types of surface reflectance behavior, either without leveraging scene priors~\cite{liu2019Phasor,lindell2019acoustic}, based on simple priors such as non-negativity or sparsity priors ~\cite{otoole2018confocal} (whose use within an iterative optimization resulted in recovery times beyond $1{.}5$ hours), or based on explicit scene priors in terms of a plane-based model as surface representations~\cite{pediredla2017reconstructing,tsai2019beyond}. Additionally, Heide et al.~\cite{heide2019non} aimed at higher reconstruction accuracy by introducing a factored NLOS light transport representation, which accounts for partial occlusions and surface normals and allows for a factorization approach for inverse time-resolved light transport.
Tsai \etal~\cite{tsai2017geometry} leveraged the properties of first-returning photons from three-bounce light paths (i.e. direct bounce NLoS measurements) and their respective time of flight to guide a space carving. However, such first-returning photons cannot be received for all surface points of the NLoS object as this approach has problems in handling smooth convex objects with Lambertian reflectance, which creates light paths belonging to the tail of the light transient, as well as smooth non-convex objects with specular reflectance due to interreflections on the NLoS object that complicates the separation from the three-bounce light path from higher-bounce light paths. Furthermore, in case of non-smooth object surfaces, specular reflectance makes certain lighting/sensing pairs not receive any photon.
Furthermore, Liu \etal\cite{liu2019analysis} investigated a generic description for three-bounce light path NLoS measurements and analyzed
the captured NLoS information in the spatial frequency domain to get insights regarding visible and invisible features in the measurement space. 
Instead, learning richer scene priors in a supervised manner relies on large datasets as well as trainable NLoS methods for reconstruction.
These requirements are difficult to reach and limit respectively existing approaches~\cite{caramazza2018neural} to individual classes in a controlled setting, particularly due to the strongly limited datasets that only contain limitied object categories.
\hide{
Furthermore, Chen et al.~\cite{chen2020learned} showed that previous vanilla image-to-image mapping networks were not suitable for non-local NLOS reconstruction problems and instead proposed learning richer scene priors in terms of jointly learning a differentiable hidden scene representation with the reconstruction task based on a differentiable transient renderer directly from the raw transient images.
This provides robustness to surface reflectance as well as occlusion as it allows propagation from 2D to 3D based on physical models and, hence, incorporating multi-view and depth consistency, as well as backprojection to 2D based on a rendering network. \todo{CHECK: in real time rates?}
Further data-driven pipelines for NLoS reconstruction include the use of encoder-decoder network trained in an end-to-end fashion using an efficient transient renderer to map transient images directly to a depth map representation~\cite{grau2020deep} as well as, instead of temporal probing, applying steady-state NLOS imaging based on conventional intensity sensors and continuous illumination~\cite{chen_2019_nlos}, where an architecture has been trained to map diffuse indirect reflections to scene reflectance from purely synthetic data.
}
Furthermore, Chen et al.~\cite{chen2020learned} showed that previous vanilla image-to-image mapping networks were not suitable for non-local NLOS reconstruction problems. Instead, the authors proposed to learn richer scene priors in terms of jointly learning a differentiable hidden scene representation with the reconstruction task based on a differentiable transient renderer directly from the raw transient images to improve reconstruction quality. Similarly, Grau Chopite \etal~\cite{grau2020deep} used the combination of an end-to-end trained encoder-decoder network with an efficient transient renderer to map transient images directly to a depth map representation, and Chen \etal~\cite{chen2019steady} trained an architecture to map diffuse indirect reflections to scene reflectance from purely synthetic data based on applying steady-state NLOS imaging with conventional intensity sensors and continuous illumination instead of temporal probing.
Young et al.~\cite{young2020non} introduced a model to compute a joint albedo-normal representation.
Even 3D human pose estimation from transient images has been tackled leately based on a learnable inverse point spread function (PSF) for converting raw transient image data into deep feature vectors that are used together with a neural humanoid control policy derived from observed interactions with a physics simulator~\cite{isogawa2020optical}.
In addition, new acquisition techniques have facilitated in-the-wild applications~\cite{scheiner2020seeing,bijelic2020seeing}.
%

Whereas the community has focused mostly on retrieving 2D targets or volumetric albedo instances of the hidden scene, NLoS surface reconstruction still seems less explored due to the challenge of decoding surfaces from NLoS ToF captures{}.
%
However, the reconstruction of surfaces in the NLoS setting is a relevant problem as surface elements (orientation and local vicinity) play an essential role in light propagation, and tessellated representations provide more detailed information regarding the surface.
The few works along this direction include the use of inverse rendering, where a template mesh is iteratively transformed to match the query measurement~\cite{iseringhausen2020non,tsai2019beyond}.
Recent advances in learning implicit scene representations also have a high potential for facilitating surface representations in the NLoS context and, hence, could be the key to unlock powerful NLoS surface reconstruction techniques.
Respective approaches~\cite{2019occupancy,park2019deepsdf,xu2019disn,saito2019pifu} circumvent direct shape decoding by predicting implicit indications of the 3D space that the shape occupies.
However, these methods suffer from poor generalization to unseen classes and detail loss.
A growing number of works have been introduced to address these issues \cite{peng2020convolutional,deng2020cvxnet,genova2020local,chabra2020deep,sitzmann2020implicit}.
Recently, Shen \etal~\cite{shen2021non} explored the success of Neural Radiance Fields (NeRF)~\cite{mildenhall2020nerf} in the context of non-line-of-sight (NLOS) imaging, where an implicit representation in terms of a multi-layer perceptron (MLP) has been used to represent the neural transient field and the consideration of radiance along rays has been replaced by measuring the transient over spherical wavefronts.
In contrast, we introduce a novel non-line-of-sight scene representation, \textit{Occlusion Fields}, that unifies the treatment of recoverability with the reconstruction itself and allows modeling the NLoS surface as the decision boundary of a neural network that discriminates points that are wall-visible
from those that are occluded behind the hidden target.
This allows inferring adaptively tessellated surfaces from time-of-flight measurements of moderate resolution, thereby overcoming memory-inefficient volumetric representations, as well as recovering features beyond the Fermat path criterion, while being robust to self-occlusion and allowing for end-to-end training.

\section{Main Idea}
We focus on the problem of reconstructing the $3D$ surface of non-line-of-sight targets (see Figure ~\ref{fig:nlosScene+mainIdea}). In order to capture signals from hidden object $X$, an observer equipped with a time-resolved sensor and a laser casts short light pulses onto the visible wall and records the secondary reflections. This process can be simply written as:

\begin{equation}
m = T (X)
\end{equation}

where $m$ represents the measured \textit{transient image} and $T$ is the light transport operator. Therefore, reconstructing the target $X$ from $m$ involves solving the inverse problem. Note here that only the front-face of the hidden target casts light onto the wall, and thus the shape $X$ contained in $m$ is an open surface.

One way to alleviate this difficulty consists in considering descriptions of the scene that jointly depend on the NLoS target and the scan aperture. In particular, we observe that the NLoS surface $X$ belongs to the interface boundary between the wall-visible space and the target-occluded region behind $X$ (see Figure \ref{fig:nlosScene+mainIdea}). Note that the occluded region is a closed surface delimited not only by the NLoS target but also the lateral interface between visible area and the projected shadow hull. In the next section, we depart from this observation and show how sampling the occlusion function allows to extract NLoS surfaces from the transient measurement.

\vspace{-4mm}
\section{Mathematical Formulation}
We now proceed to present the image formation model of transient measurements and introduce our representation in this context.

\subsection{Image Formation Model}
Throughout the remaining of this manuscript, we consider the confocal setting for acquiring transient images from the relay wall \cite{otoole2018confocal}. Given the NLoS surface $X$ with albedo $\rho$, the total light contribution at $\vec{s_i}$ and time $\tau$ from all points $\vec{p}=(x,y,z)$ in the hemisphere $\Omega$ can be computed as:

\begin{eqnarray}\label{eq:transient_equation}
m(\vec{s_i}, \tau) = \int\int\int_{\Omega} \rho(x,y,z) f(s_{i_x}, s_{i_y}, x, y, z) \nonumber \\ 
\times \ \delta(2\sqrt{(s_{i_x} - x)^2 + (s_{i_y} - y)^2 + z^2} - \tau c) 
\ dx dy dz
\end{eqnarray}

where:

\vspace{-4mm}
\begin{eqnarray}\label{eq:geometric_factor}
f(s_{i_x}, s_{i_y}, \vec{p}) = v(\vec{s_i}, \vec{p}) \ \frac{\cos^j(\vec{n}, \vec{w})}{\lvert \vec{s_i} - \vec{p} \rvert^l}
\end{eqnarray}

is the \textit{geometric factor} containing the local visibility $v$ between $\vec{p}$ and $\vec{s_i}$, the normal $\vec{n}$ at $\vec{p}$ and $\vec{w}$ is the direction joining $\vec{s_i}$ and $\vec{p}$. Here, the exponents $(j,l)$ prescribe either diffuse surface characteristics $(2,4)$ or retroreflective surface characteristics $(4,2)$.

\subsection{NLoS Occlusion Model}\label{section:occlusion_model}
We define the \textit{occlusion} function as the \textit{orthogonal complement} of the visibility $v$ in equation \ref{eq:transient_equation}. That is, given a sensing point $s_i$ and surface $X$, we define the \textit{local occlusion} of a point $\vec{p}$ as:

\[\textit{o}_{s_i}(X, \vec{p}) = \overline{v(\vec{s_i}, \vec{p})} =\begin{cases} \label{eq:local_occlusion}
1, & \mbox{ if $\vec{p}$ occluded by X w.r.t.  $s_i$} \\
0,& \mbox{ if $\vec{p}$ not occluded by X w.r.t.  $s_i$}
\end{cases} \]

Then, we define the \textit{global occlusion} $\mathscr{O}$ of $\vec{p}$ given $X$ as the product operator among all sensing points $\{s_i\}_{i=1}^{N}$:

\begin{eqnarray}\label{eq:occlusion_def}
\mathscr{O}(X, \vec{p}) &=& o_{s_{1}}(X, \vec{p}) \times o_{s_{2}}(X, \vec{p})  ... \times  o_{s_{N}}(X, \vec{p}) \nonumber  \\ &=& \prod_i^{N} o_{s_{i}}(X, \vec{p}) \in \{0,1\}
\end{eqnarray}

Equation \ref{eq:occlusion_def} implies that a point $\vec{p}$ is occluded if it is so for all sensing positions. Conversely, if a point is visible it is so for at least \textit{one} sensing position $s_i$. This product can be modified to match different recovery criteria, e.g a point is visible if it is so for at least $k$ virtual sensors.

In order to compute $\mathscr{O}(X, \vec{p}$) according to Equation (\ref{eq:occlusion_def}) one must solve the inverse problem $X=T^{-1}(m)$. Nevertheless, $X$ and $m$ represent the target up to a linear transformation. In the remainder, we refer to $X$ and $m$ indistinguishably as the shape measurement.

\subsection{NLoS Occlusion Networks}
For the purpose of this work, we parametrize the occlusion function as a binary classifier $G$ that distinguishes wall-visible from occluded points given the surface $X$ (see Figure \ref{fig:pipeline}). Functionally, this can be written as:

\begin{equation}\label{eq:functional_inverse_model}
\mathscr{O}(m, \vec{p}) \rightarrow G_{\alpha}(F_{\beta}(m), H_{\gamma}(\vec{p})) = G_{\alpha}(i, q)
\end{equation}

where $F=T^{-1}$ represents an approximate inverse, $i$ is shape code learned from $m$ and $q$ corresponds to a positional encoding of the point $\vec{p}$ . $\alpha$ and $\beta$ are learnable parameters, and $\gamma$ might be alternatively trainable or not. In order to train our models, we feed our networks with transient images and occlusion samples of NLoS scenes. Similar to \cite{2019occupancy,peng2020convolutional}, we realize the occlusion function as a probability distribution $\mathscr{O}(m, \vec{p}) \rightarrow [0,1]$ over $3D$ space and minimize the binary-cross-entropy error over samples of points $K$:

\begin{figure*}\label{fig:pipeline}
	\centering
	\adjincludegraphics[width=0.95\columnwidth,trim={0 6cm 0 3.5cm},clip]{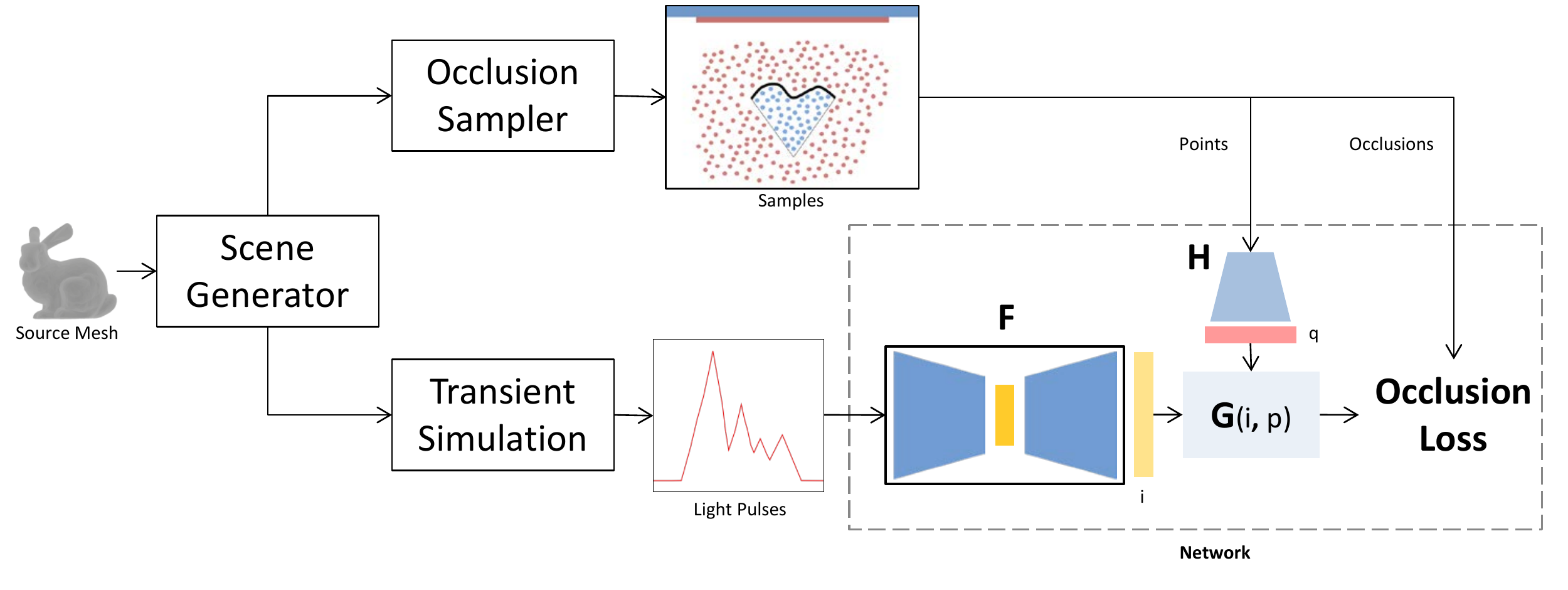}
	\centering
	\caption{\textbf{Reconstruction pipeline:} Instance pairs of transients and occlusions are generated by randomizing a pool of source meshes in the NLoS scene. Our network takes as inputs the time-of-flight measurement and points batches sampled in the hidden region. Each point is classified as either occluded or visible.}
\end{figure*}

\begin{equation}\label{eq:loss}
L_{occl}(\hat{ \theta}) = \frac{1}{\left|B\right|}\sum_{m=1}^{B} \sum_{n=1}^{K} BCE(G_{\hat{\theta}} (i_m, \vec{p}_{mn}), \ \mathscr{O}^{gt}_{mn})
\end{equation}

\noindent where $\hat{\theta}$ denotes the sets of learnable parameters $\alpha, \beta, \gamma$. 

The choices of $L_{occl}$, $G$, $F$ and $H$ may specify different NLoS-occlusion models. For example, equations \ref{eq:occlusion_def} and \ref{eq:loss} could also be defined as a SDF field with a regression loss. We found the latter to be beyond the scope of this work as it involves explicit calculation of the shadow hull behind the target.

\subsection{Representation Features}\label{section:features}
\begin{figure}[t]
	\centering
	\rotatebox{90}{\small ~Raw Prediction\textcolor{white}p} 
	\includegraphics[trim=11mm 8mm 5mm 3mm,clip,width = 0.20\columnwidth]{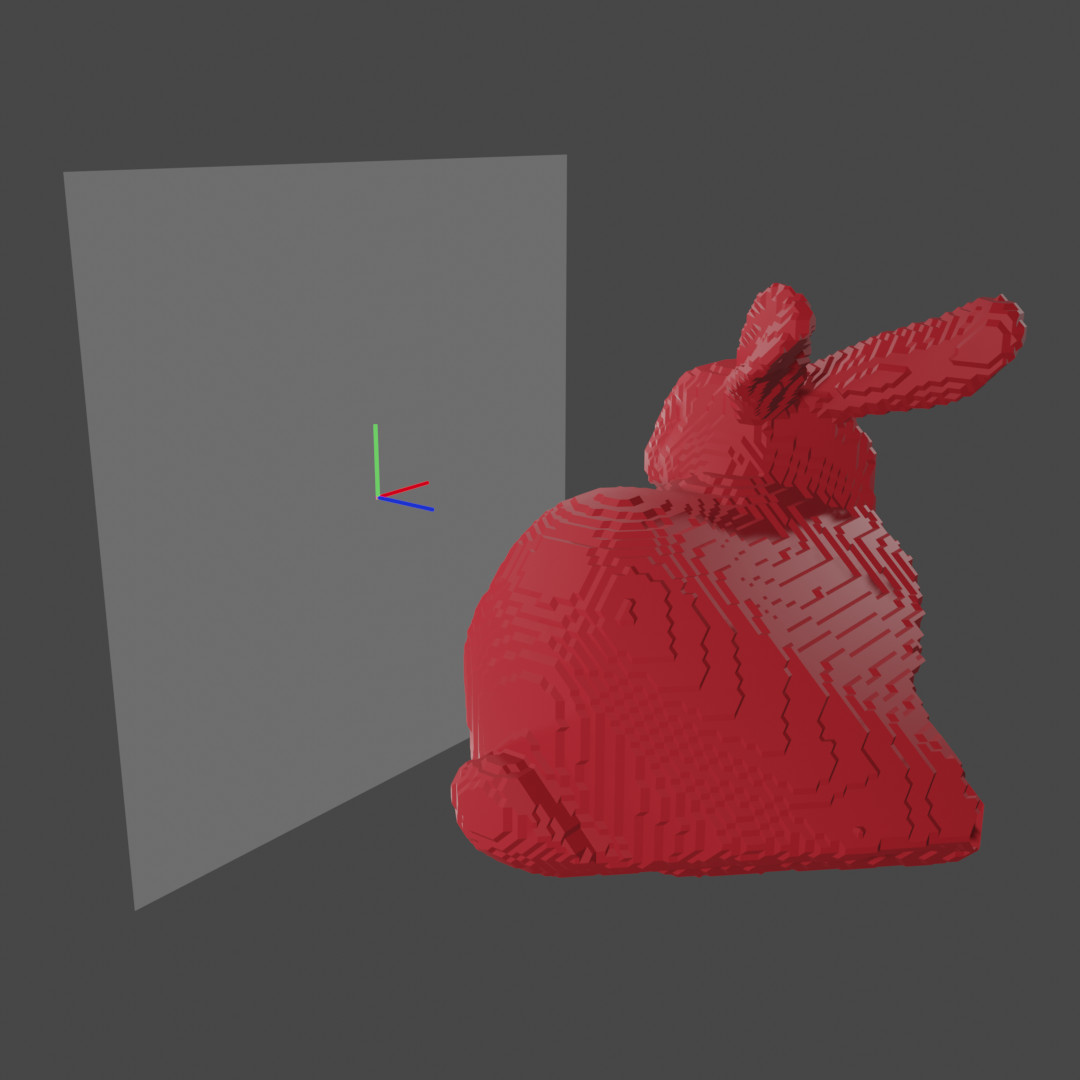}\vspace{0.25mm}
	\includegraphics[trim=11mm 8mm 5mm 3mm,clip,width = 0.20\columnwidth]{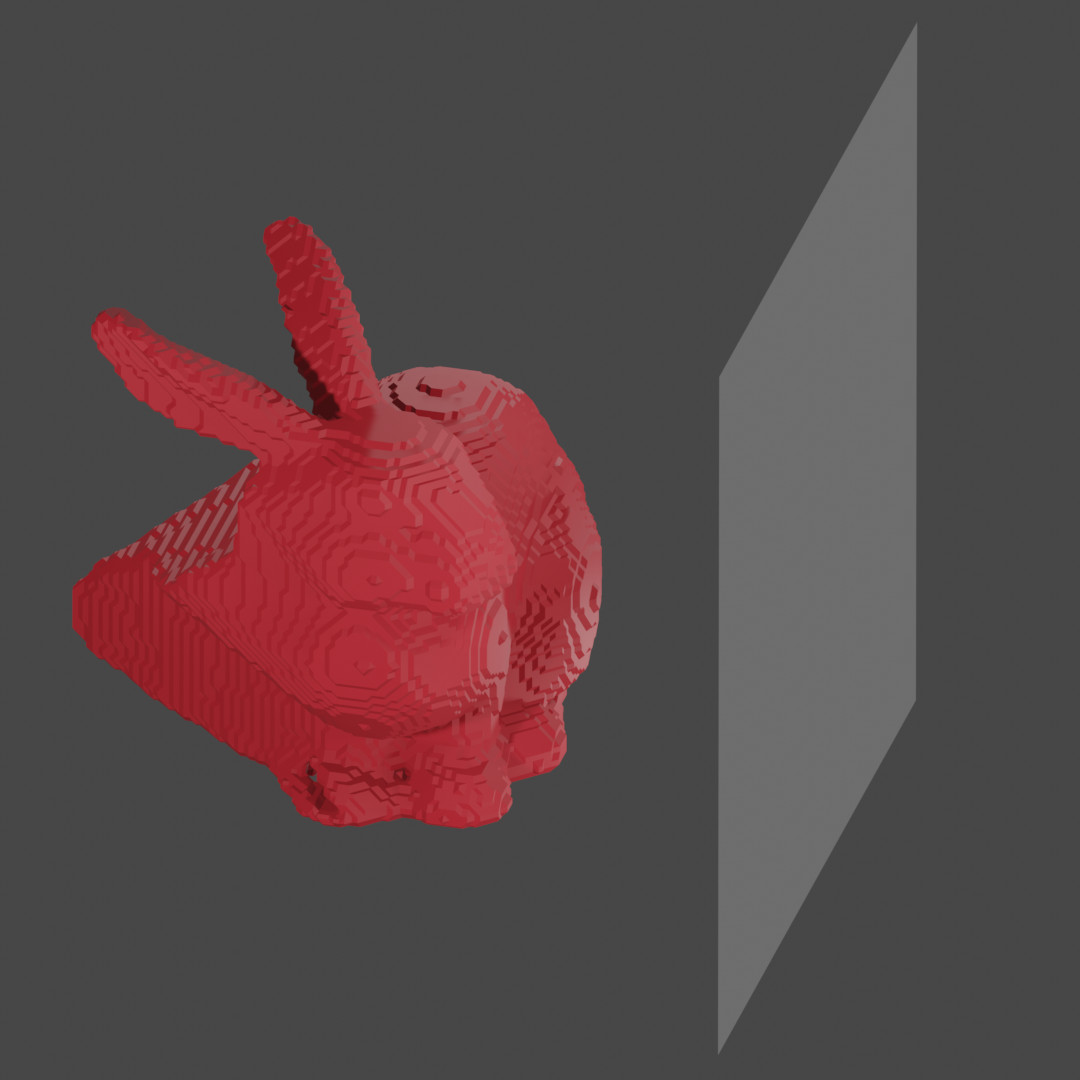}\hspace{5mm}
	\rotatebox{90}{\small~~NLoS Surface\textcolor{white}p} 
	\includegraphics[trim=11mm 8mm 5mm 3mm,clip,width = 0.20\columnwidth]{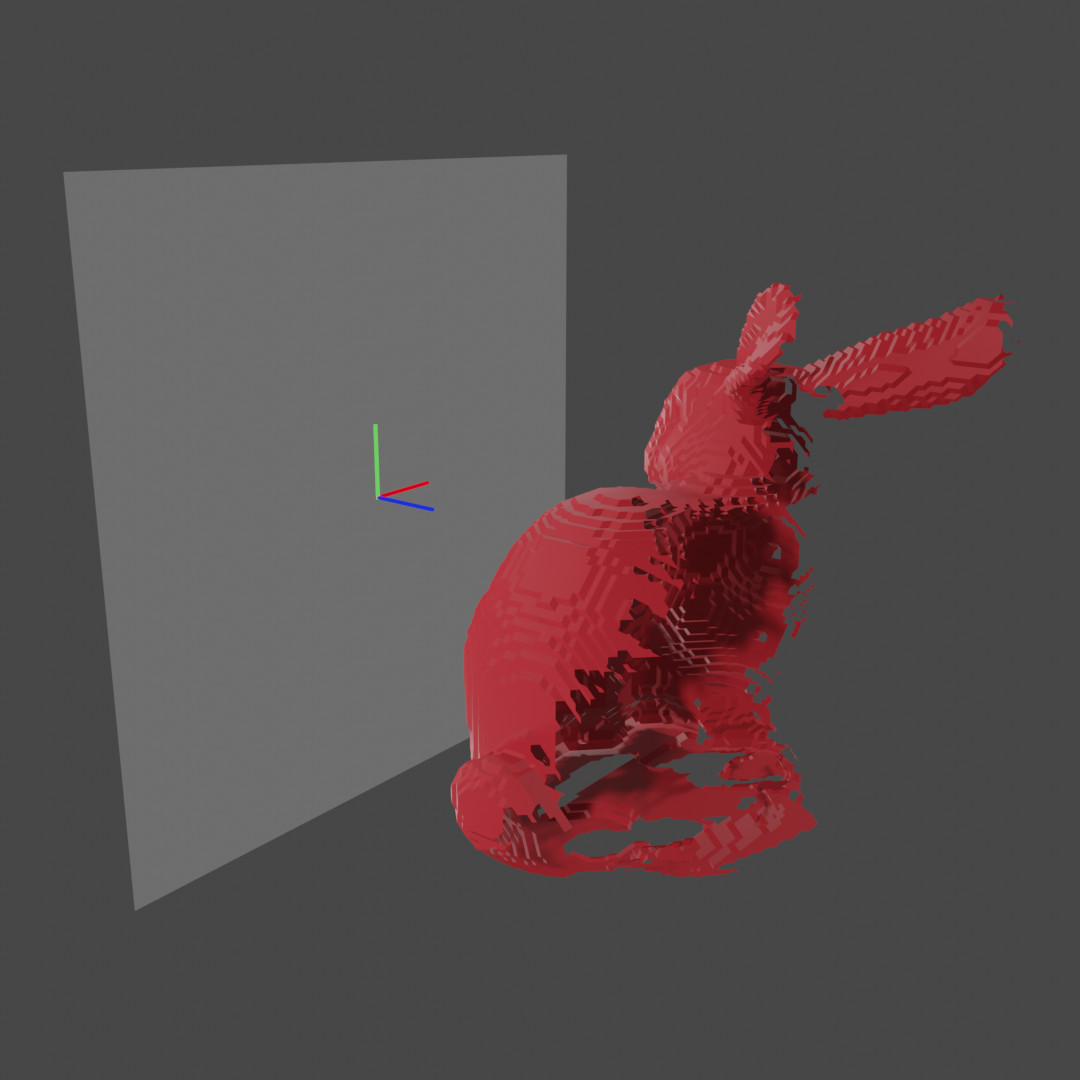}\vspace{0.25mm}
	\includegraphics[trim=11mm 8mm 5mm 3mm,clip,width = 0.20\columnwidth]{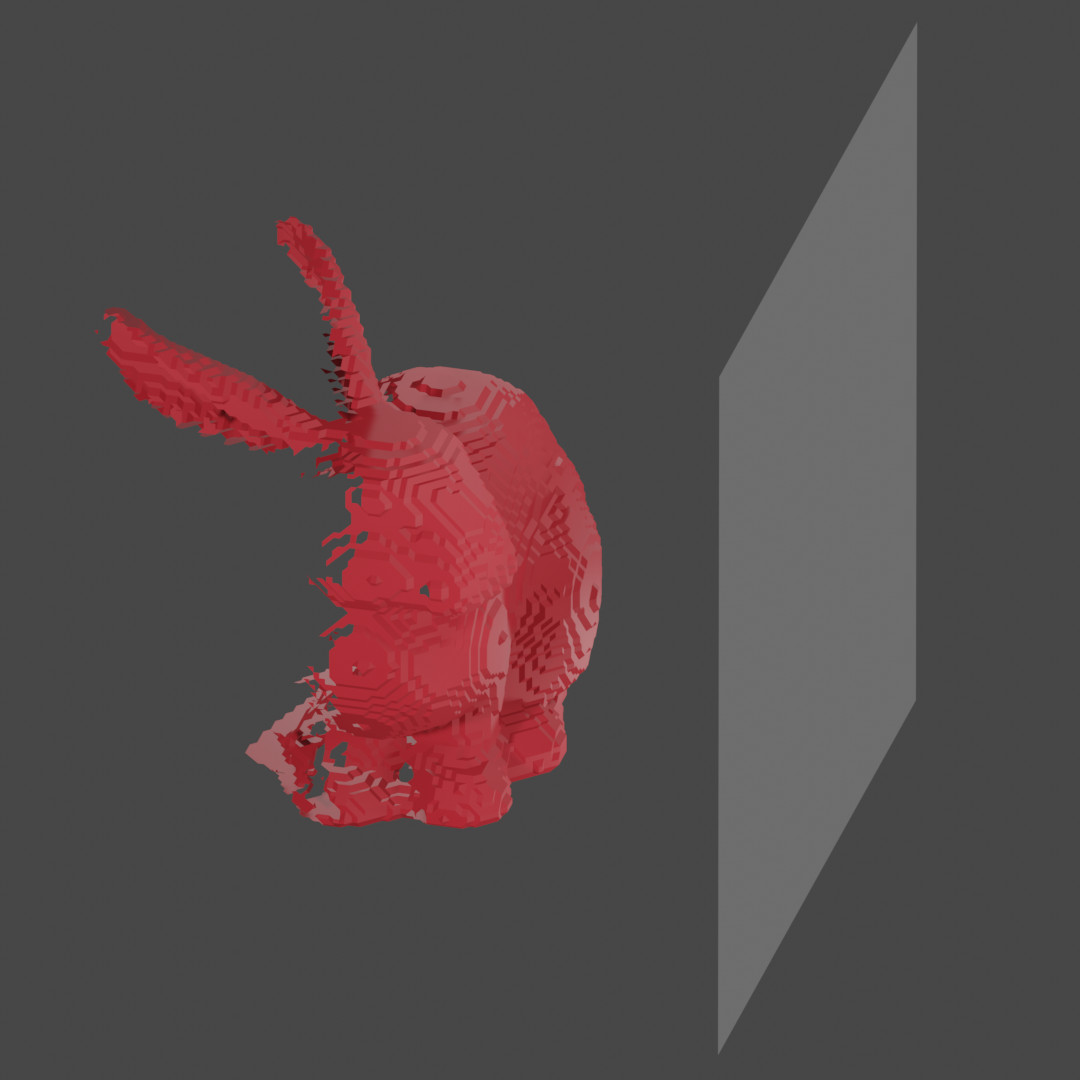}\\[0.1mm]
	%
	\caption{Example of Surface Extraction. \textbf{Left:} Raw prediction by our network after applying marching cubes. \textbf{Right:} Extracted surface after occlusion test. (Single-scene optimization)}
	\label{fig:surface_extraction}
	\vspace{-4mm}
\end{figure}

\noindent \textbf{Surface extraction:} Our networks predict occlusion scores for points sampled in the hidden scene. After applying marching cubes at inference, we obtain a closed mesh that couples the NLoS surface with the overall occlusion hull. In order to segment NLoS surface from shadow, we apply the same occlusion test used for data generation to the retrieved primitives, i.e we remove triangles whose centroids are occluded in the sense of \ref{eq:occlusion_def}. This allows us to retrieve the open NLoS surfaces that effectively cast light onto the wall (see Figure~\ref{fig:surface_extraction}).

\noindent \textbf{Recovery properties:} A major benefit of our representation is that of retrieving shape features beyond the specular Fermat criteria \cite{xin2019theory,liu2019analysis}. A simple way of visualizing this is by noting that the definitions for local and global occlusion do not consider normals. This feature naturally sacrifices accuracy in the primitives orientations but offers in turn the gain of retrieving any visible point at the cost of an approximate normal. This choice follows the original idea of \cite{kirmani2009looking}, which proposes that all surface points casting light onto wall regions can be reconstructed. As resolving non-Fermat features remains a very hard problem for physically-based NLoS and low-resolution setups, we use this design choice to motivate our goal of learning data-priors for the task of NLoS dense reconstruction.

\noindent \textbf{Self-occlusion}: Another benefit of our representation implies that a shape primitive is recoverable if it is locally visible for at least $k$ wall sensors. This way, portions of the hidden surface that are occluded for some wall regions may be visible for others, thus making them globally visible according to \ref{eq:occlusion_def} and hence recoverable.

\noindent \textbf{Architecture:} The choice of $L_{occl}$, $G$, $F$ and $H$ specify different occlusion models. We design our pipeline to be compatible with state-of-the-art architectures developed for 3D reconstruction \cite{sitzmann2020implicit,chabra2020deep,chibane2020implicit,park2019deepsdf}. In these frameworks, the general idea is to encode measurements and sampling points separately prior to computing the score function. Since we focus on binary occlusion throughout this work, we evaluate existing networks for that case (\cite{peng2020convolutional,2019occupancy}) and make no major claims on architectural contributions. We tailor the networks to process time-of-flight inputs and train them end-to-end. Alternatively, $F$ could be modeled as an explicit inverse as in \cite{chen2020learned}, or a pre-trained module that solves the inverse problem (e.g for albedo, depth maps, etc). Analogously, $H$ could be realized, for example, as MLP modules with sine-cosine encodings as in \cite{mildenhall2020nerf,tancik2020fourier}.

\section{Results and Experiments}\label{section:results}

\textbf{Scene Configuration Space:} In order to predict unseen targets in the NLoS region we must endorse our network with translational, rotational and scale invariance during training. When generating training scenes, this results in a much larger space of configurations than the single box-aligned scenes used by most authors in $3D$ modeling literature. For the NLoS case each source mesh must sampled $n$ times, each of which constitutes a random affine transform of the object. This requirement imposes hard trade-offs between generalization, dataset size and computational resources. When $n$ is too small, networks may generalize poorly on arbitrary test scenes. If $n$ its too large then dataset generation can result prohibitively expensive in either runtime, training time and/or disk storage. 

\noindent \textbf{Datasets:}  Given the size of the configuration space, we consider smaller categorical partitions than the one introduced by Choy et. al. \cite{choy20163d} (44,000 models across 13 categories of the ShapeNet database \cite{shapenet:paper}) when generating our $3D$ scenes. We sample random configurations from the categories cars, cameras, mugs, couches, bikes, guns and our own class collections of flat geometric figures, letters of different fonts and statues from the ``Scan the World'' database \cite{scantheworld}. In order to keep storage requirements and training time feasible, we consider a maximum budget of 70,000 scenes over pools of 5-7 categories, each containing 100-600 meshes. Our scenes are generated within unit cube space using moderate ranges for the affine transformations (see supplemental).

To render transient images, we use the physically-based strategy by Iseringhausen \etal\cite{iseringhausen2020non} and generate volumes of $32\times 32 \times 256$ with 32\,ps and 64\,ps resolutions in order to simulate low-resolved setups of those by previous approaches \cite{otoole2018confocal,lindell2019wave}. We compute occlusions with our occlusion/visibility test by sampling points partly over a shape's surface and uniformly over the entire $3D$ scene, as done by Park \etal~\cite{park2019deepsdf}. For most experiments we used 400,000 points per scene.

\noindent \textbf{Representation Implementation:} We use Nvidia's OptiX framework to implement our occlusion/visibility test. An OptiX context is launched to trace $m$ rays from each sampling point to each virtual sensor. If a ray unobstructedly intersects the wall, then the point is visible, otherwise it is occluded. Overall, our implementation achieves close to linear runtime over all of its inputs (\# sampling points, \# virtual sensors, \# mesh triangles) as shown in the supplemental.

\noindent \textbf{Training:} We tailored the architectures from Peng \etal~\cite{peng2020convolutional} and Mescheder \etal~\cite{2019occupancy} in Pytorch in order to compute binary occlusions from transient images. We let our trainings run for about 3--4 days (40-60 epochs) using batch sizes of 6-8 scenes, a $20k$ batch of points, and then choose the predictive model as the one with best validation score. Results of experiments in this manuscript correspond to trainings done with the convolutional architecture, as this one showed better predictive performance on sharp features.

\noindent \textbf{Error metrics:} We optimize our networks to minimize the binary-cross-entropy on training scenes. Similar to most works we find that the most robust metrics to evaluate our models are F-score, intersection-over-union (IoU) and the Chamfer distance computed on the tessellated surfaces.

\subsection{NLoS 3D Reconstruction}
Throughout this work, we establish the method by Iseringhausen \etal~\cite{iseringhausen2020non} as the state-of-the-art baseline for NLoS mesh reconstruction. The approach by \cite{tsai2019beyond} requires mesh initializations similar to the final object, thus assuming strong prior knowledge of the sought target. We argue that this method could be used as a latter refinement step, but we do not consider it to be a competitive baseline for fair comparisons. Due to reasons of space in this manuscript we refer the reader to the supplemental material for complementary evaluations and predictions.

\begin{figure*}[t]
	\centering
	\rotatebox{90}{\small~~~~~~GT\textcolor{white}p} 
	\includegraphics[trim=11mm 8mm 5mm 3mm,clip,width = 0.16\columnwidth]{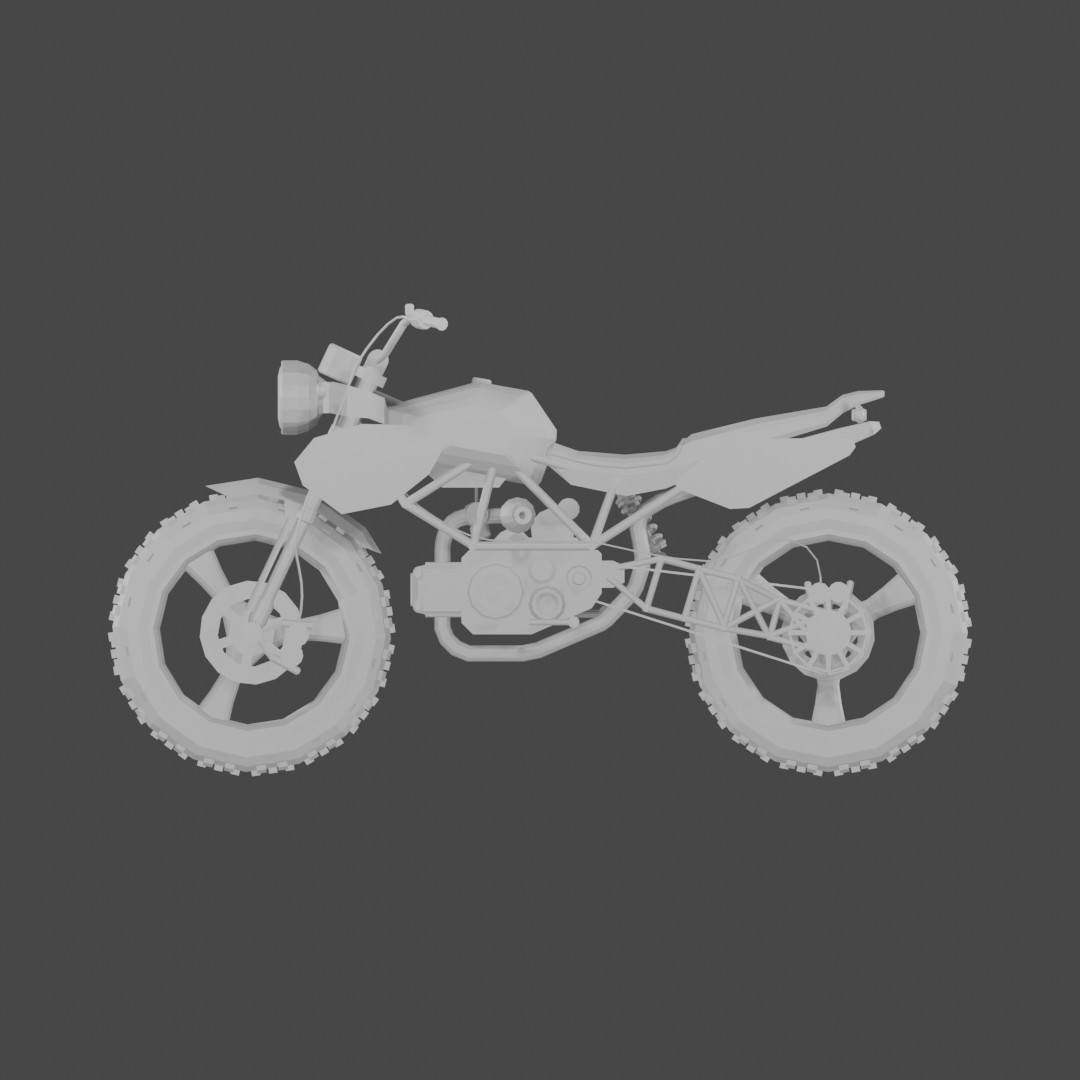}%
	\includegraphics[trim=11mm 8mm 5mm 3mm,clip,width = 0.16\columnwidth]{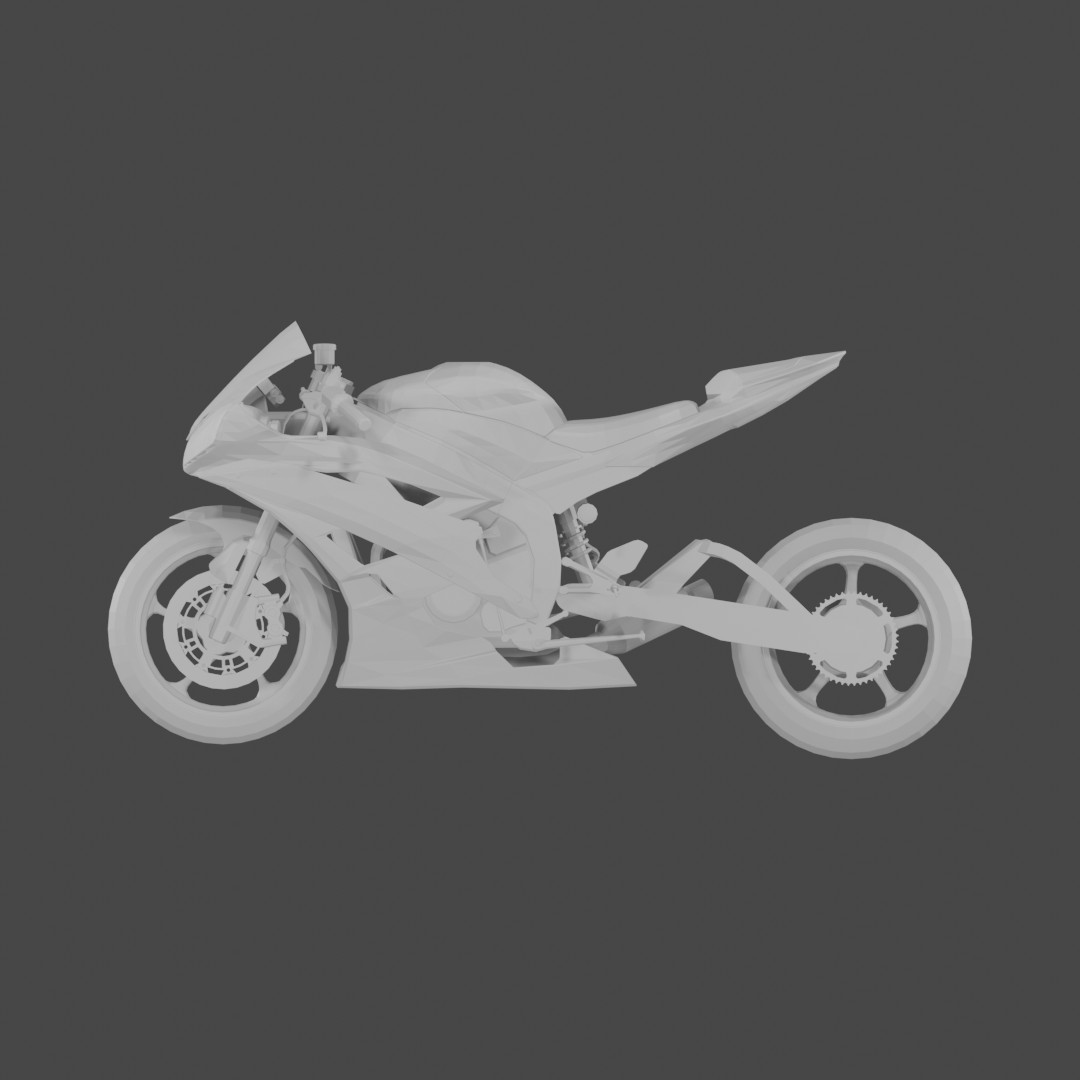}%
	\includegraphics[trim=11mm 8mm 5mm 3mm,clip,width = 0.16\columnwidth]{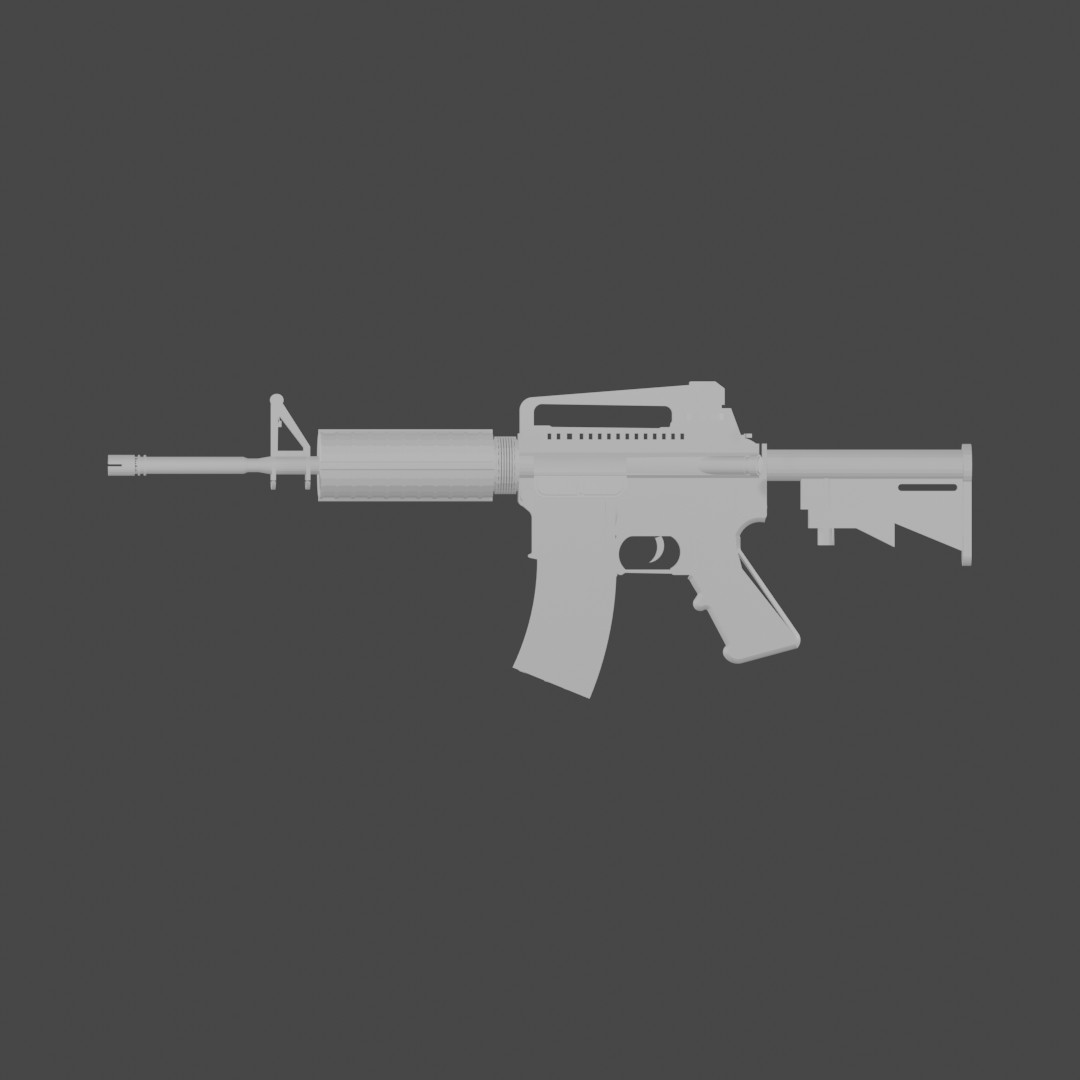}%
	\includegraphics[trim=11mm 8mm 5mm 3mm,clip,width = 0.16\columnwidth]{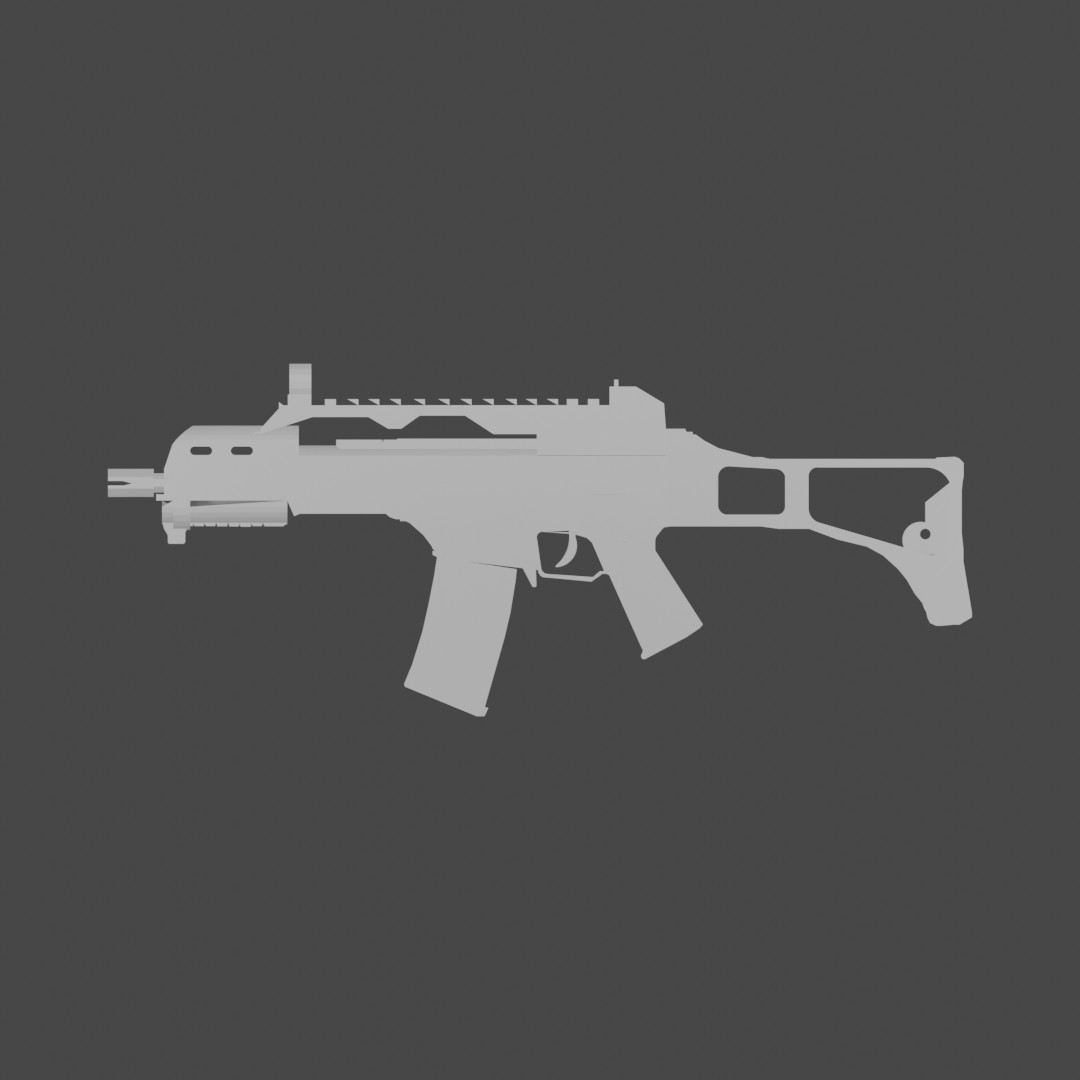}%
	\includegraphics[trim=11mm 8mm 5mm 3mm,clip,width = 0.16\columnwidth]{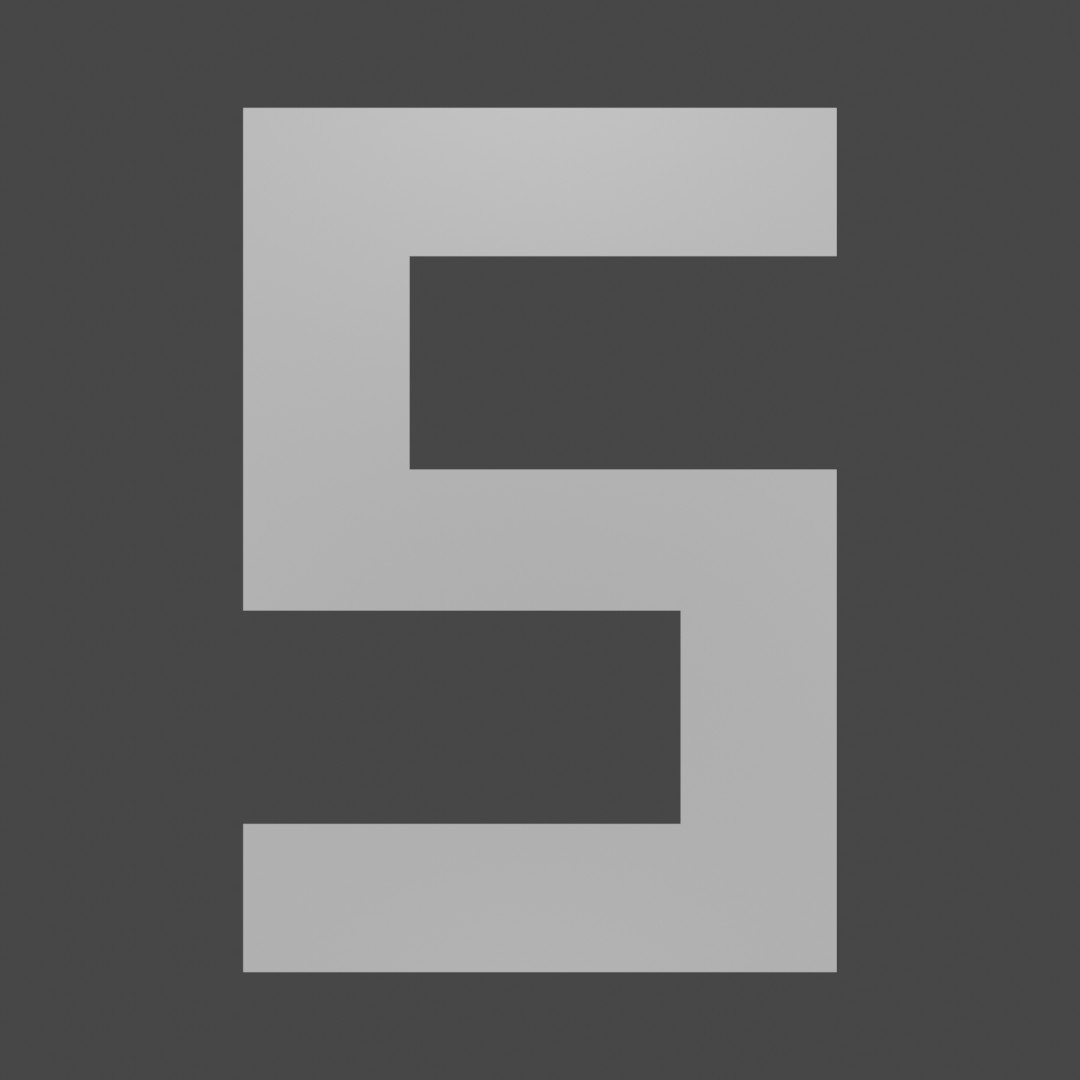}
	\includegraphics[trim=11mm 8mm 5mm 3mm,clip,width = 0.16\columnwidth]{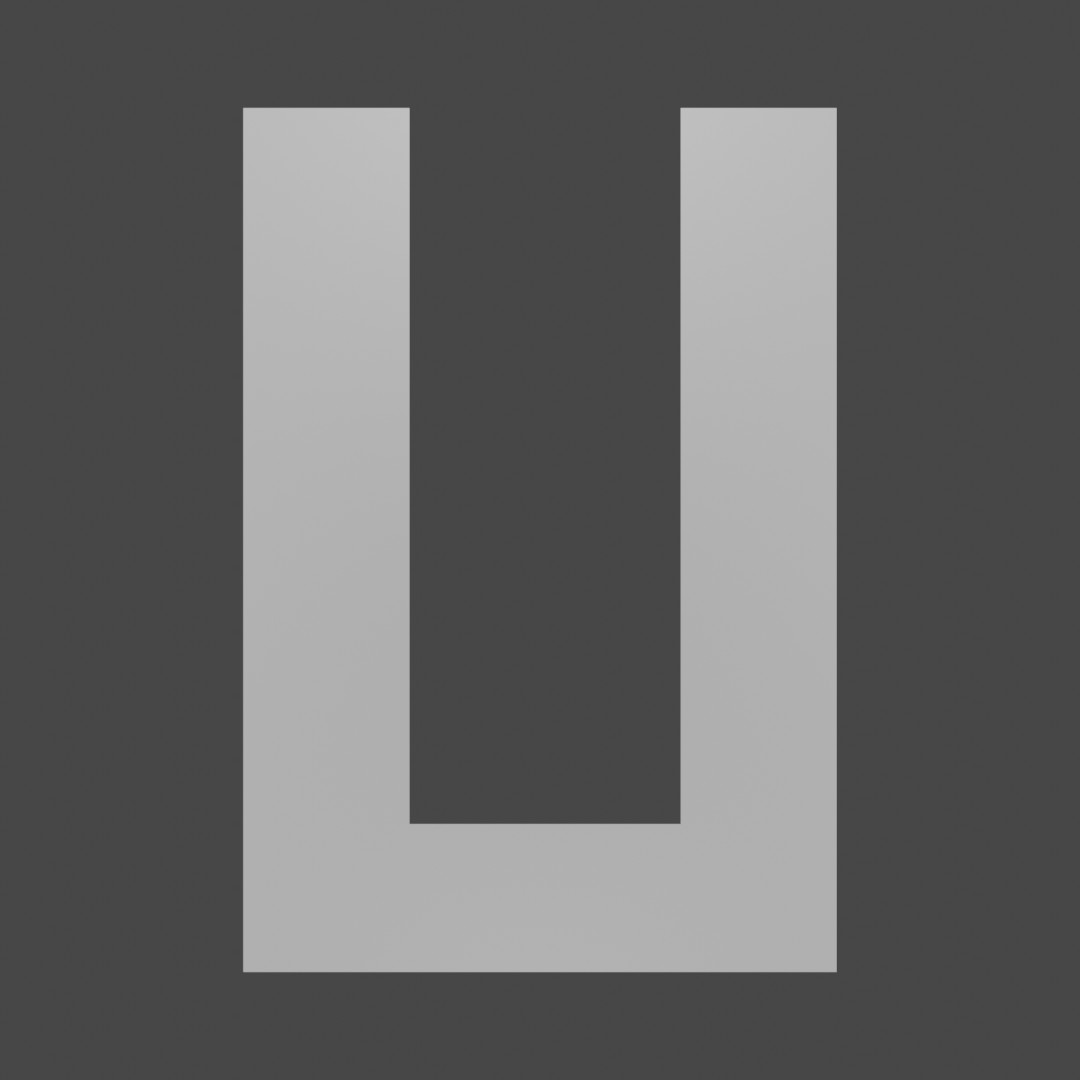}\\[1mm]
	\rotatebox{90}{\small Iseringhausen\textcolor{white}p} 
	\includegraphics[trim=11mm 8mm 5mm 3mm,clip,width = 0.16\columnwidth]{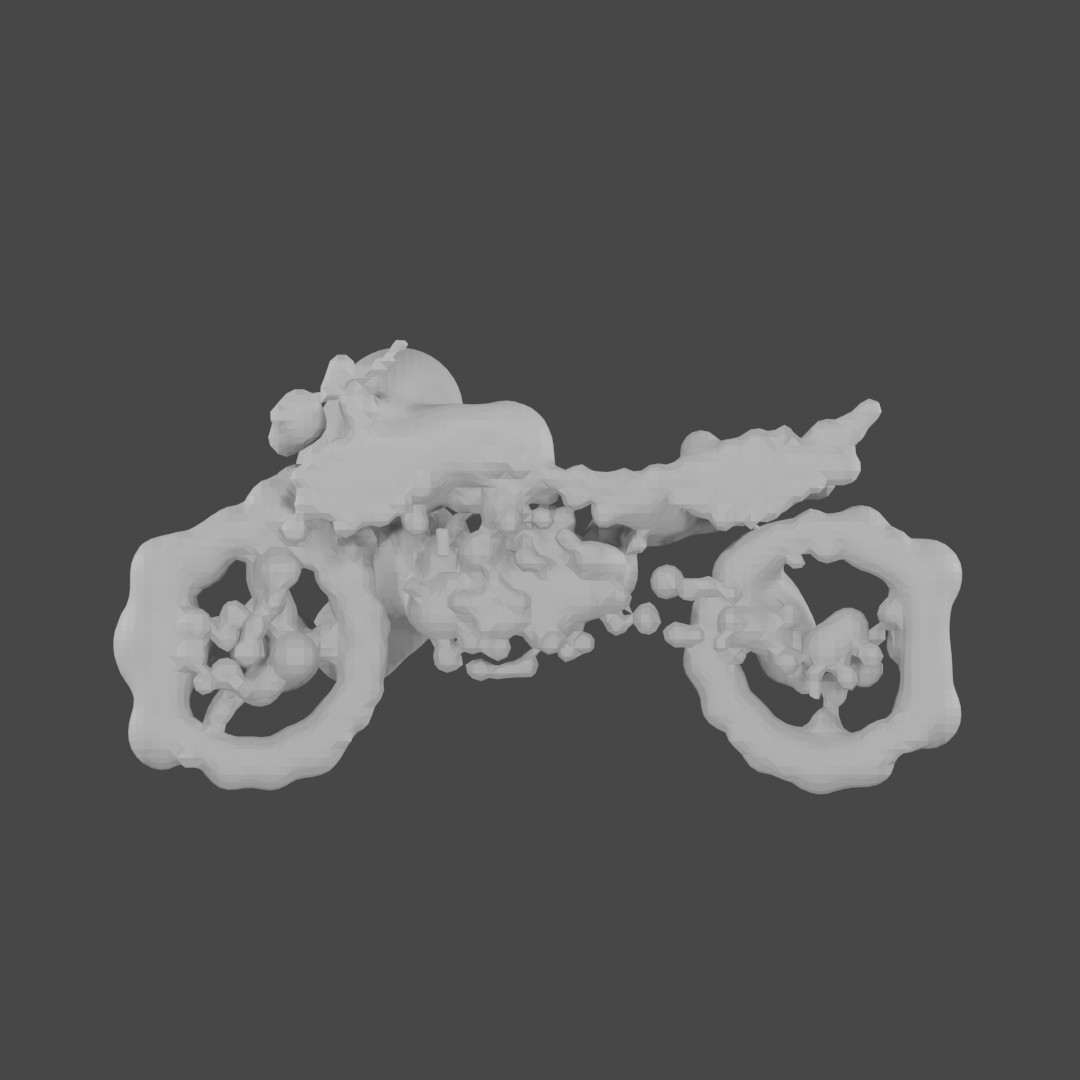}%
	\includegraphics[trim=11mm 8mm 5mm 3mm,clip,width = 0.16\columnwidth]{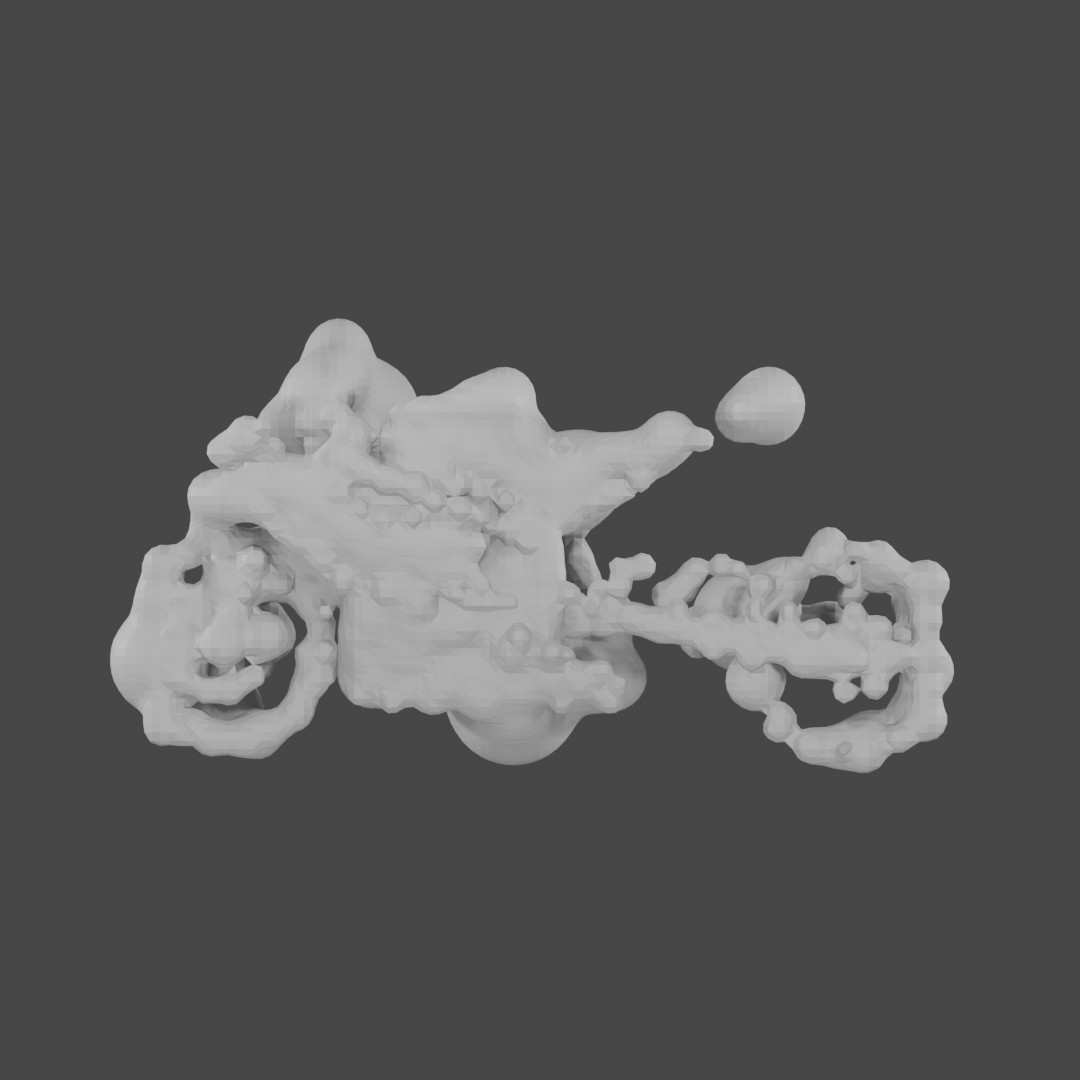}%
	\includegraphics[trim=11mm 8mm 5mm 3mm,clip,width = 0.16\columnwidth]{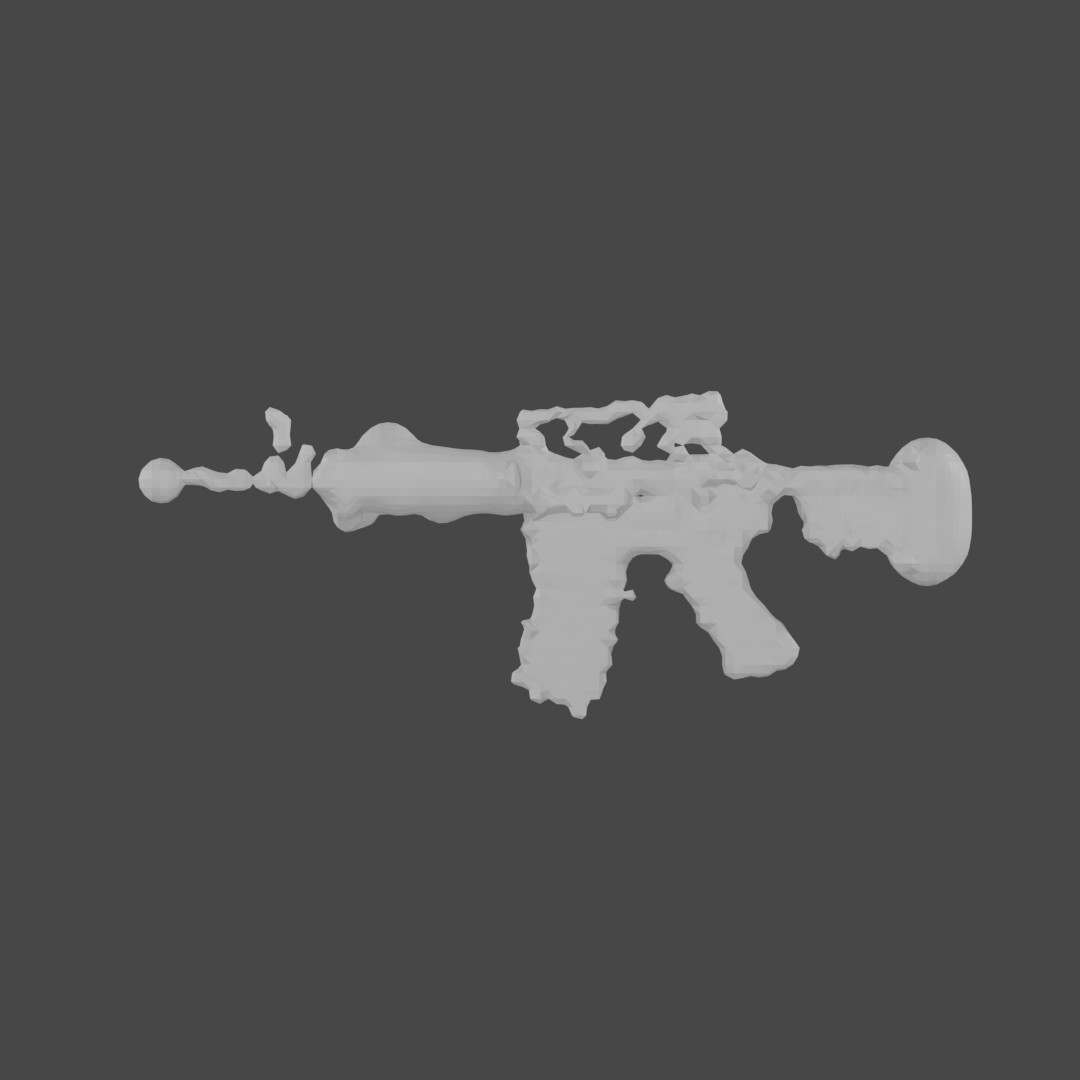}%
	\includegraphics[trim=11mm 8mm 5mm 3mm,clip,width = 0.16\columnwidth]{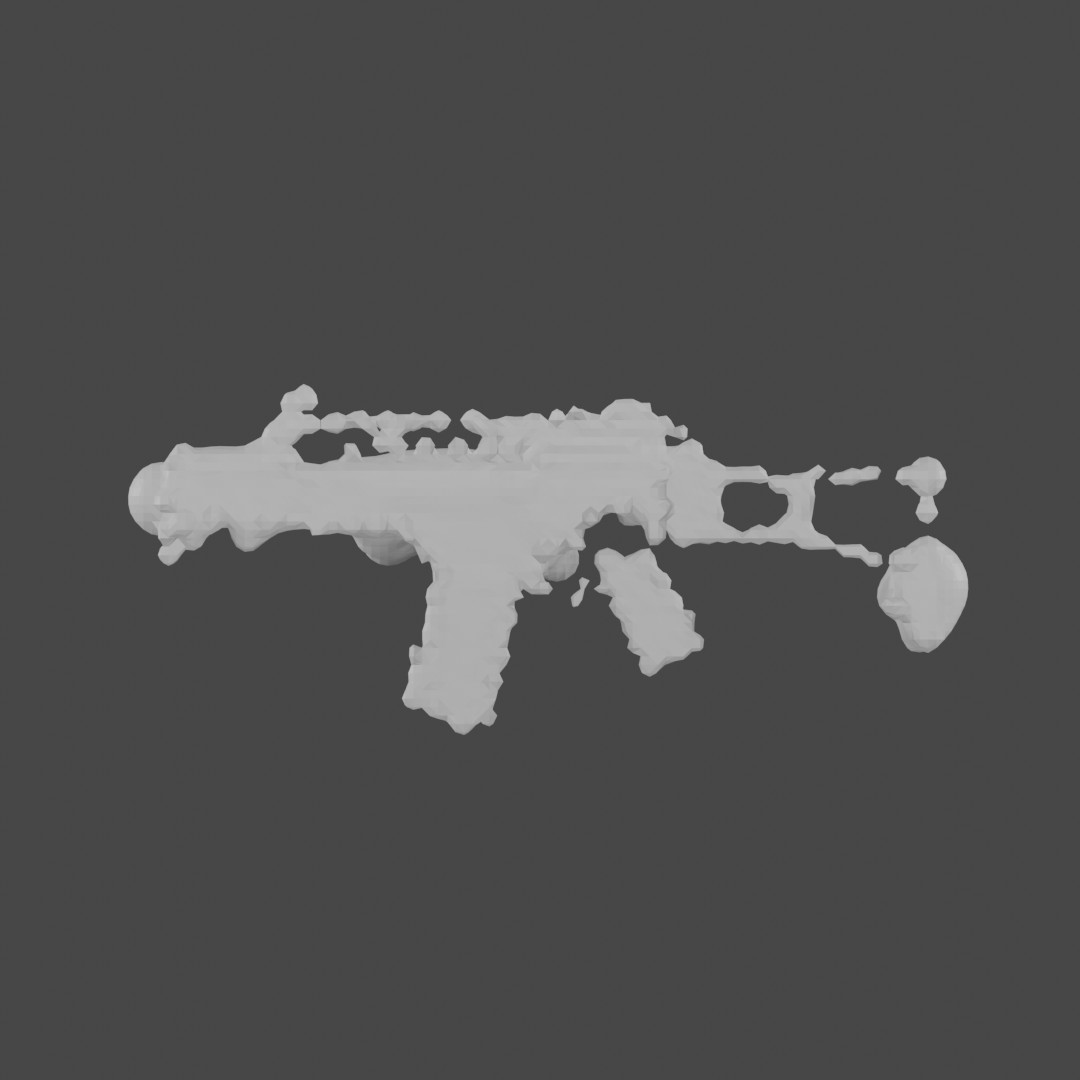}%
	\includegraphics[trim=11mm 8mm 5mm 3mm,clip,width = 0.16\columnwidth]{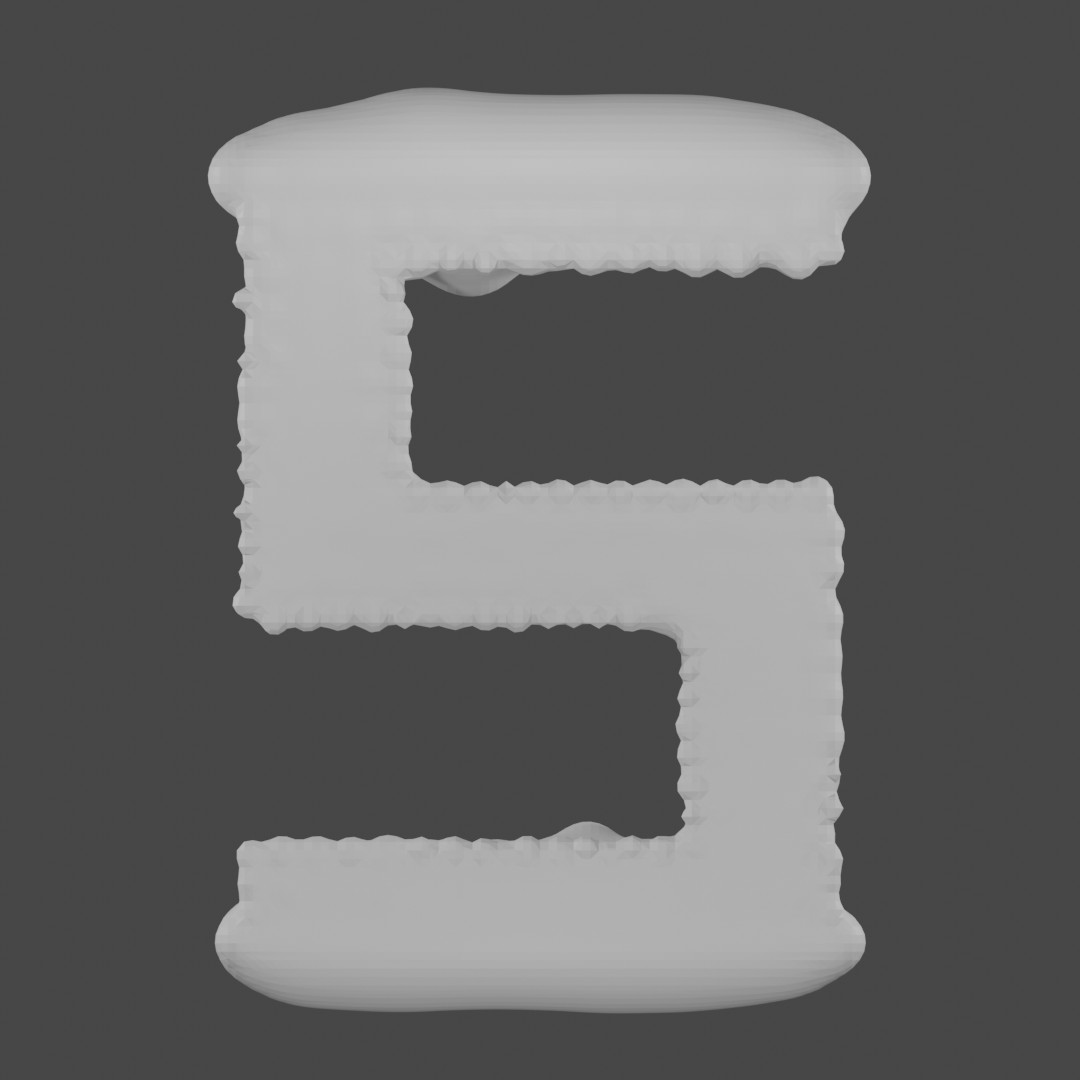}
	\includegraphics[trim=11mm 8mm 5mm 3mm,clip,width = 0.16\columnwidth]{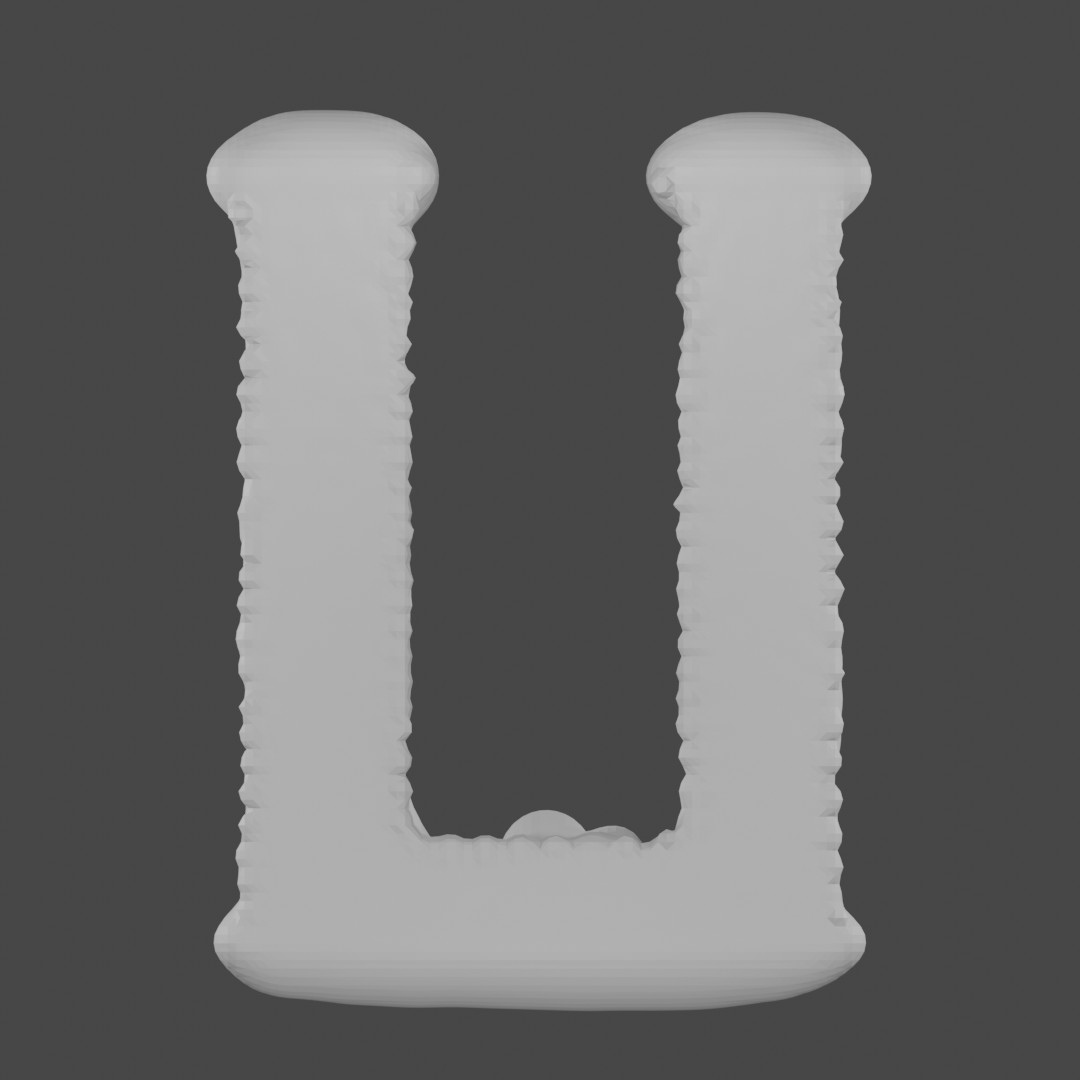}\\[1mm]
	\rotatebox{90}{\small~~~~~Ours\textcolor{white}p} 		
	\includegraphics[trim=11mm 8mm 5mm 3mm,clip,width = 0.16\columnwidth]{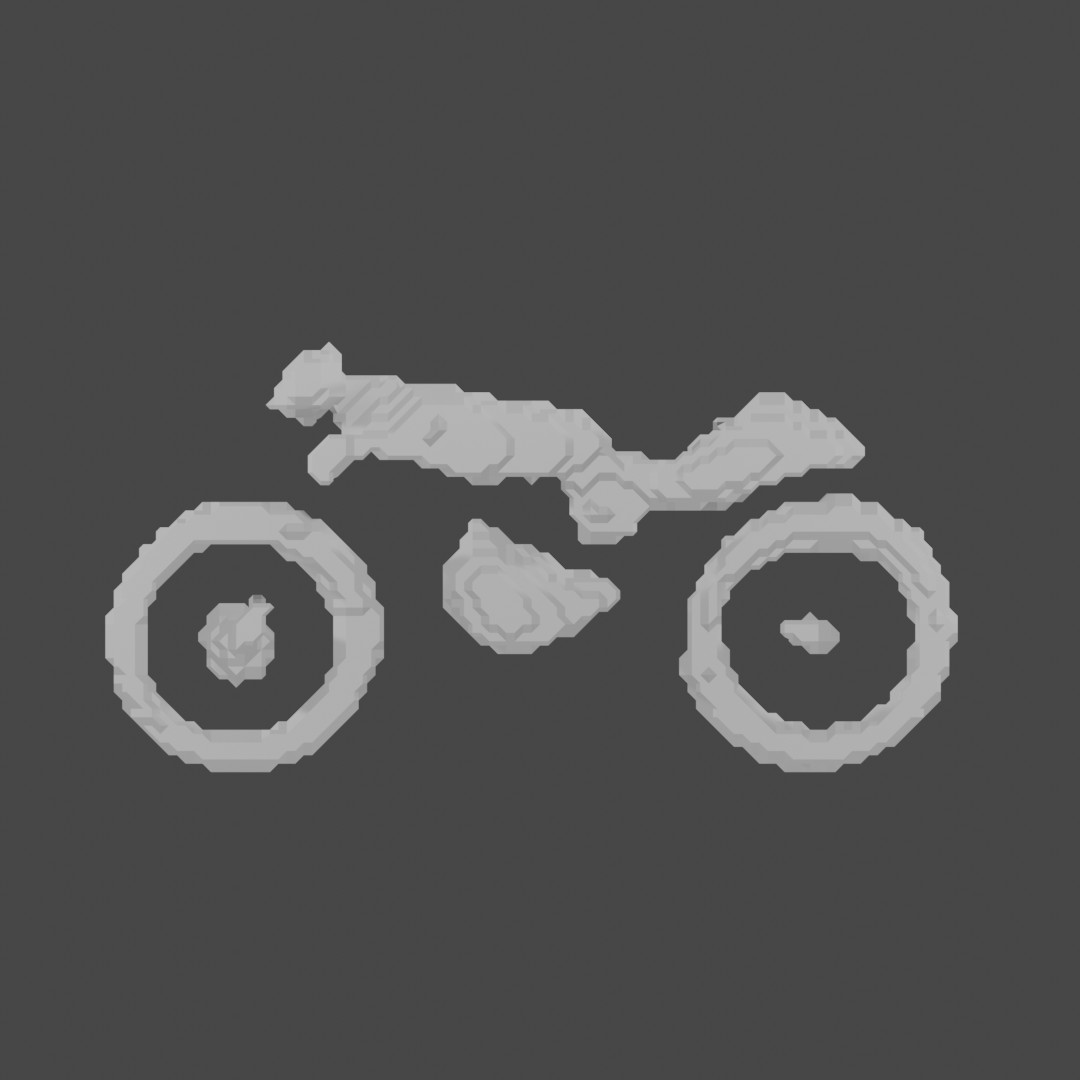}%
	\includegraphics[trim=11mm 8mm 5mm 3mm,clip,width = 0.16\columnwidth]{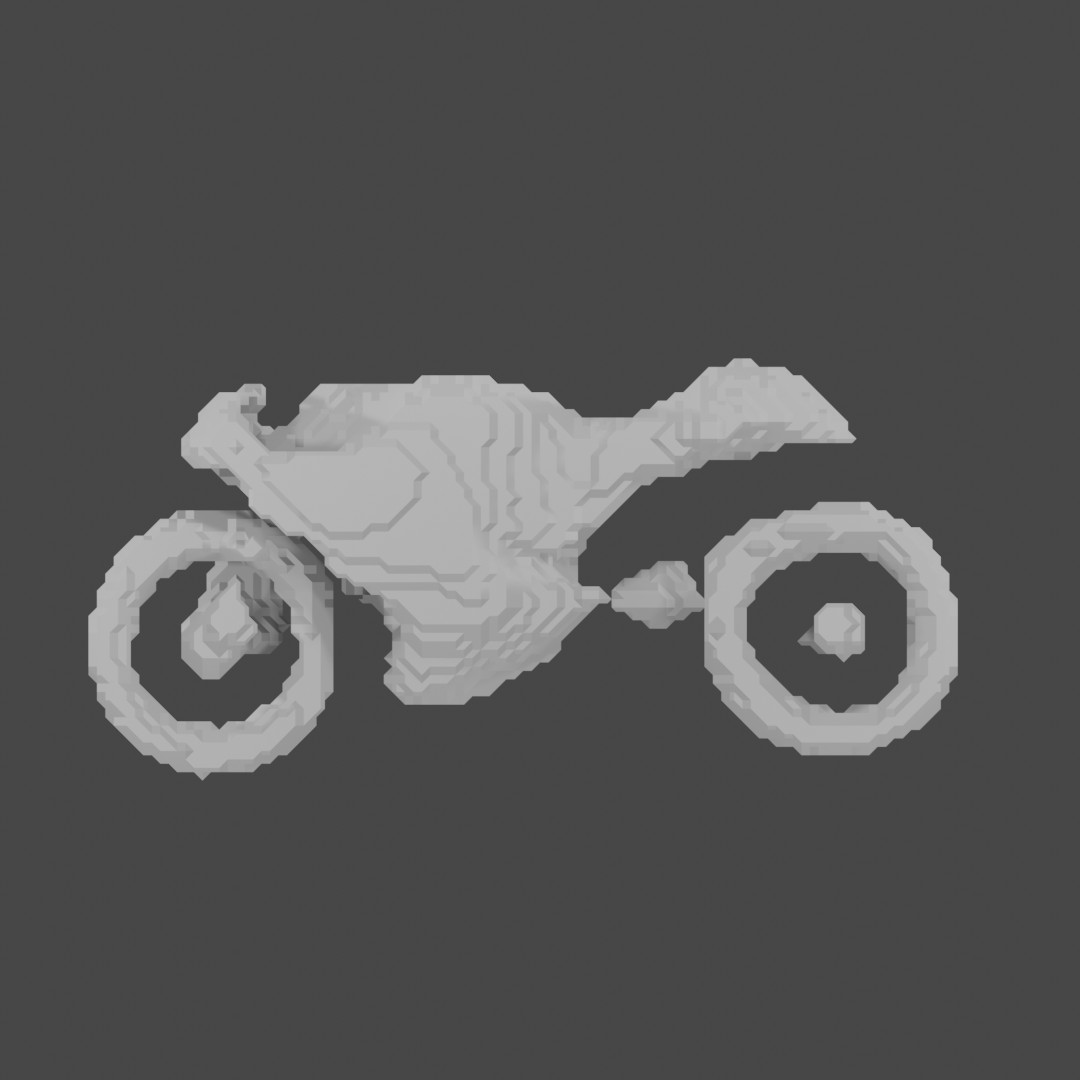}%
	\includegraphics[trim=11mm 8mm 5mm 3mm,clip,width = 0.16\columnwidth]{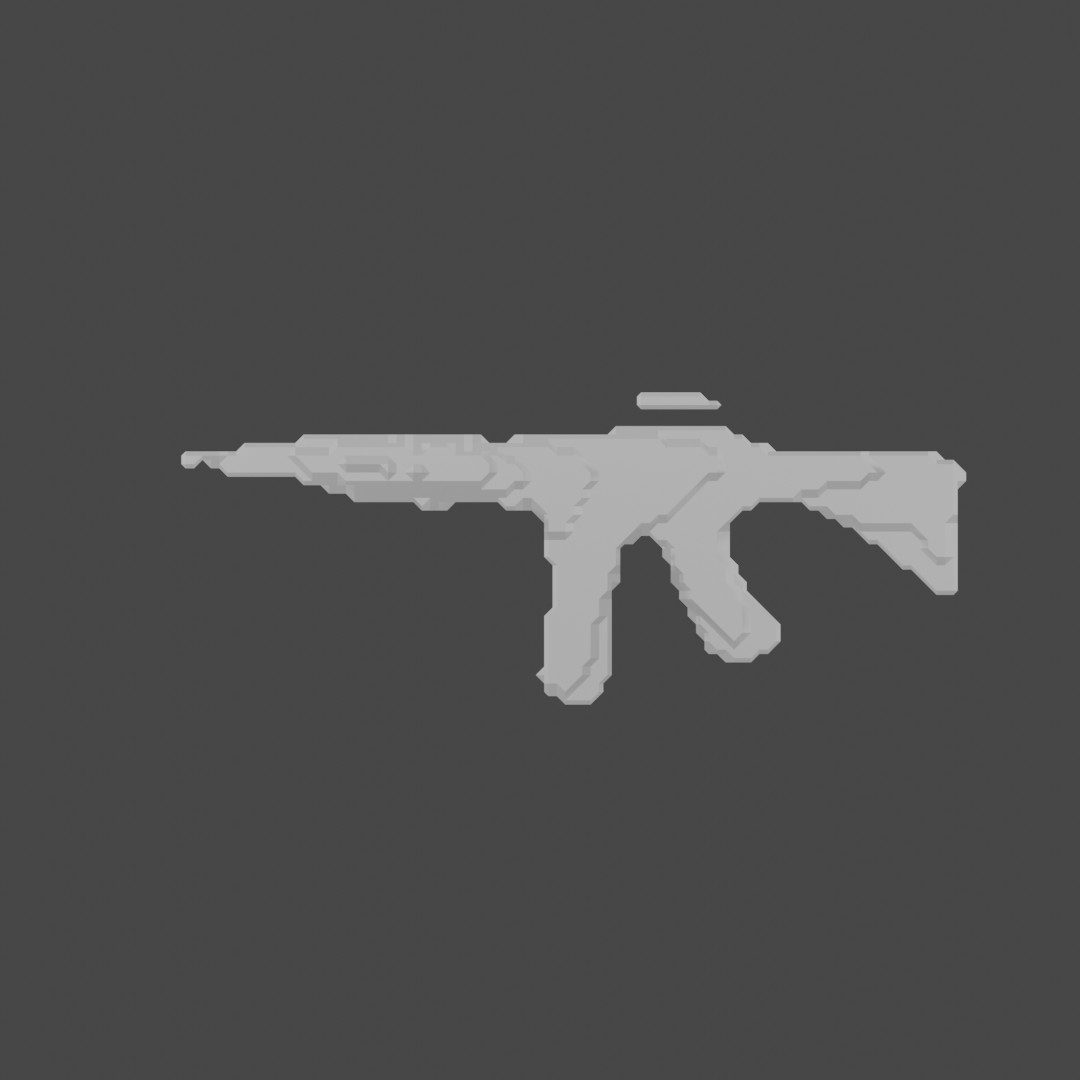}%
	\includegraphics[trim=11mm 8mm 5mm 3mm,clip,width = 0.16\columnwidth]{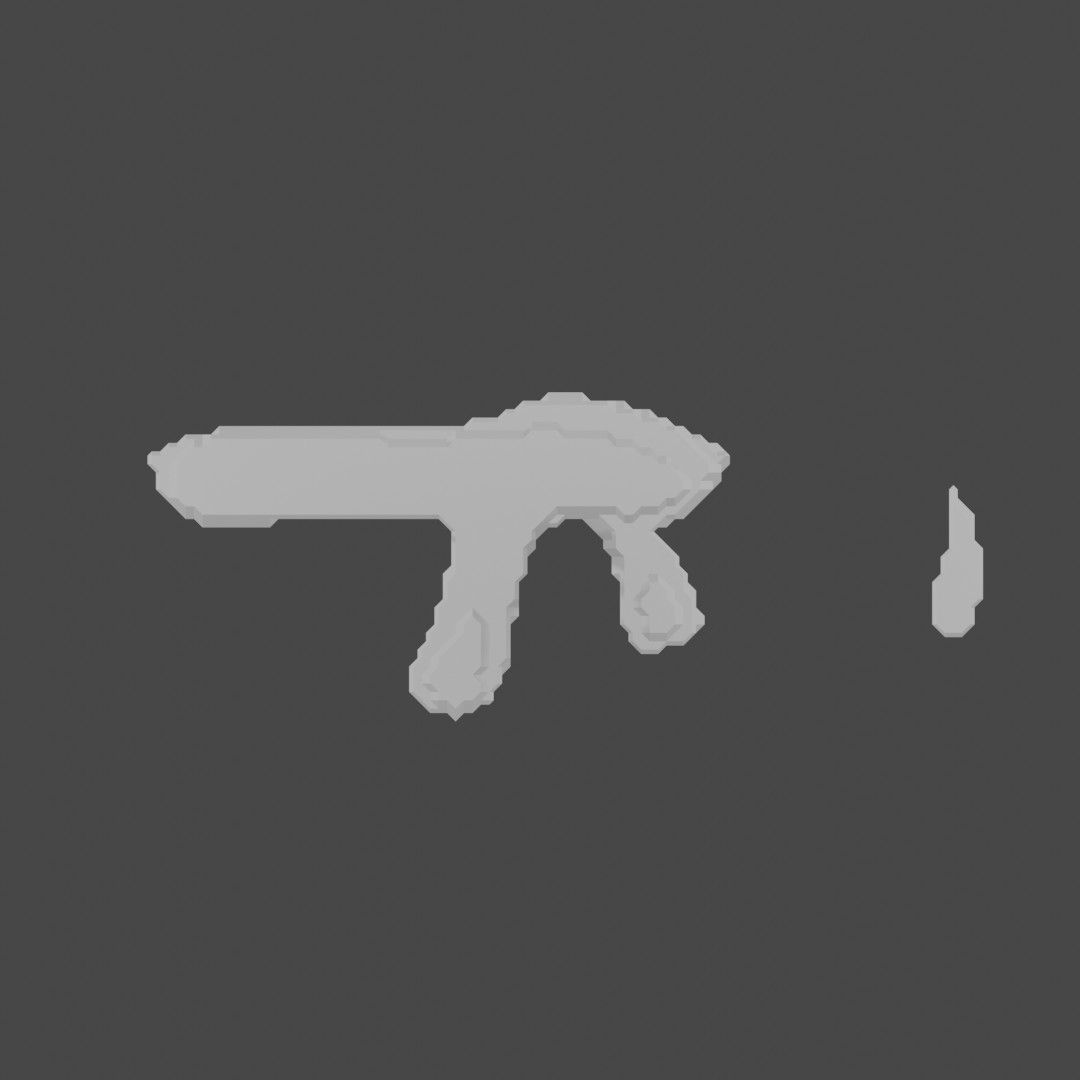}%
	\includegraphics[trim=11mm 8mm 5mm 3mm,clip,width = 0.16\columnwidth]{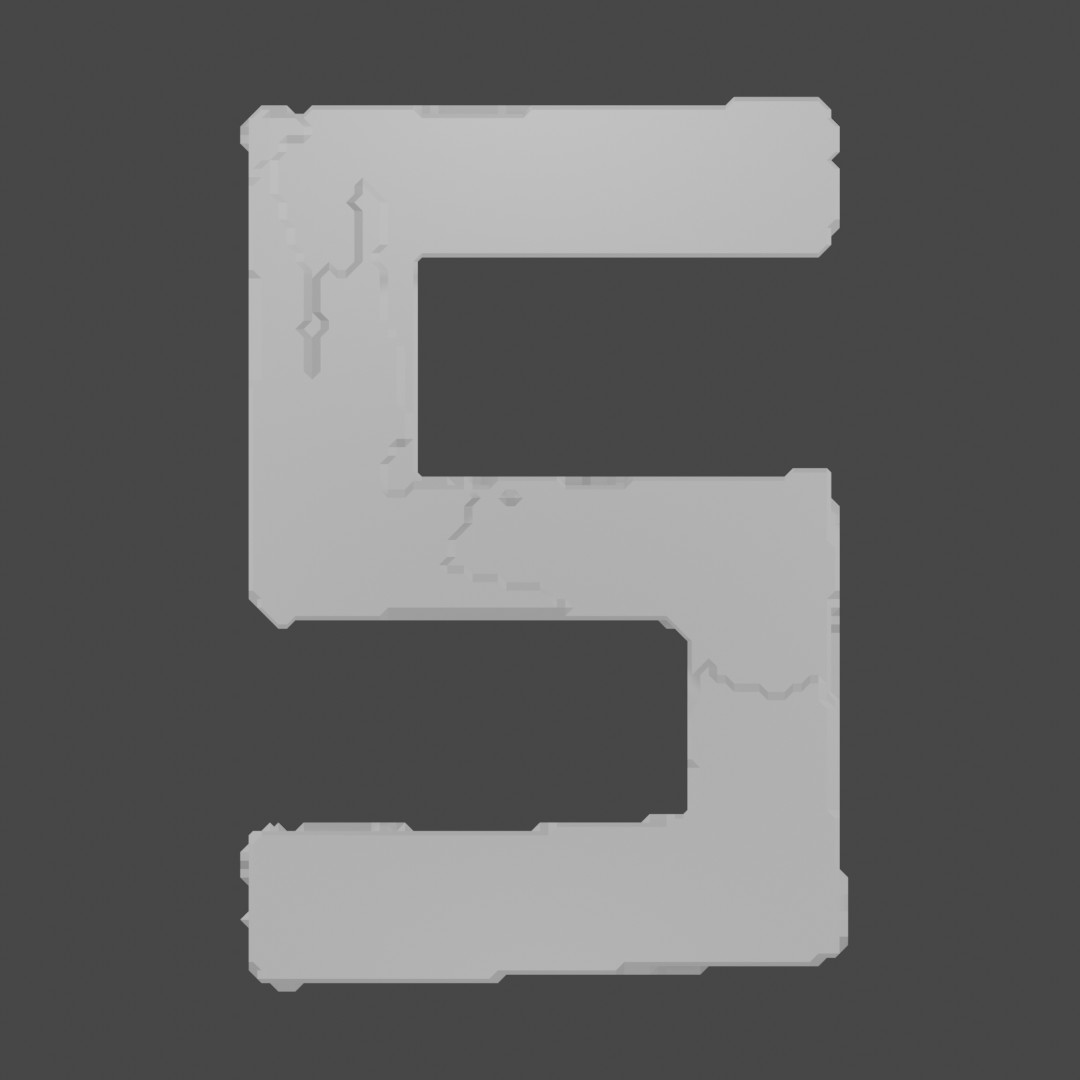}
	\includegraphics[trim=11mm 8mm 5mm 3mm,clip,width = 0.16\columnwidth]{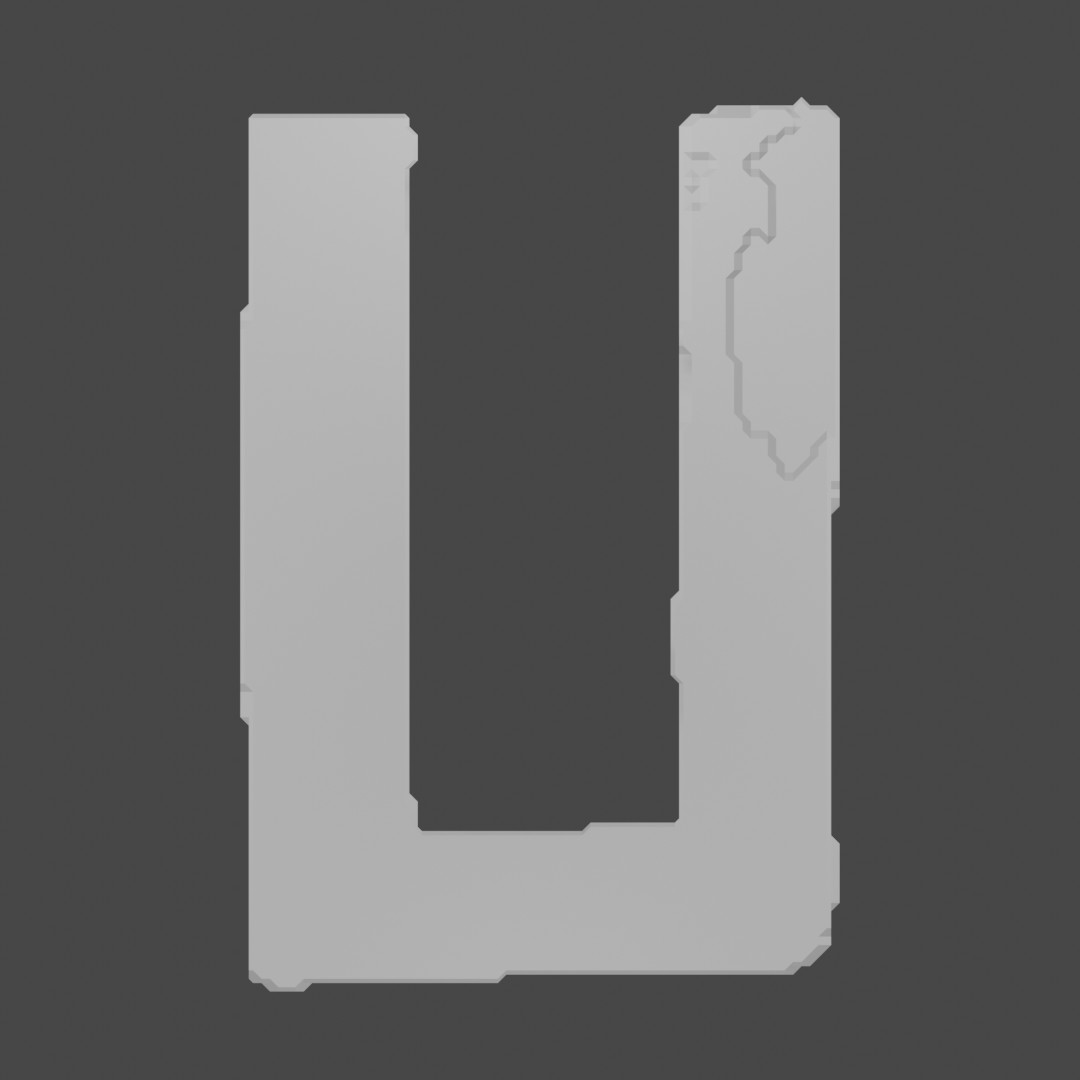}\\[1mm]
	%
	\vspace{-2mm}
	\caption{Comparison between our model and the one by \cite{iseringhausen2020non}. Our model (bottom row) predicts surfaces with less noise and sharper edges than the baseline (middle row). The scanned area on the wall is 70\,cm$\times$70\,cm large, sampled with an array of $32\times 32$ scan positions at 32\,ps resolution, resembling a low-res capture of the one by \cite{otoole2018confocal}. Targets are placed 35\,cm from the wall.}
	\label{fig:results_1}
\end{figure*}

Figure \ref{fig:results_1} shows reconstructions on unseen scenes after training over 40,000 examples sampled over three classes of objects. Overall, our model is able to make competitive predictions compared to the baseline method while producing smoother surfaces. For the bikes and guns scenes we observe that our method misses thin features of the shapes but predicts global structure with less noise and that remains more faithful to the underlying geometry (see Table \ref{tab:bikegunletter_examples_chamfer}). In the case of the letter scenes our approach predicts surfaces with sharp edges, thus indicating high-frequency capability.

\begin{table}[htbp]
	\makebox[\textwidth][c]{
		\begin{tabular}{cc|cccccc}
			\midrule
			Model 			& & ~Bike 1~ 	& ~Bike 2~ 		& ~Gun 1~ 	& ~Gun 2~  	& ~Letter 1~ 	& ~Letter 2~ \\ \hline
			Conv-OccNet   	& & 0.487  		& 0.352			& 0.200		& 0.387		& 0.285			& 0.229 	\\
			CBN-Cell   		& & 0.509  		& 0.318			& 0.147		& 0.449		& 0.304			& 0.488 		\\
			Iseringhausen   & & 1.268		& 0.864			& 0.559		& 0.622	 	& 1.006			& 1.053		\\
			\bottomrule
		\end{tabular}%
	}
	\vspace{1mm}
	\caption{Chamfer distances ($\times 10^{-3}$) for example datasets from Figure \ref{fig:results_1} computed within a unit cube scene. Both models trained with our representation (Conv-OccNet and CBN-cell) perform similarly on the test sets while outperforming the method by \cite{iseringhausen2020non}.}
	\label{tab:bikegunletter_examples_chamfer}%
\end{table}%

\begin{figure*}[t]
	\centering
	\rotatebox{90}{\small~~~~~~GT\textcolor{white}p} 
	\includegraphics[trim=11mm 8mm 5mm 3mm,clip,width = 0.16\columnwidth]{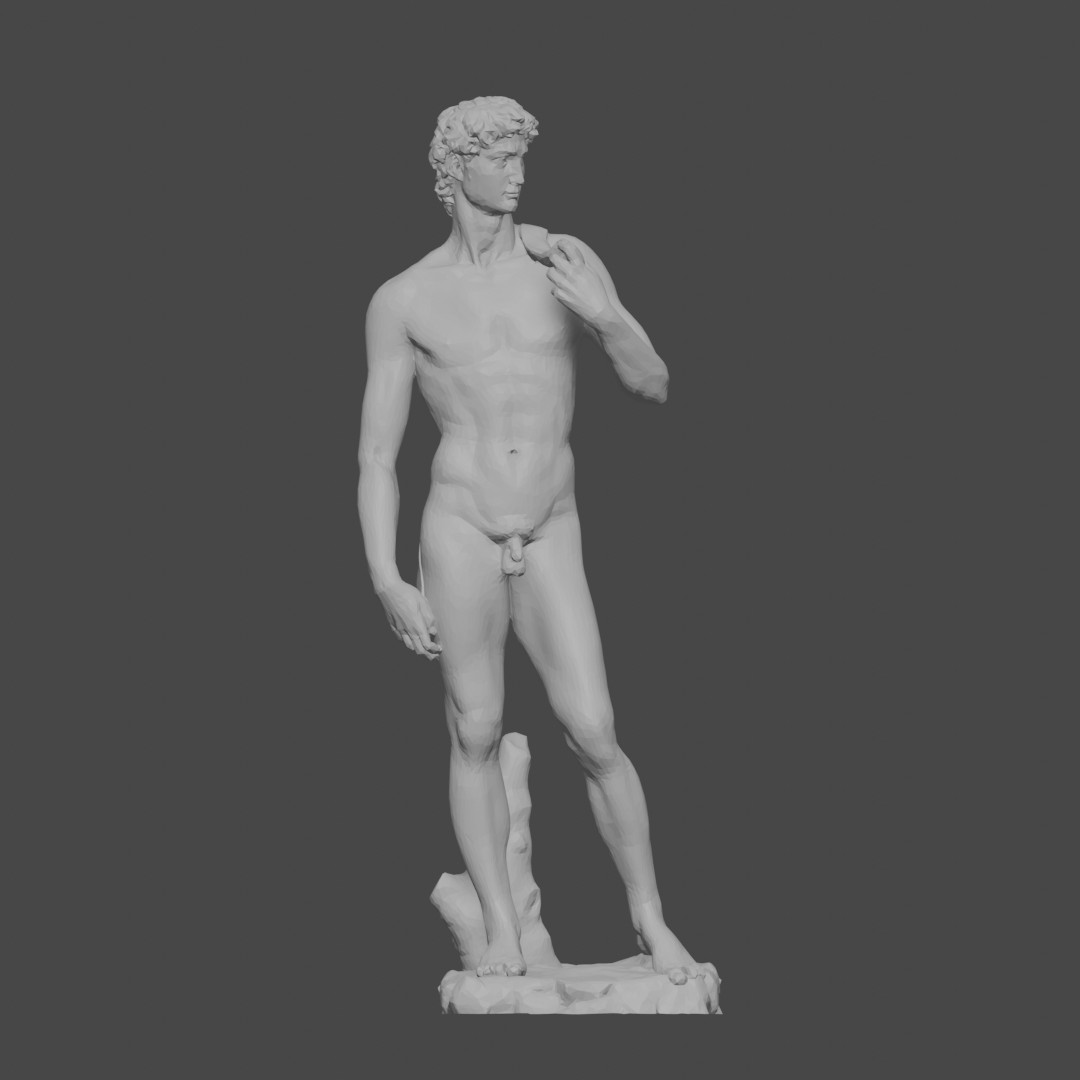}%
	\includegraphics[trim=11mm 8mm 5mm 3mm,clip,width = 0.16\columnwidth]{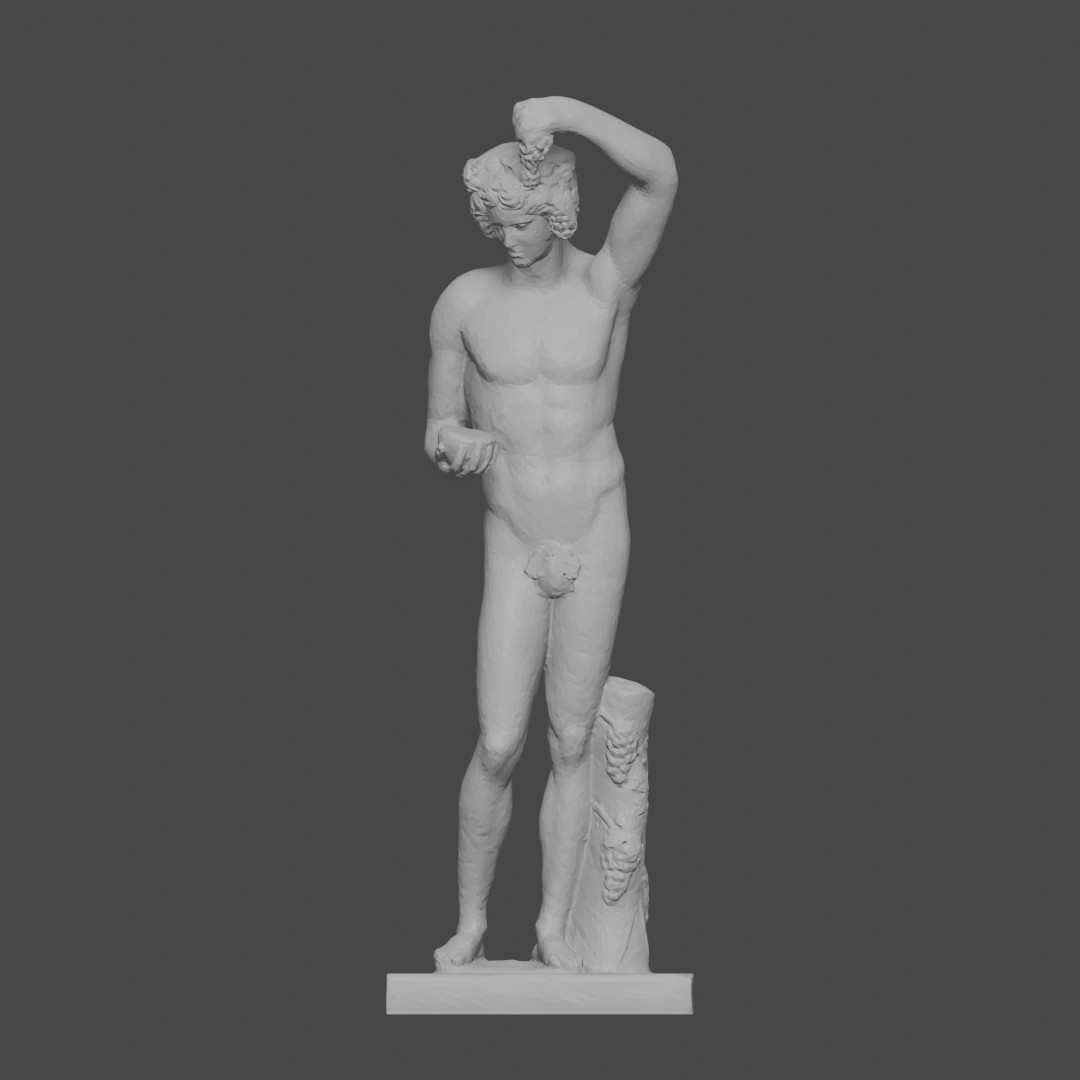}%
	\includegraphics[trim=11mm 8mm 5mm 3mm,clip,width = 0.16\columnwidth]{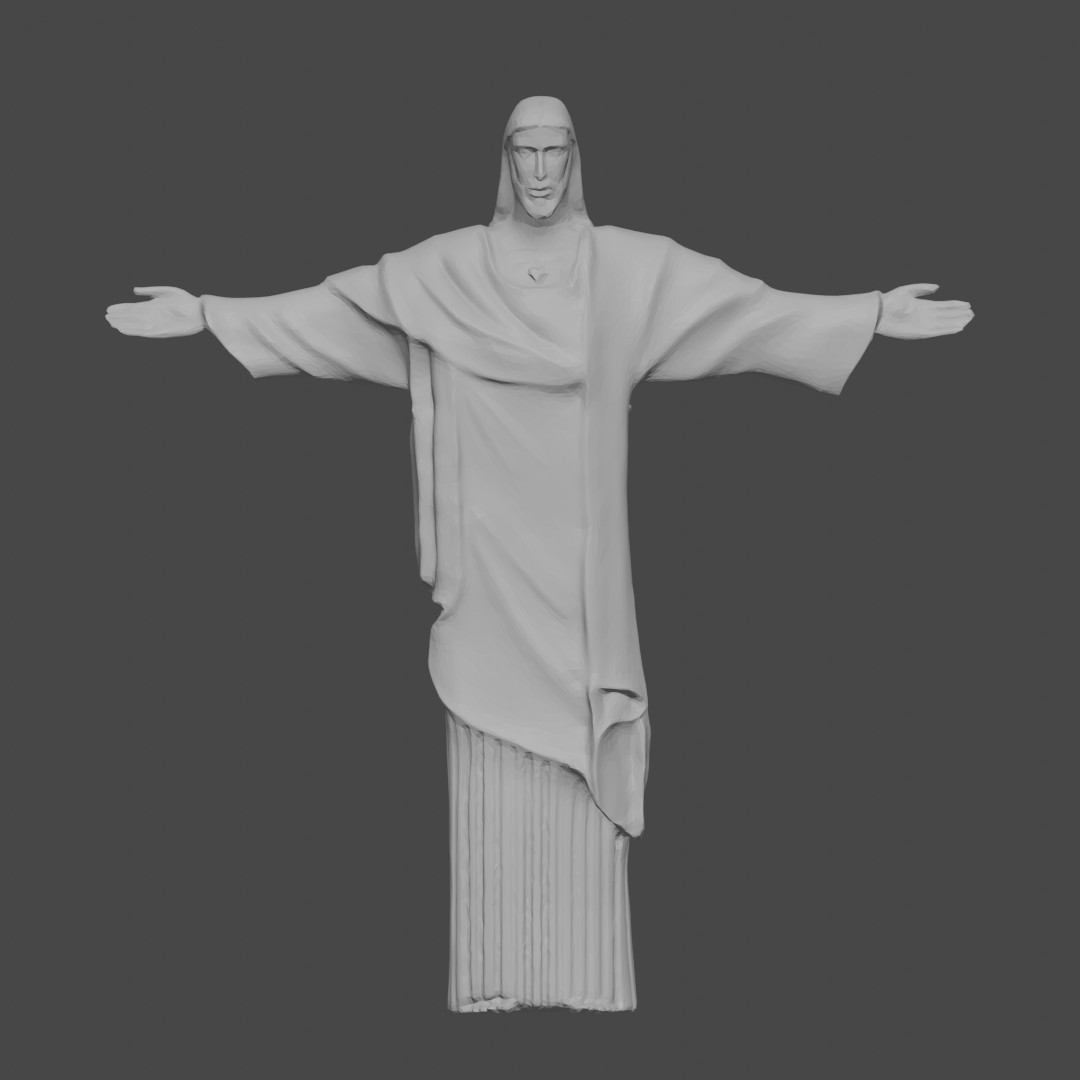}%
	\includegraphics[trim=11mm 8mm 5mm 3mm,clip,width = 0.16\columnwidth]{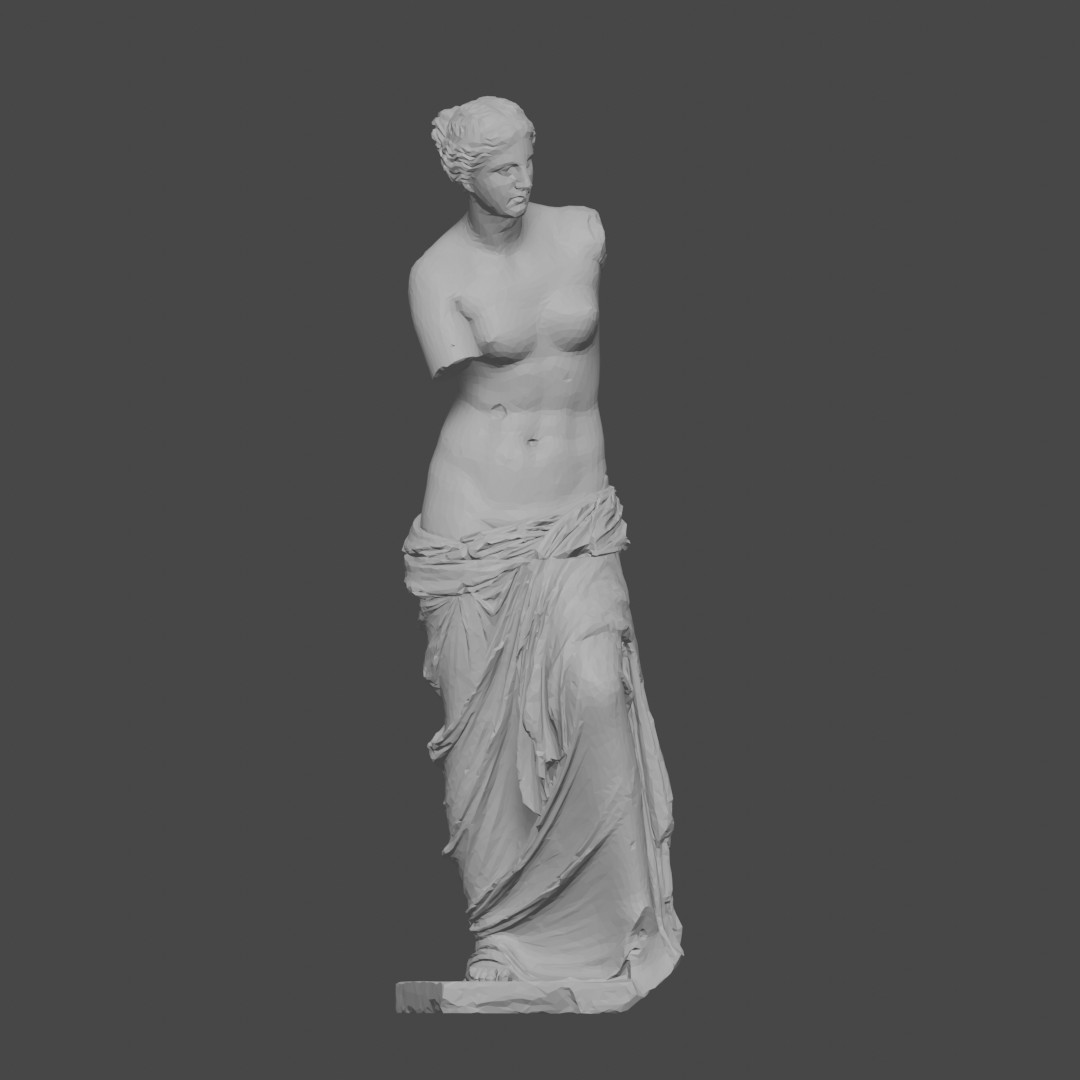}%
	\includegraphics[trim=11mm 8mm 5mm 3mm,clip,width = 0.16\columnwidth]{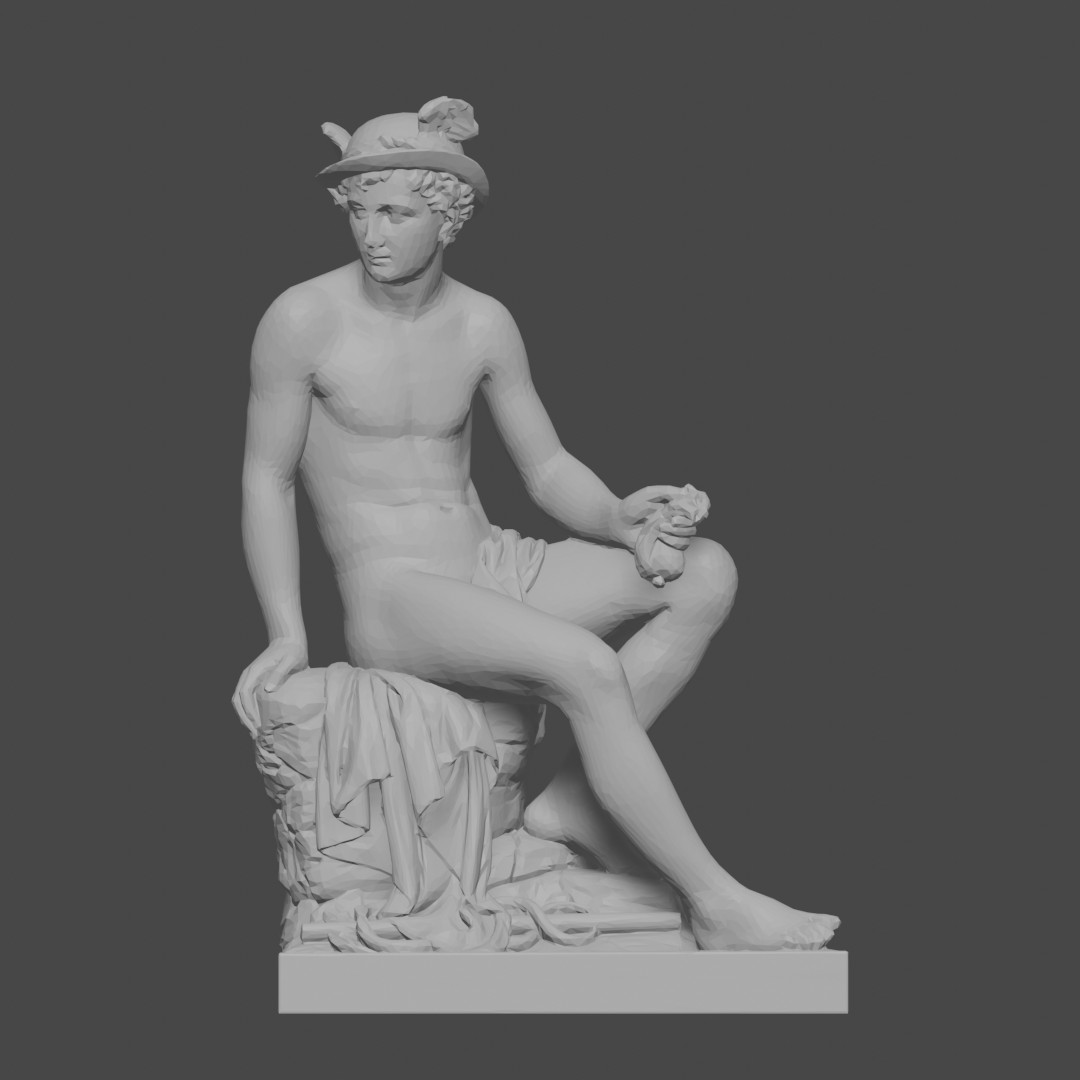}
	\includegraphics[trim=11mm 8mm 5mm 3mm,clip,width = 0.16\columnwidth]{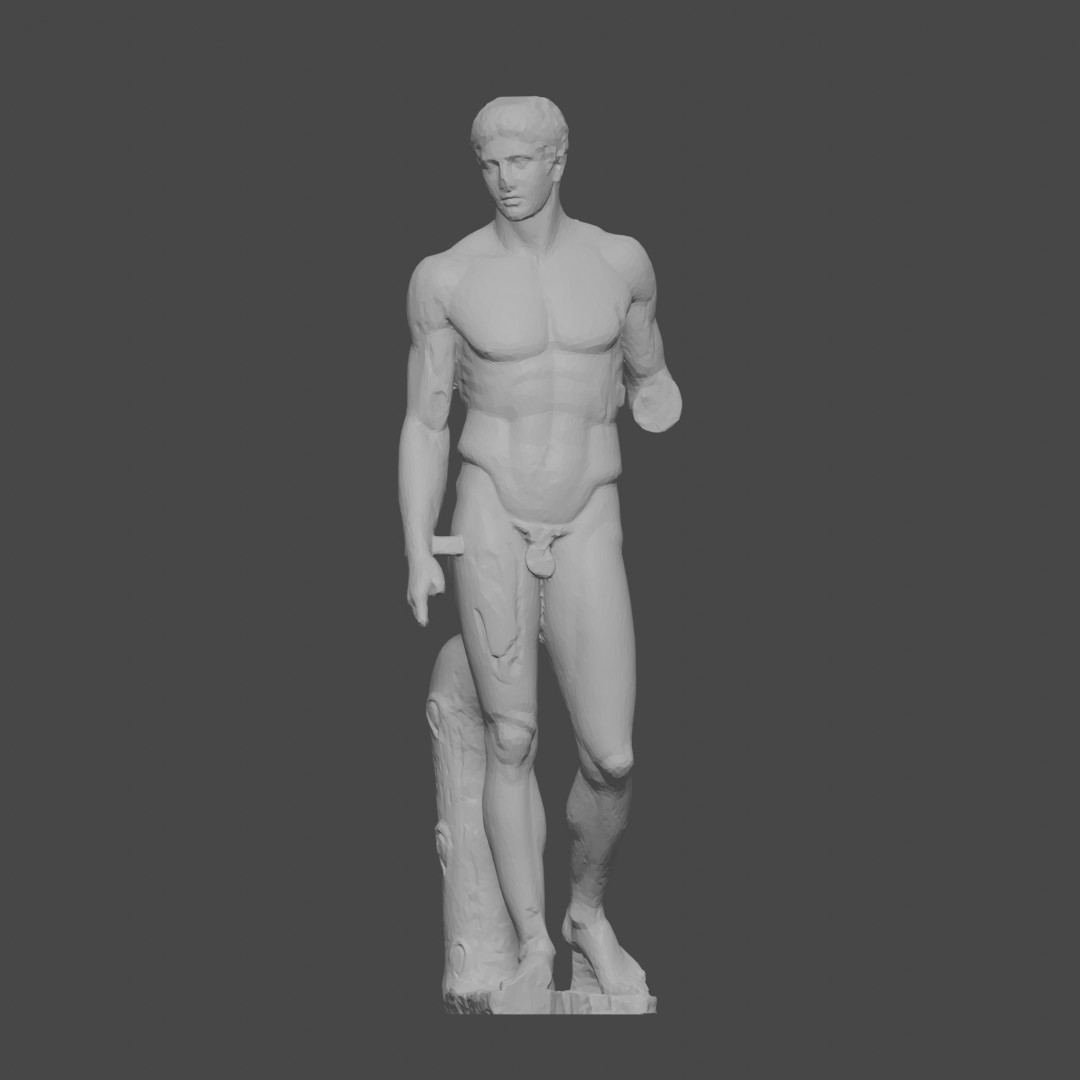}\\[1mm]
	\rotatebox{90}{\small~~~~~Ours\textcolor{white}p} 		
	\includegraphics[trim=11mm 8mm 5mm 3mm,clip,width = 0.16\columnwidth]{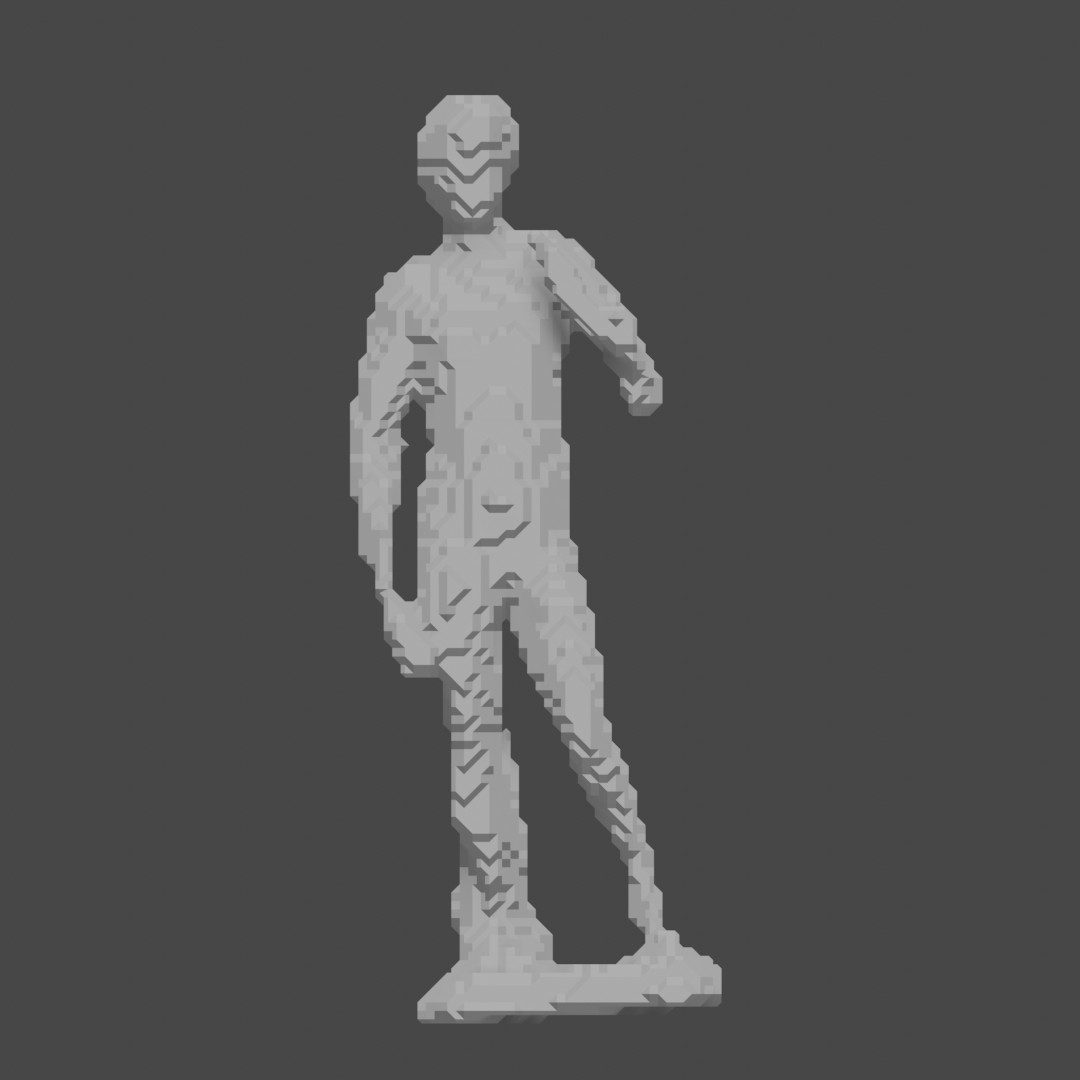}%
	\includegraphics[trim=11mm 8mm 5mm 3mm,clip,width = 0.16\columnwidth]{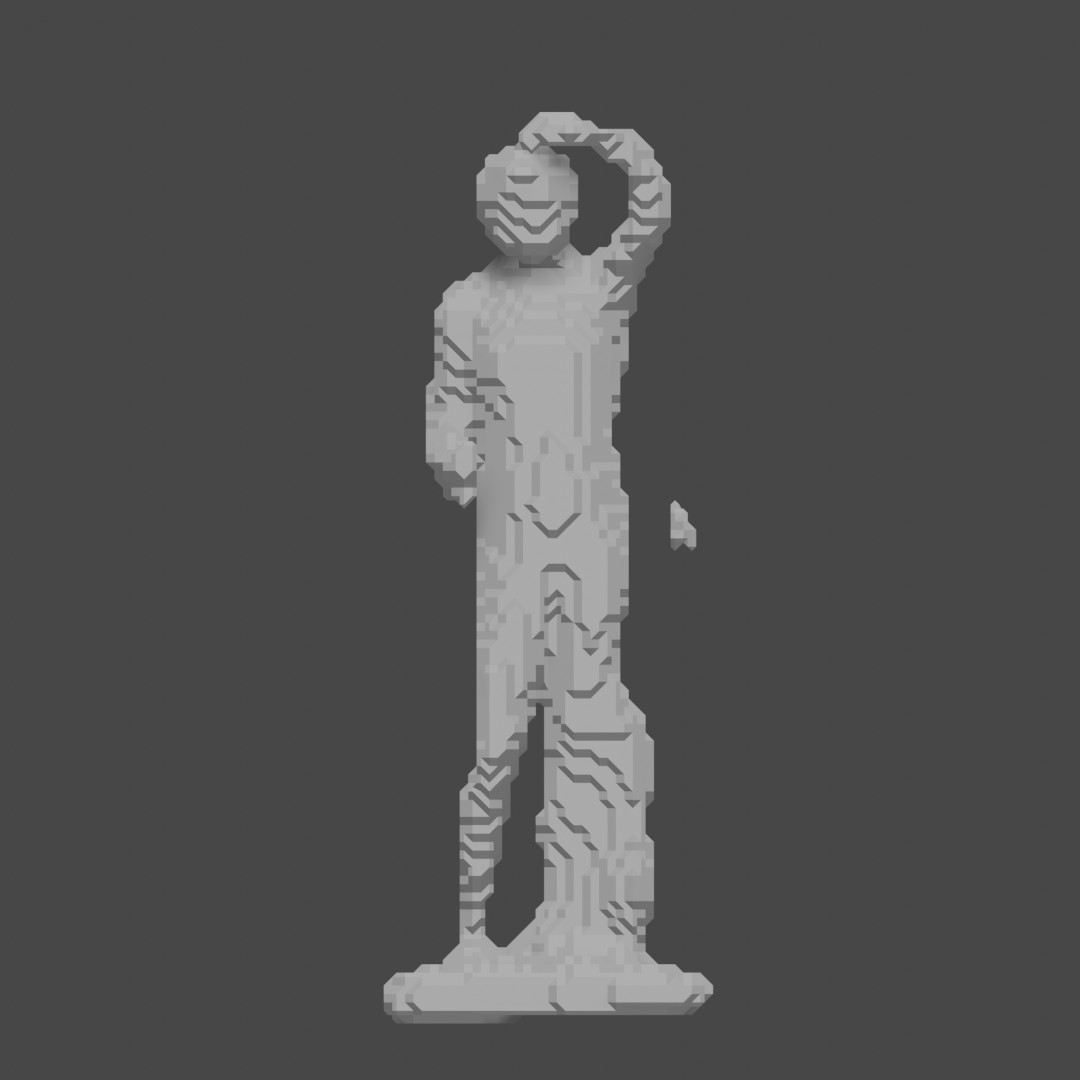}%
	\includegraphics[trim=11mm 8mm 5mm 3mm,clip,width = 0.16\columnwidth]{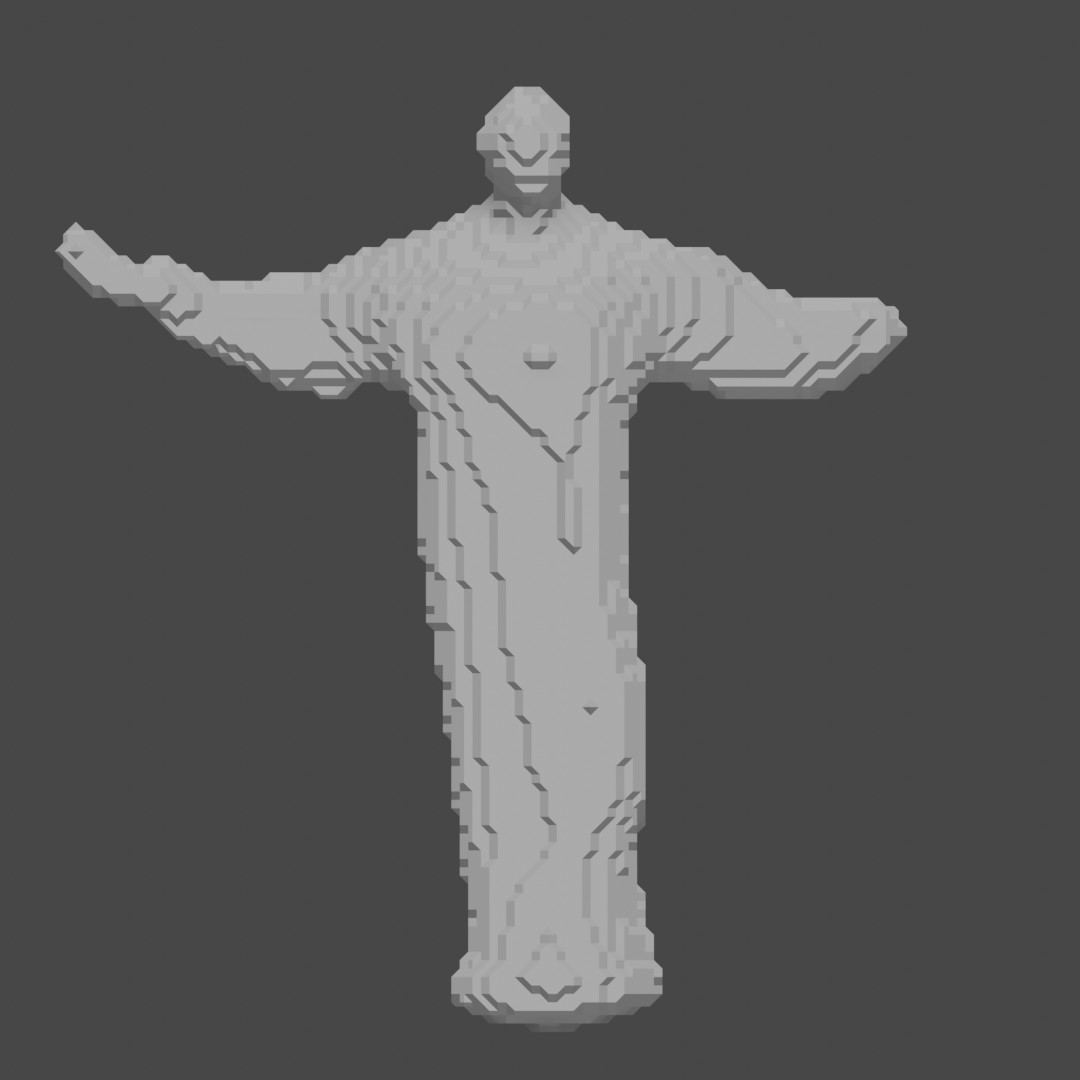}%
	\includegraphics[trim=11mm 8mm 5mm 3mm,clip,width = 0.16\columnwidth]{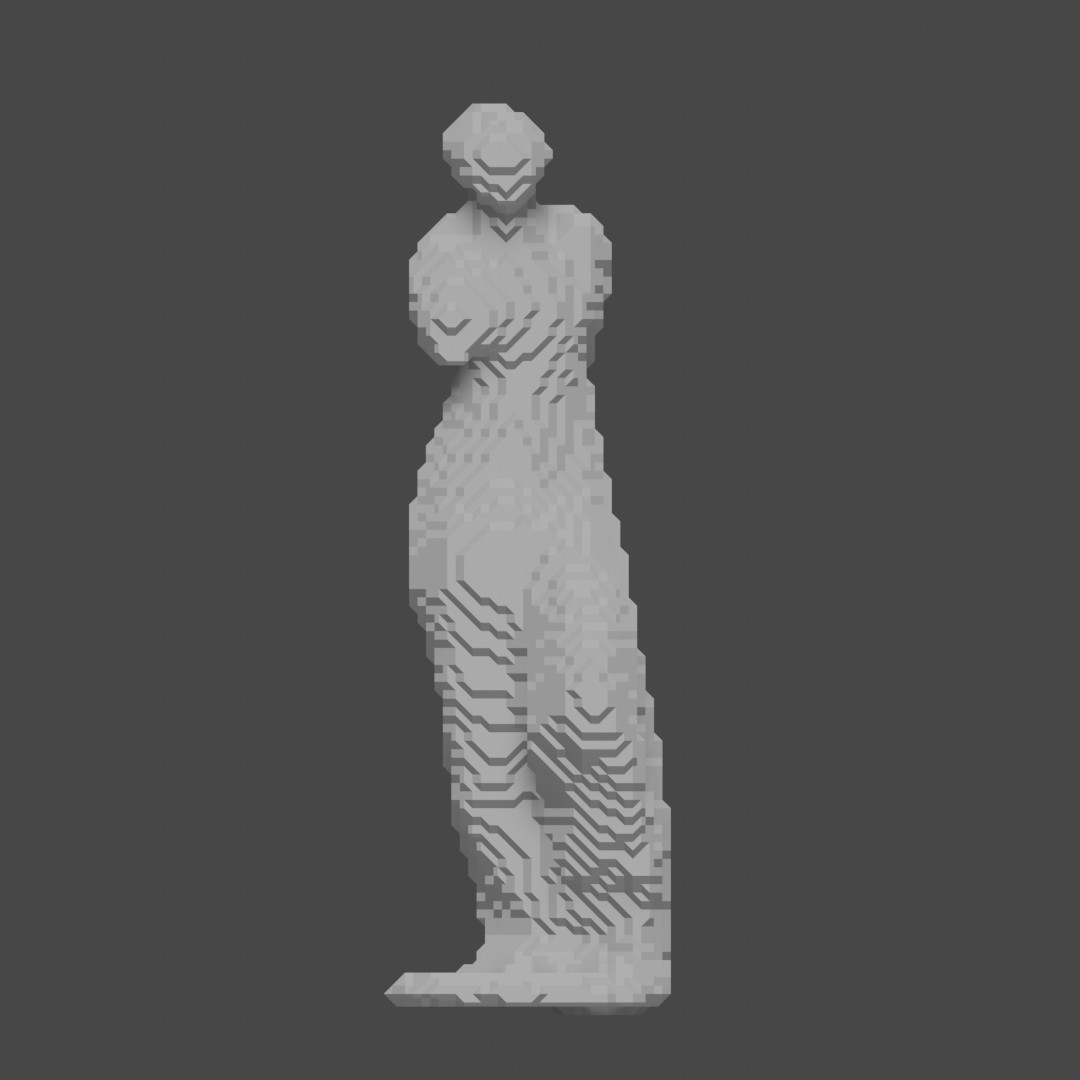}%
	\includegraphics[trim=11mm 8mm 5mm 3mm,clip,width = 0.16\columnwidth]{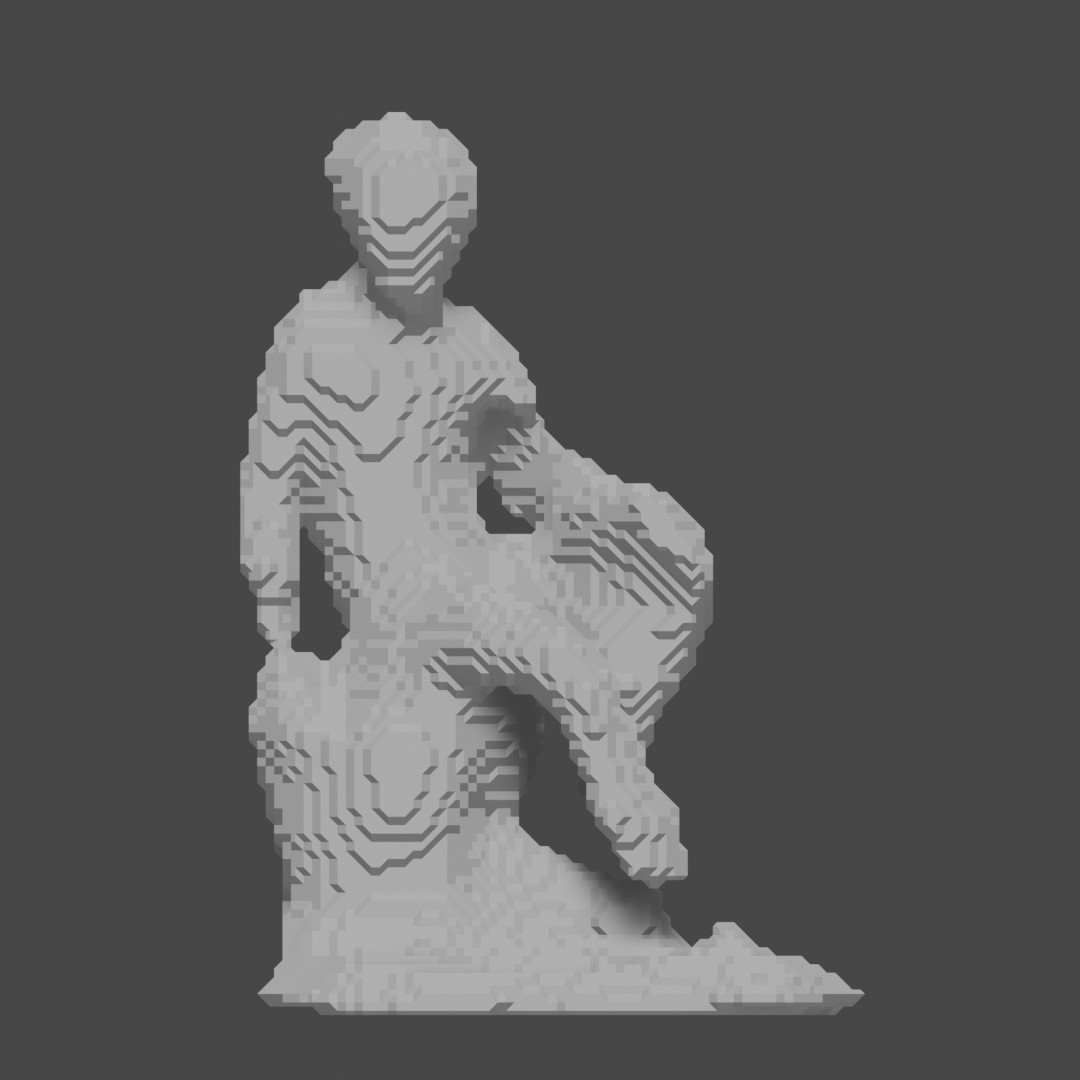}
	\includegraphics[trim=11mm 8mm 5mm 3mm,clip,width = 0.16\columnwidth]{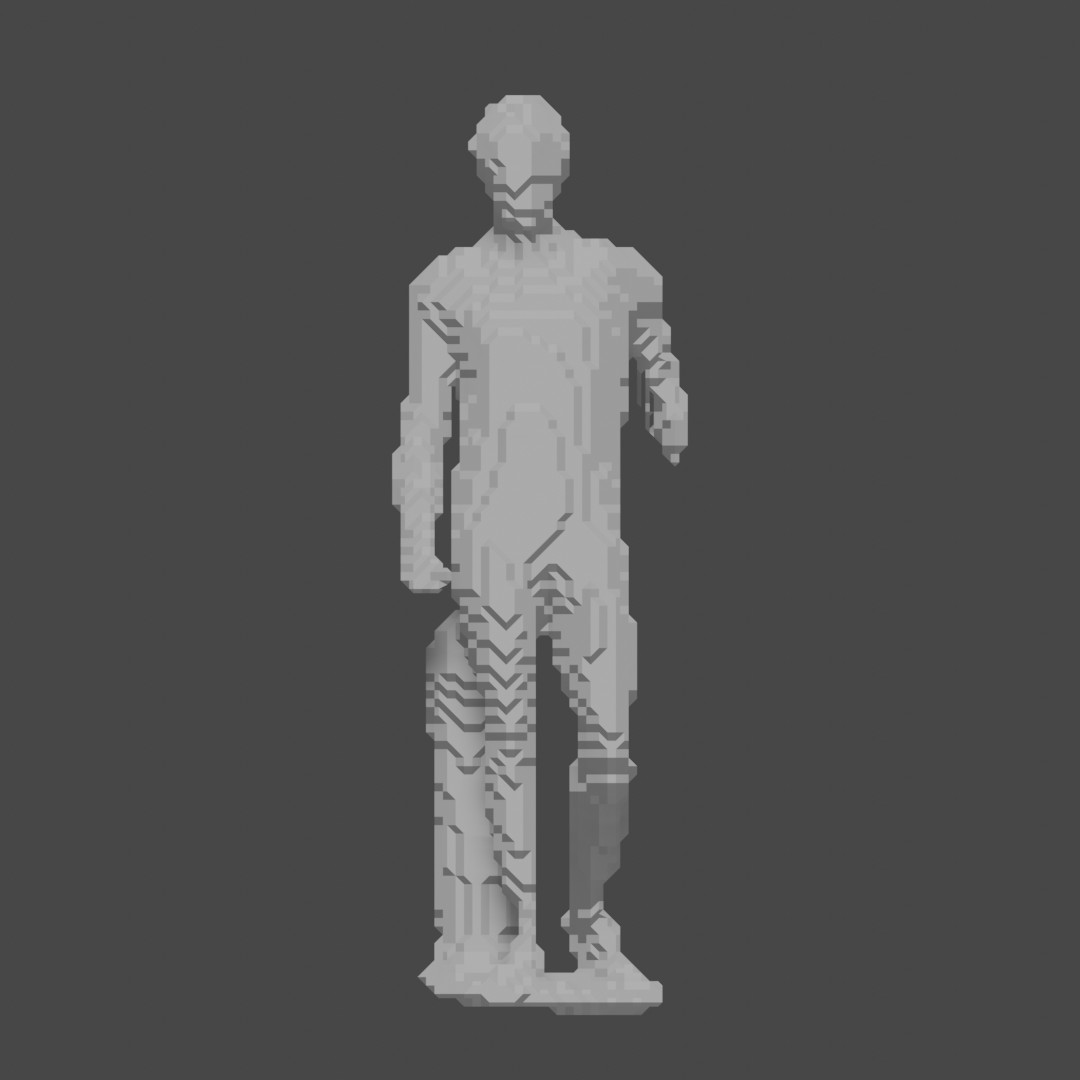}\\[1mm]
	%
	\vspace{-2mm}
	\caption{Reconstruction on statues scenes. Our representation can be used to recover complex structural features directly from time-of-flight measurements of moderate resolution. The scanned area on the wall is 2\,m$\times$2\,m large, sampled with an array of $32\times 32$ positions at 64\,ps resolution resembling a low-res capture of the one by \cite{lindell2019wave}.}
	\label{fig:results_1_statues}
\end{figure*}

We conducted a similar experiment for reconstructing objects of higher structural complexity. In this case, we created a class of ~500 meshes containing statues and sculptures taken from the ``Scan the World'' collection. This is a very challenging class as it contains a wide variety of body poses, shape diversity, thin features and detail. We sampled multiple configurations of the source meshes and trained our models on 40,000 scenes. Figure \ref{fig:results_1_statues} shows how our trained model correctly predicts global features of the shape (pose, body) and thin features (arms, legs).

\begin{table}[htbp]
	\centering
	\makebox[\textwidth][c]{
		\begin{tabular}{cc|ccc|ccc}
			\toprule
			&       & \multicolumn{3}{c|}{Bikes+Letters+Guns} & \multicolumn{3}{c}{Statues+Sculptures} \\
			\midrule
			Model 			& &~~~ F-Score  &  ~ IoU 	& &~~~ F-Score  & ~ IoU 	 \\ \hline
			Conv-OccNet   	& & ~~ 0.77  	&  ~ 0.62	& & ~~ 0.77 	& ~ 0.63	 \\
			CBN-Cell   		& & ~~ 0.81  	&  ~ 0.68	& & ~~ 0.77 	& ~ 0.63 	  \\
			\bottomrule
		\end{tabular}%
	}
	\vspace{1mm}
	\caption{Scene reconstruction scores over two validation partitions.}
	\label{tab:bikegunletter_partition}%
\end{table}%

We remark to the reader that all of our targets shown here correspond to completely diffuse surface cases.

\subsection{3D Reconstruction Beyond Fermat Limits and Self-Occlusion}
We proceed to illustrate two interesting properties of our representation: the recovery of non-Fermat features and self-occlusion capability. In both experiments we trained the convolutional architecture by \cite{peng2020convolutional} as it showed much faster convergence during preliminary tests.

\begin{figure}[t]
	\begin{minipage}[c]{.13\columnwidth}
		\centering \small GT 
	\end{minipage}	
	\begin{minipage}[c]{.13\columnwidth}
		\centering \small BFC
	\end{minipage}
	\begin{minipage}[c]{.13\columnwidth} 
		\centering \small Ours~~~~
	\end{minipage}
	\vspace{0.1mm}	
	
	\includegraphics[trim=11mm 8mm 5mm 3mm,clip,width = 0.13\columnwidth]{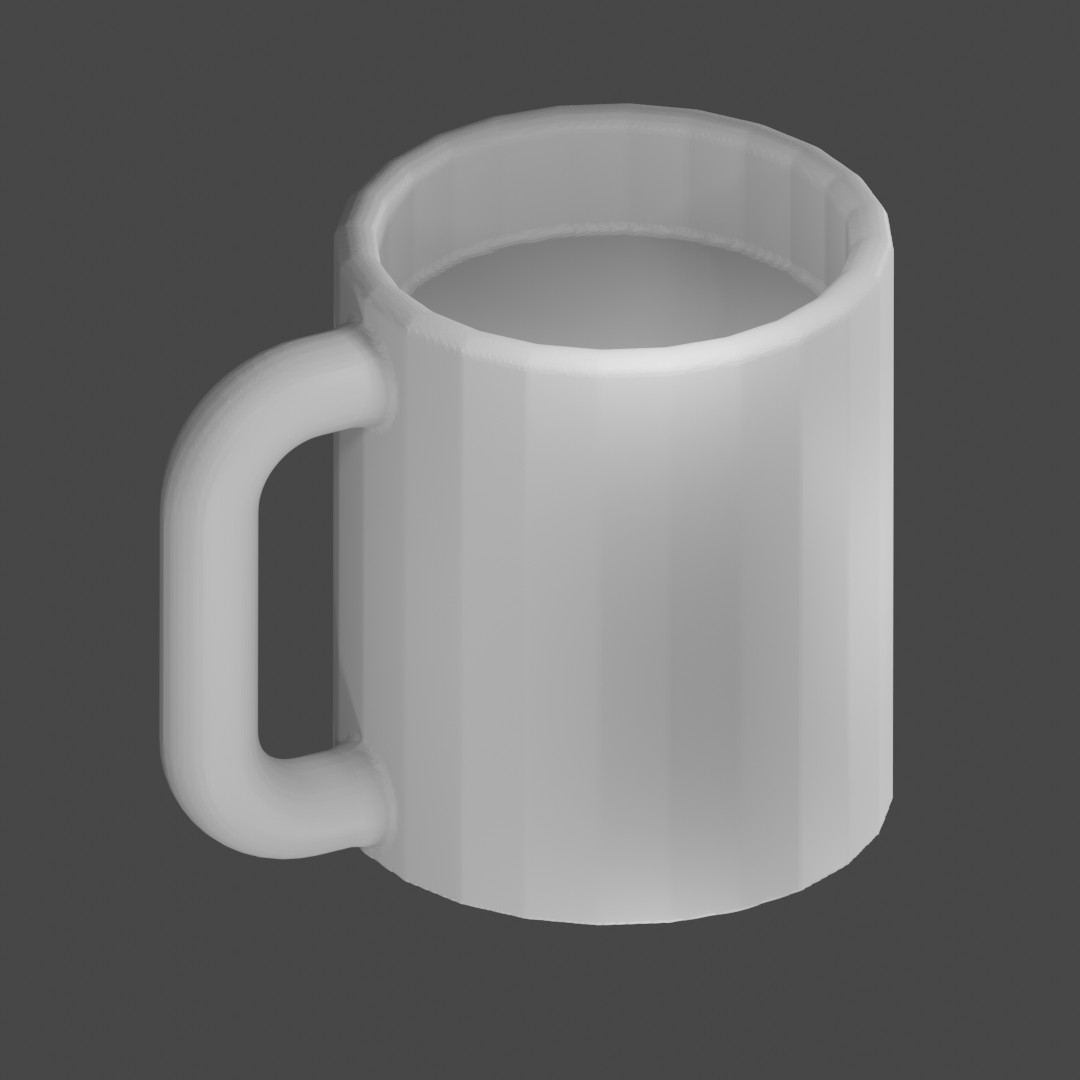}%
	\includegraphics[trim=11mm 8mm 5mm 3mm,clip,width = 0.13\columnwidth]{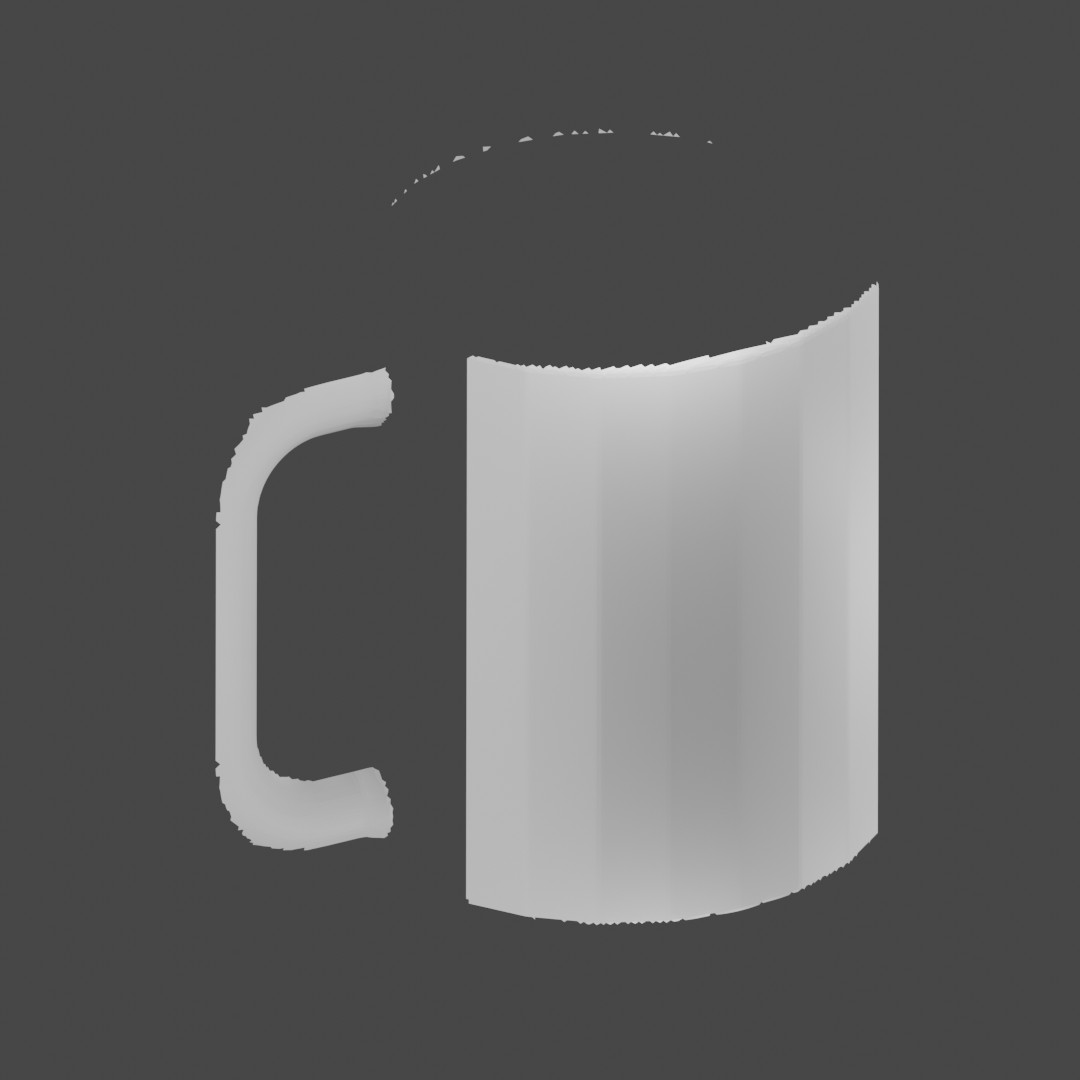}%
	\includegraphics[trim=11mm 8mm 5mm 3mm,clip,width = 0.13\columnwidth]{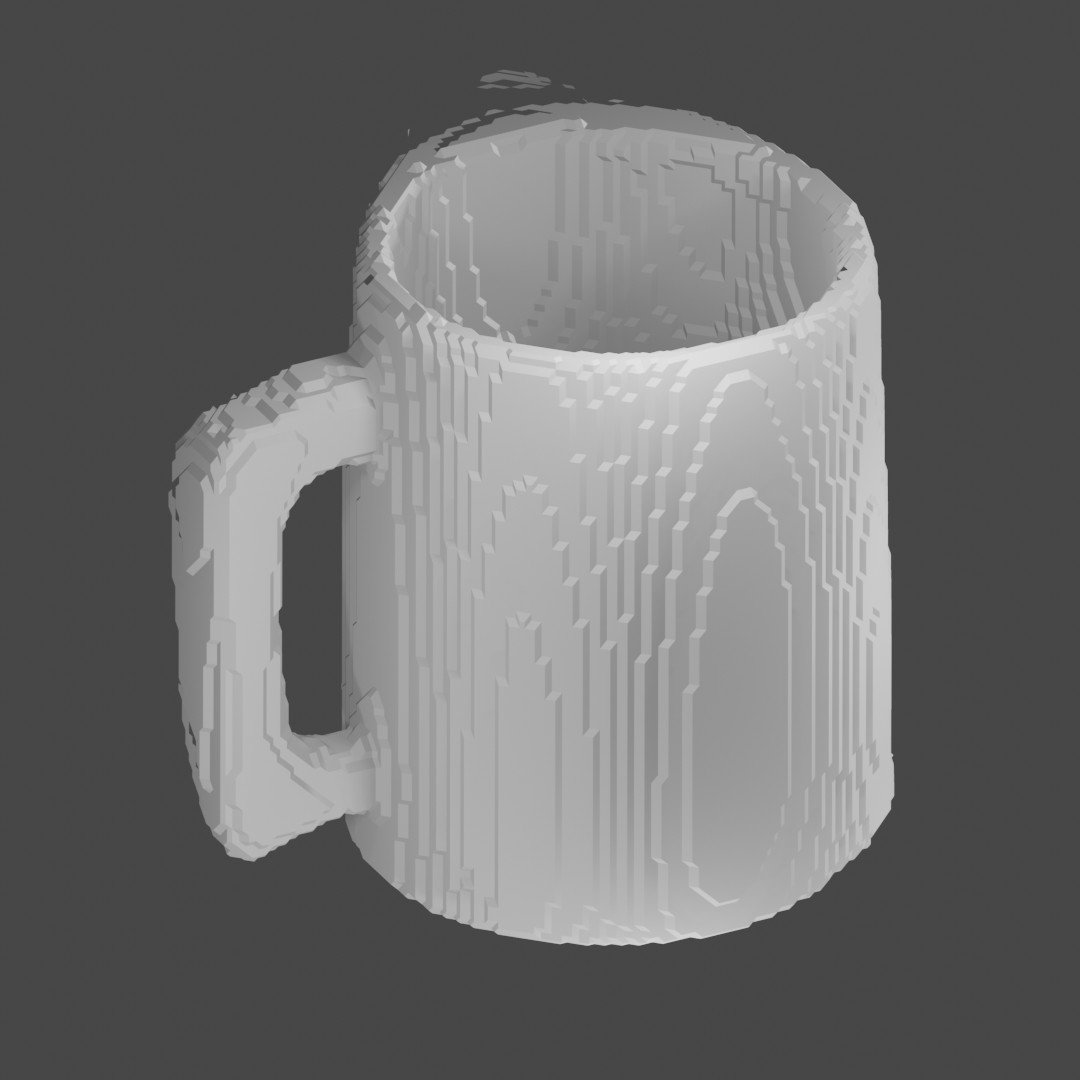}\hspace{6mm}
	\includegraphics[trim=11mm 8mm 5mm 3mm,clip,width = 0.13\columnwidth]{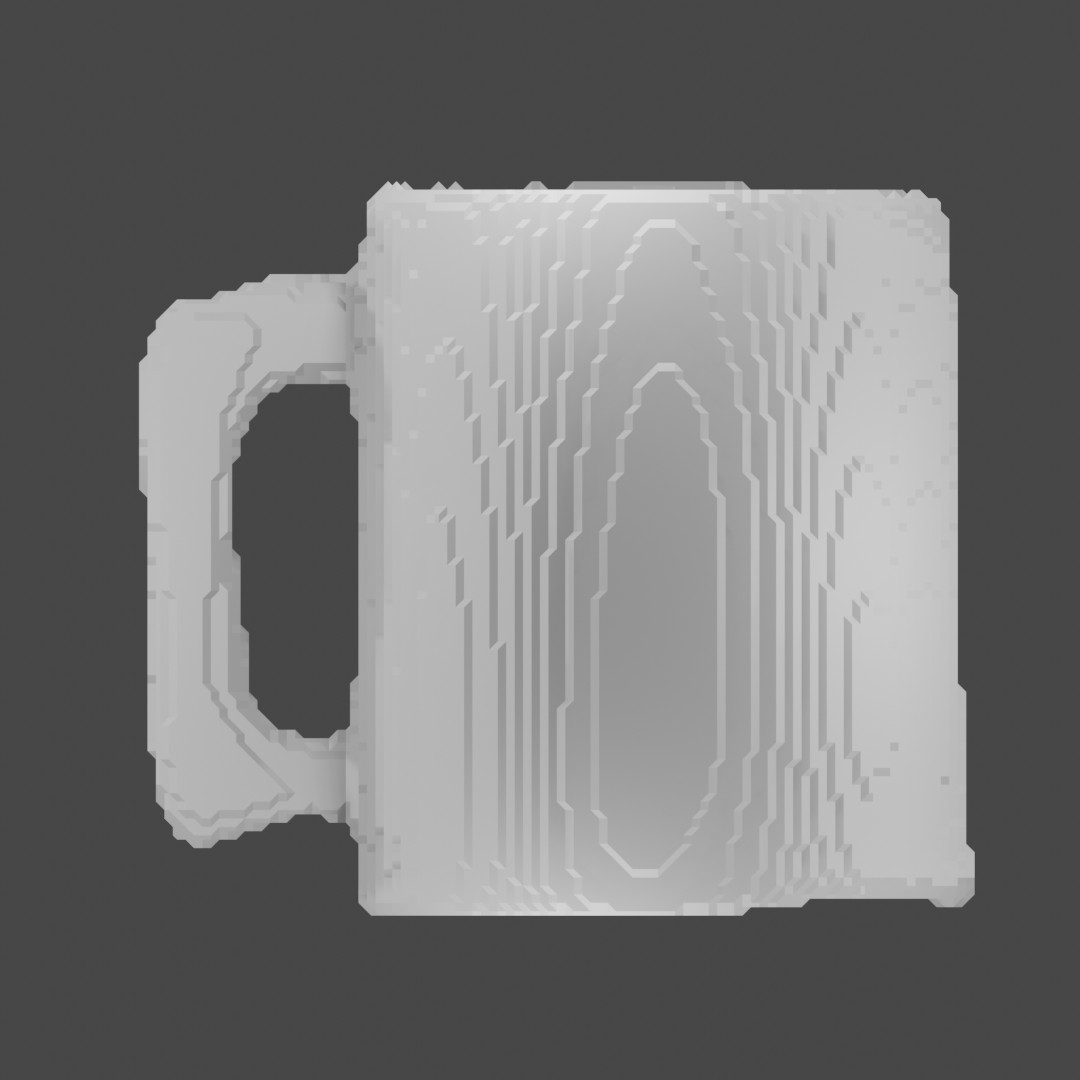}
	\includegraphics[trim=11mm 8mm 5mm 3mm,clip,width = 0.13\columnwidth]{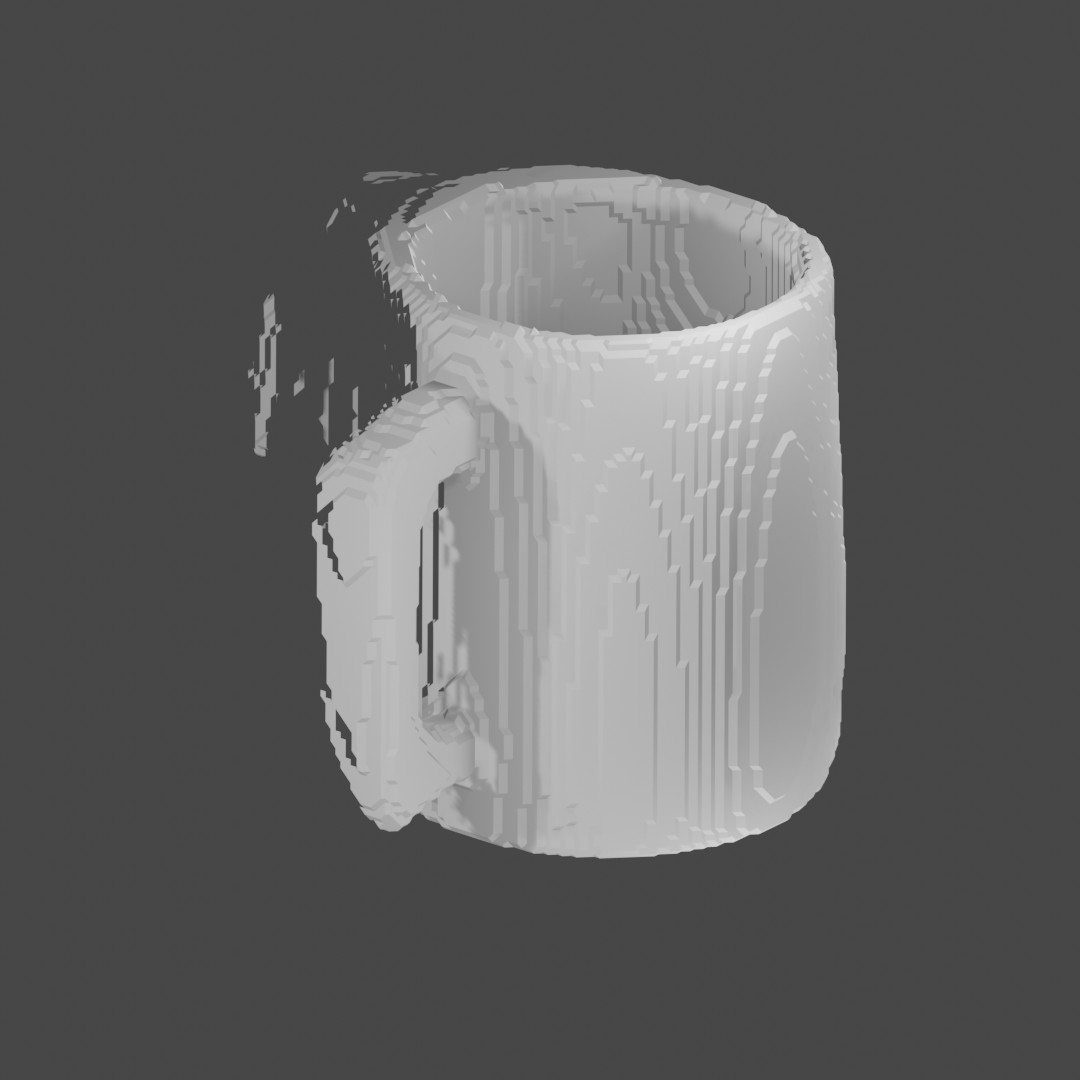} %
	\includegraphics[trim=11mm 8mm 5mm 3mm,clip,width = 0.13\columnwidth]{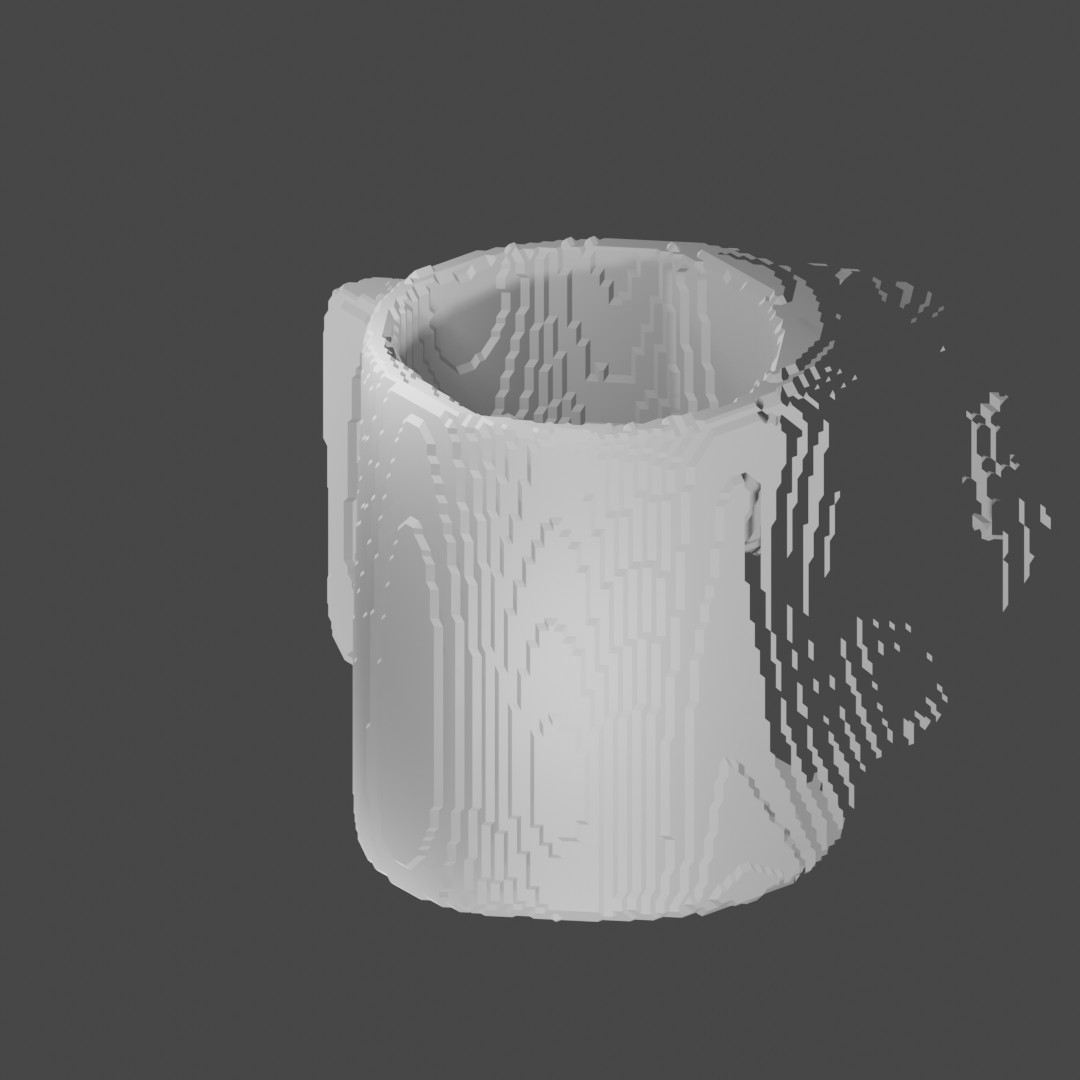} %
	\includegraphics[trim=11mm 8mm 5mm 3mm,clip,width = 0.13\columnwidth]{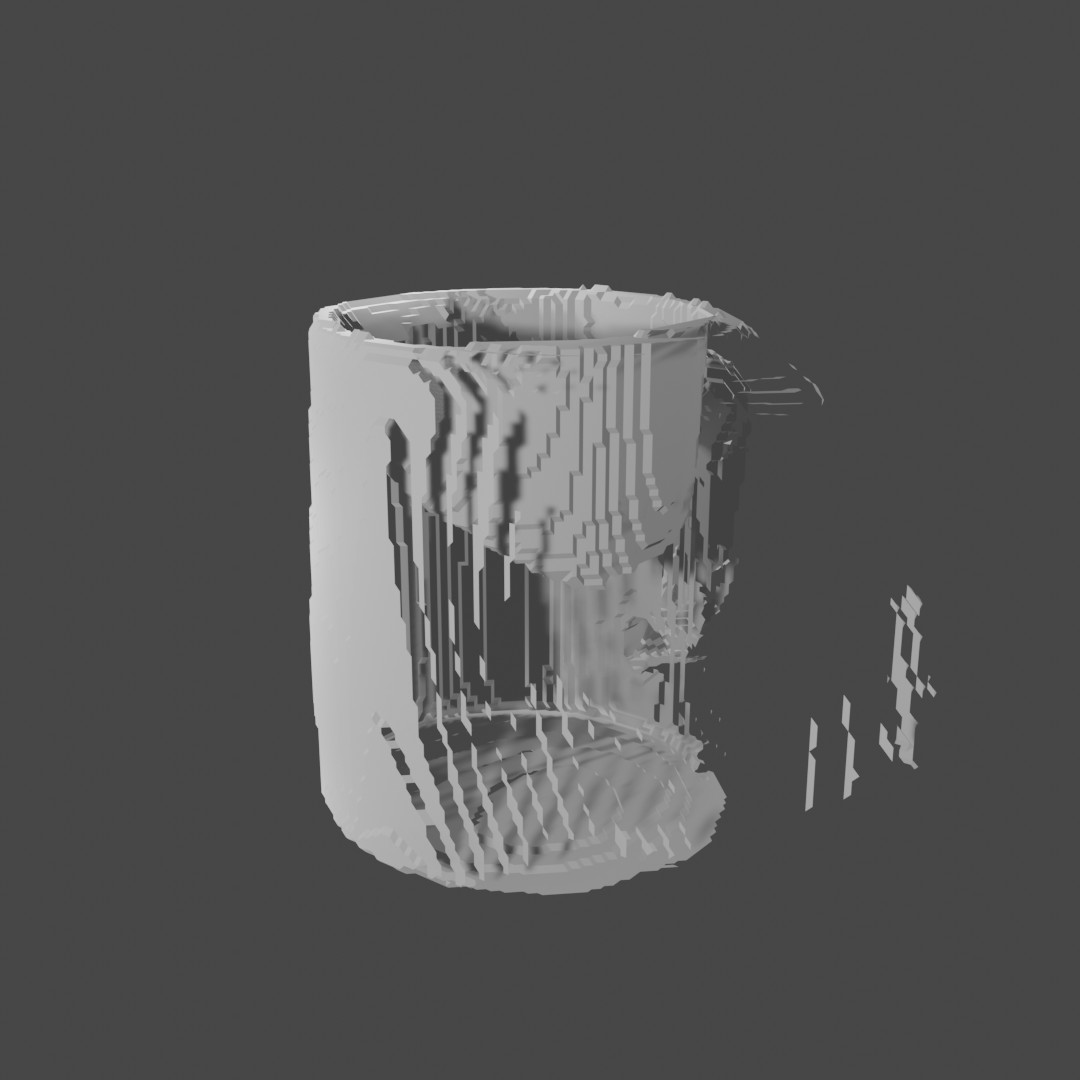}\!\!\!\\[0.2mm]%
	
	\includegraphics[trim=11mm 8mm 5mm 3mm,clip,width = 0.13\columnwidth]{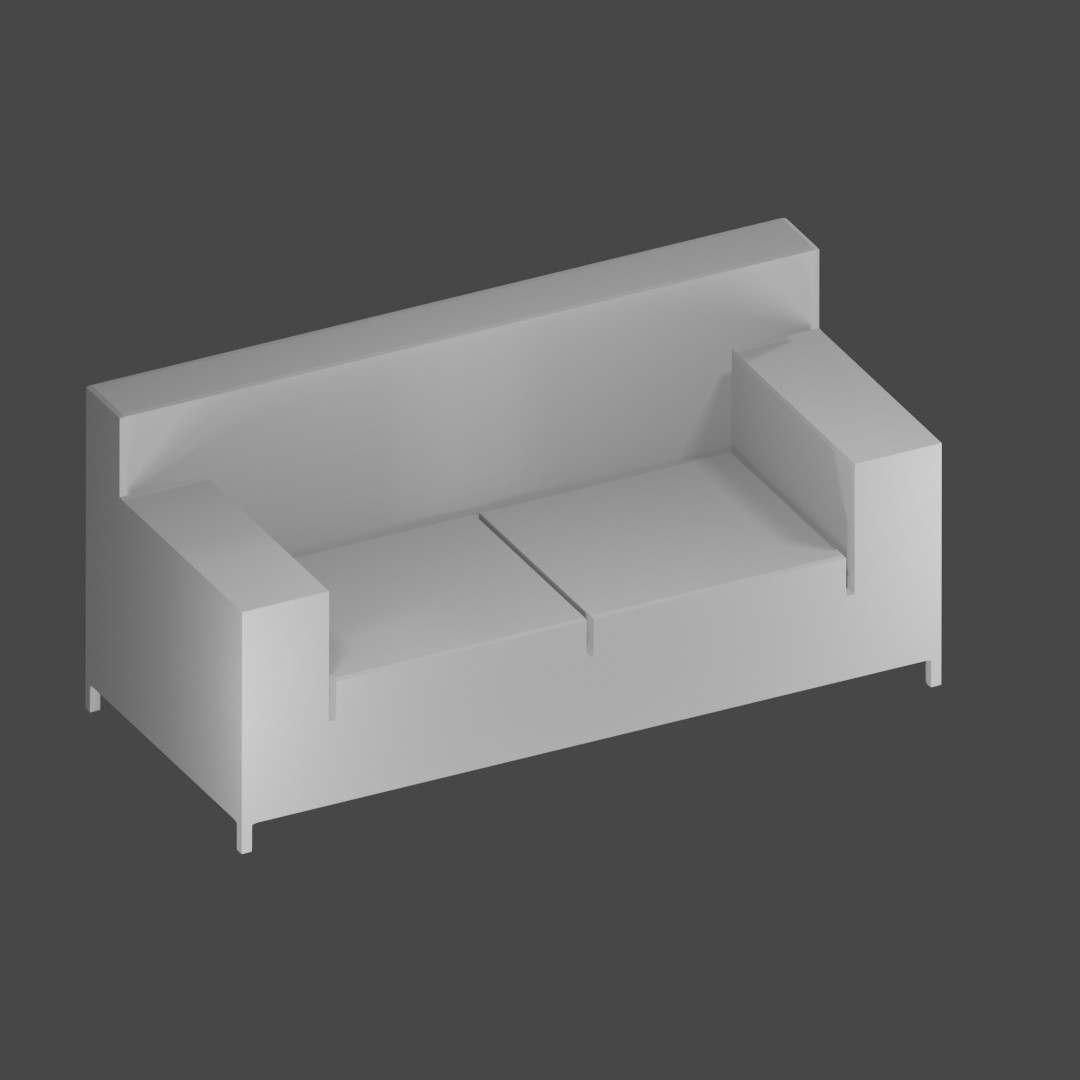}%
	\includegraphics[trim=11mm 8mm 5mm 3mm,clip,width = 0.13\columnwidth]{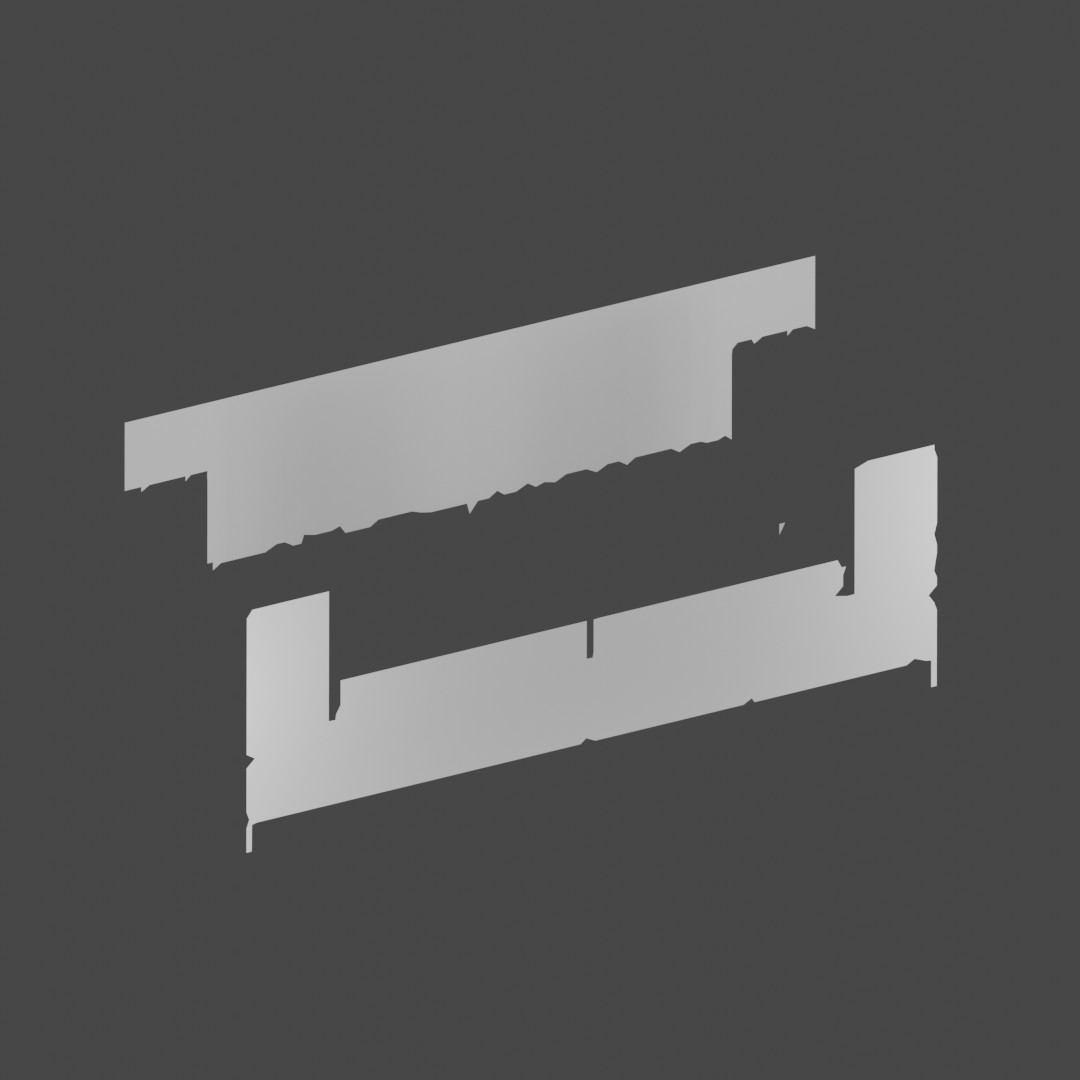}%
	\includegraphics[trim=11mm 8mm 5mm 3mm,clip,width = 0.13\columnwidth]{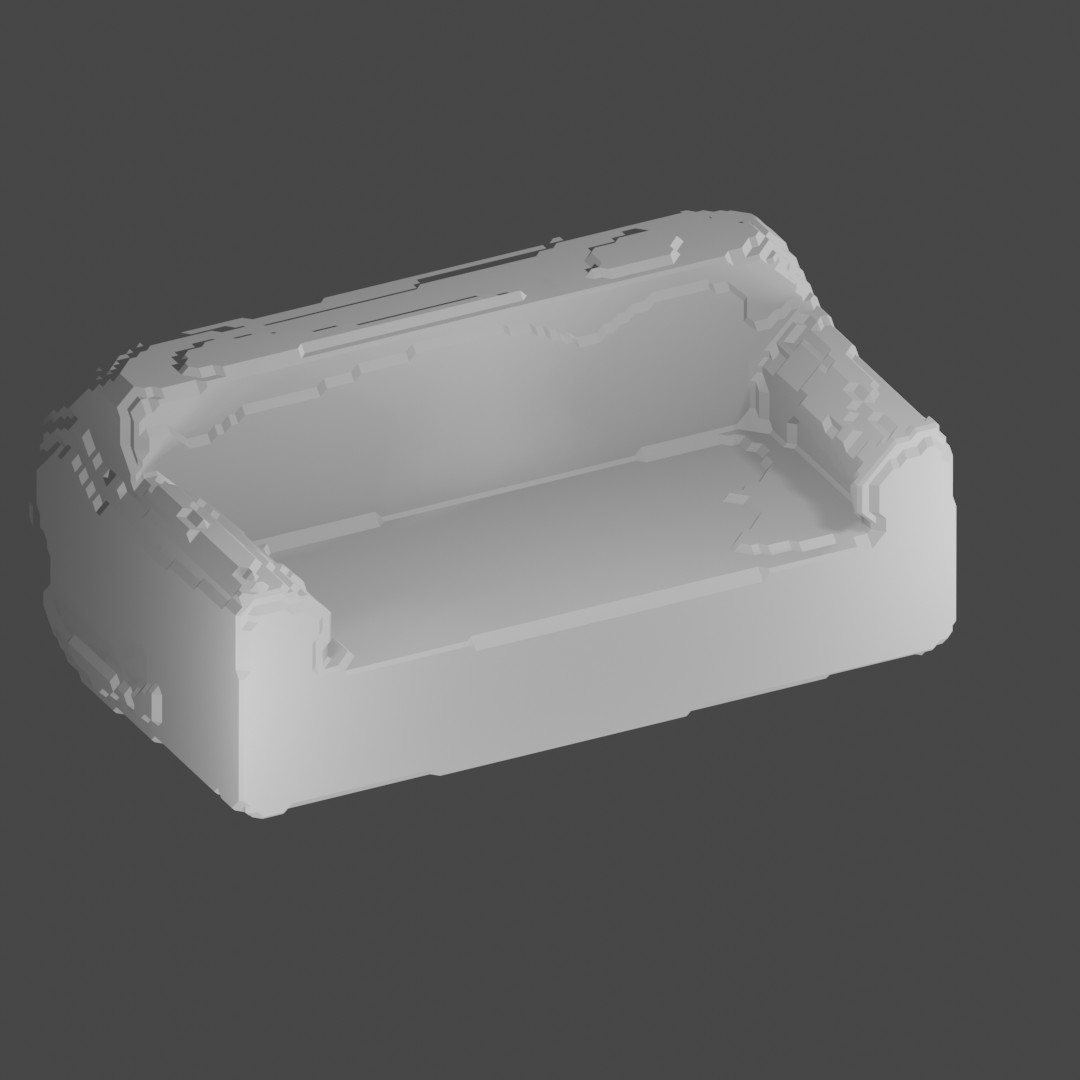}\hspace{6mm}
	\includegraphics[trim=11mm 8mm 5mm 3mm,clip,width = 0.13\columnwidth]{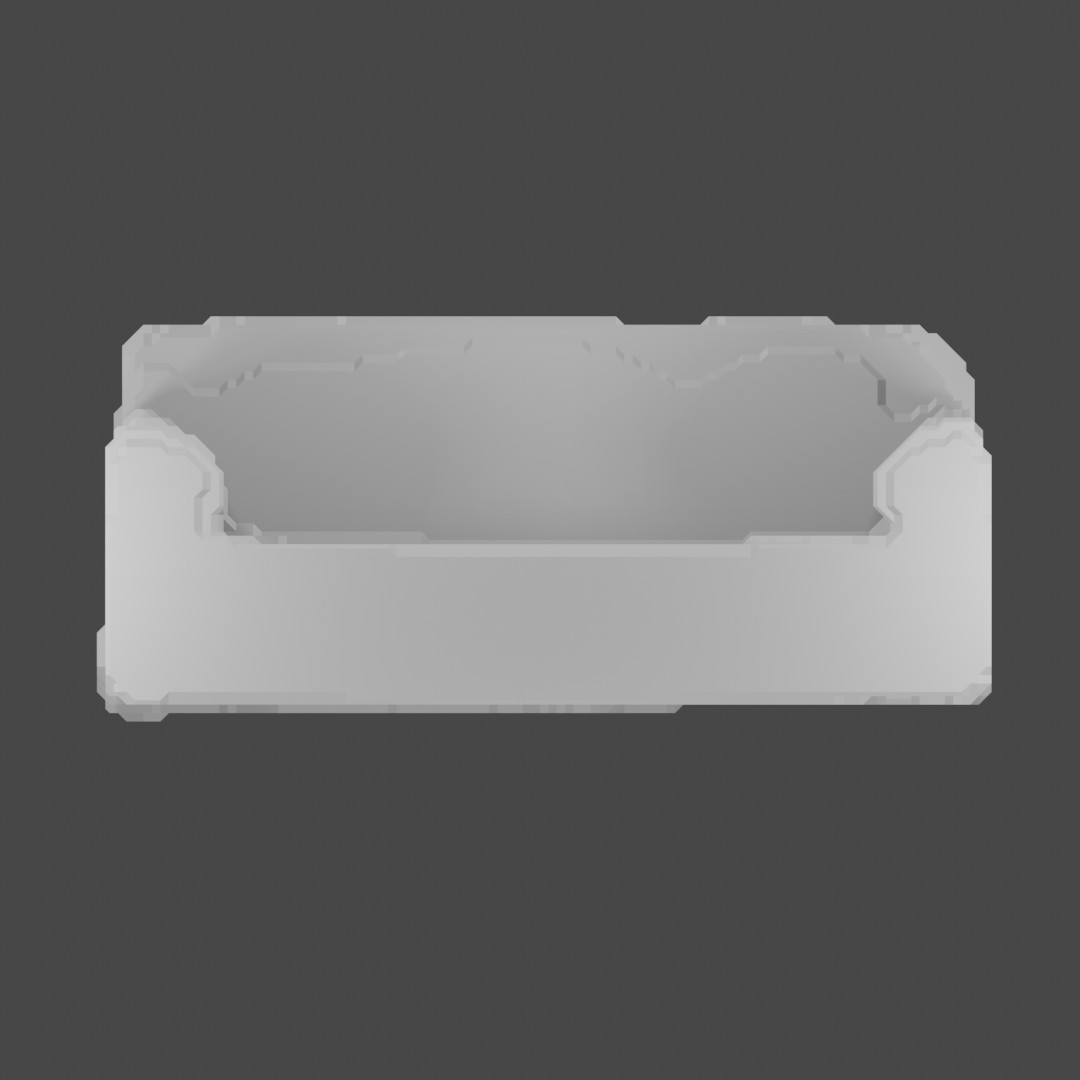} %
	\includegraphics[trim=11mm 8mm 5mm 3mm,clip,width = 0.13\columnwidth]{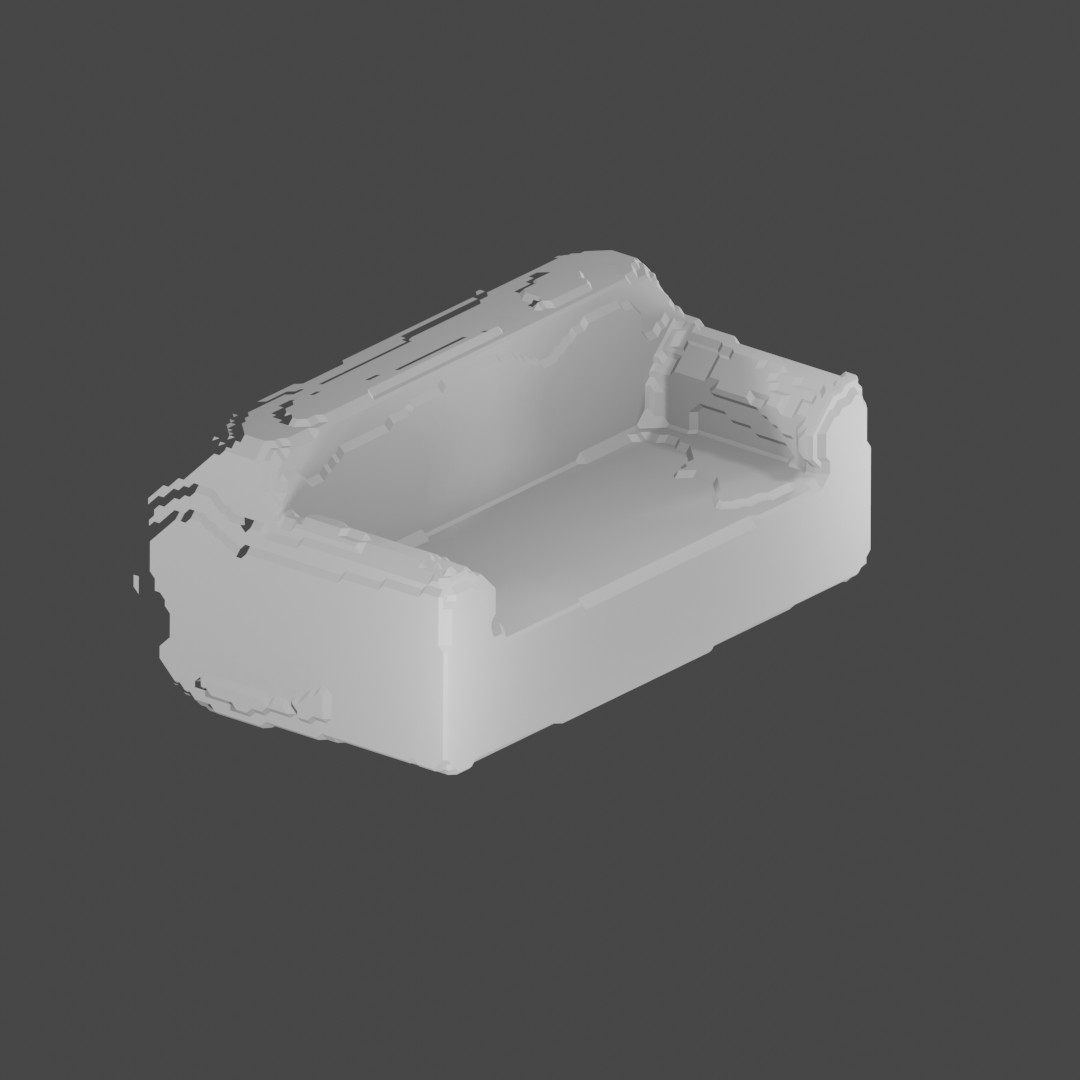} %
	\includegraphics[trim=11mm 8mm 5mm 3mm,clip,width = 0.13\columnwidth]{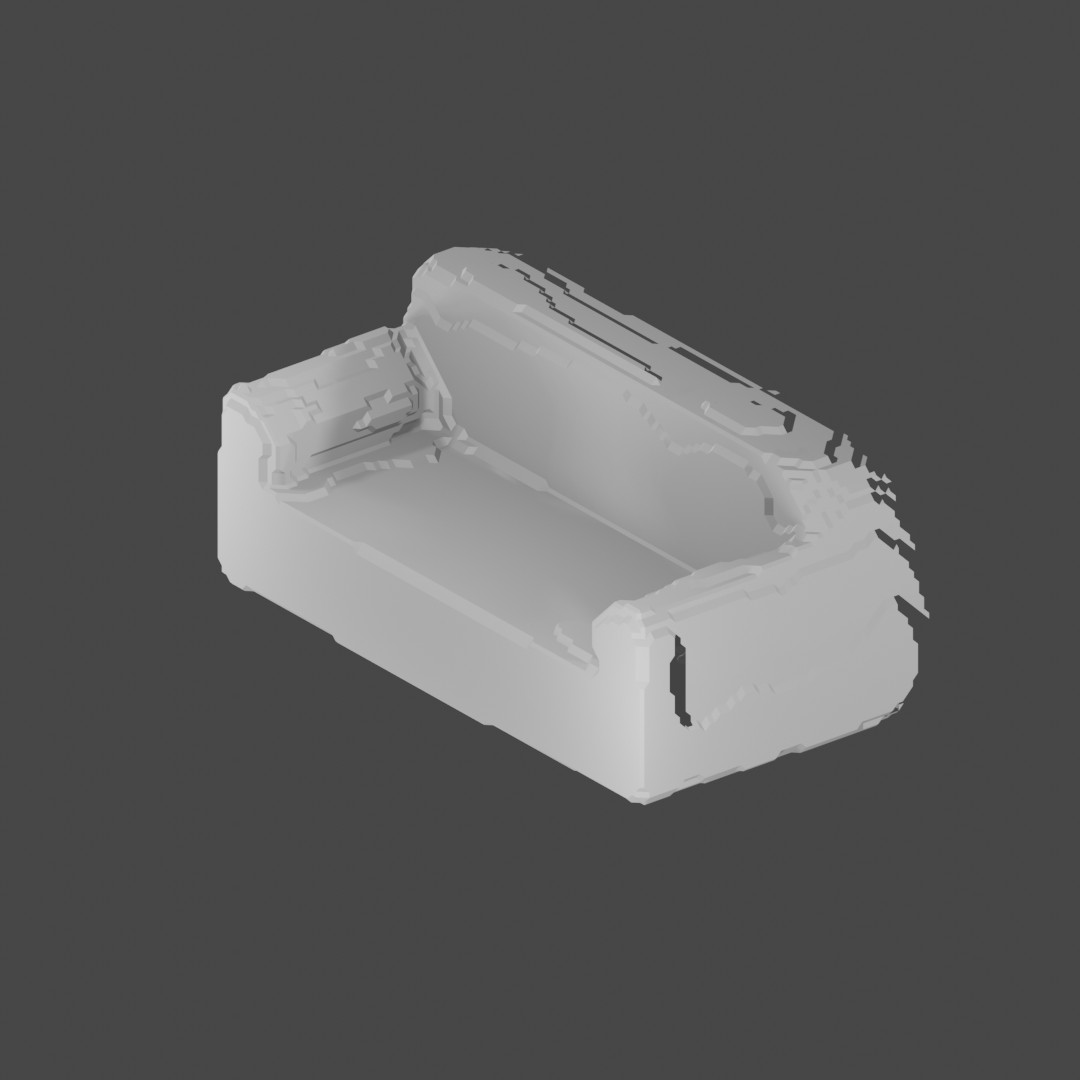} %
	\includegraphics[trim=11mm 8mm 5mm 3mm,clip,width = 0.13\columnwidth]{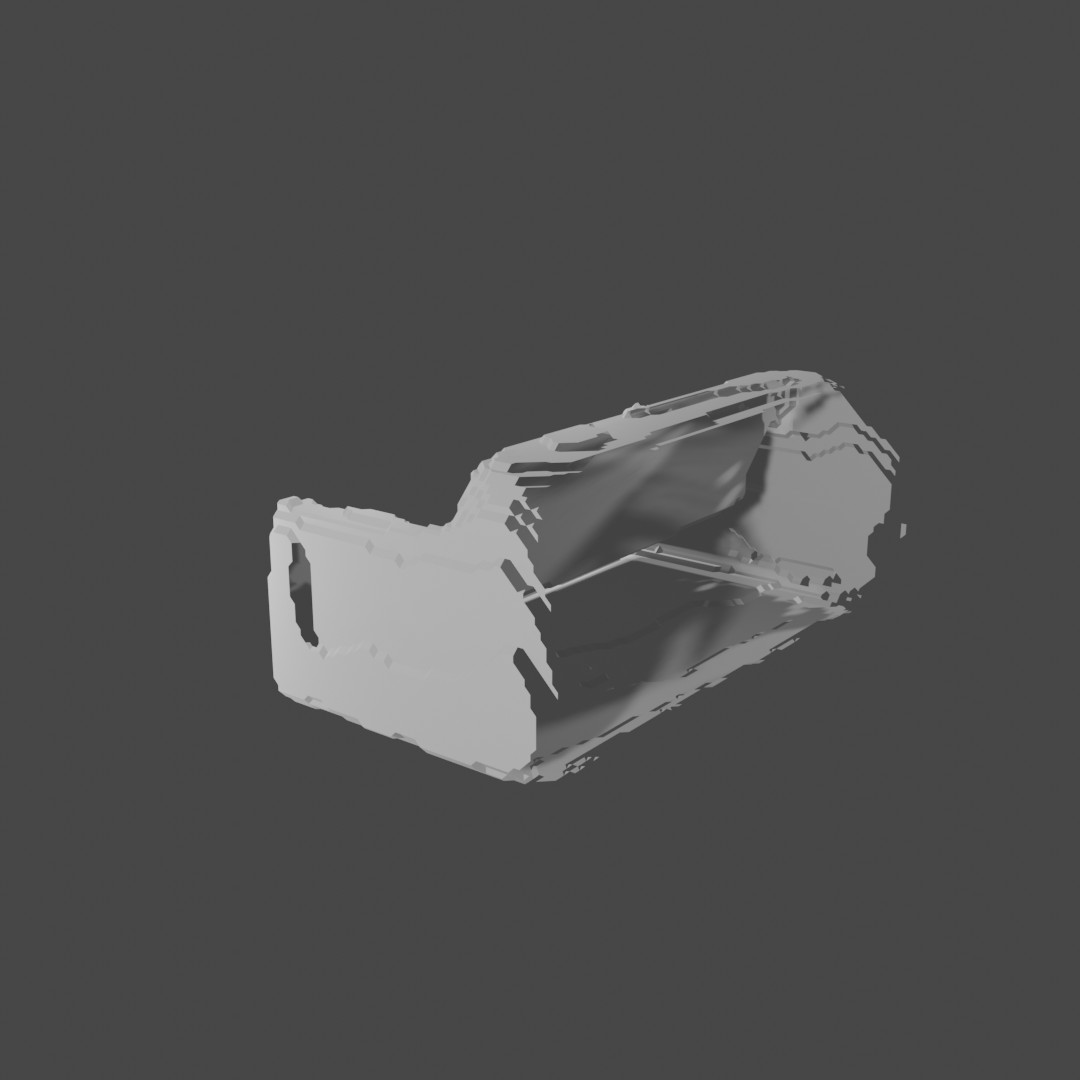}\!\!\!\\[0.2mm]%
	
	\includegraphics[trim=11mm 8mm 5mm 3mm,clip,width = 0.13\columnwidth]{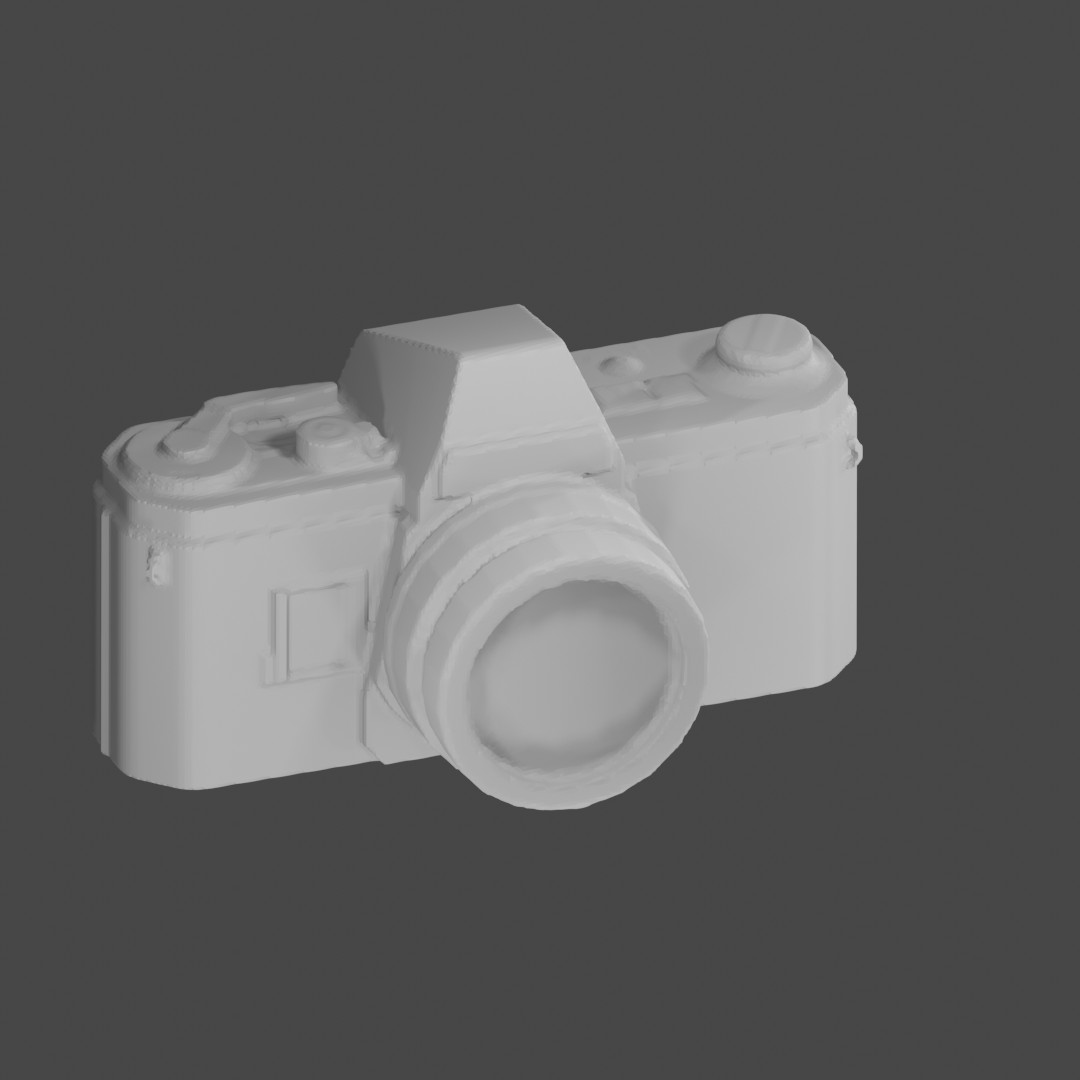}%
	\includegraphics[trim=11mm 8mm 5mm 3mm,clip,width = 0.13\columnwidth]{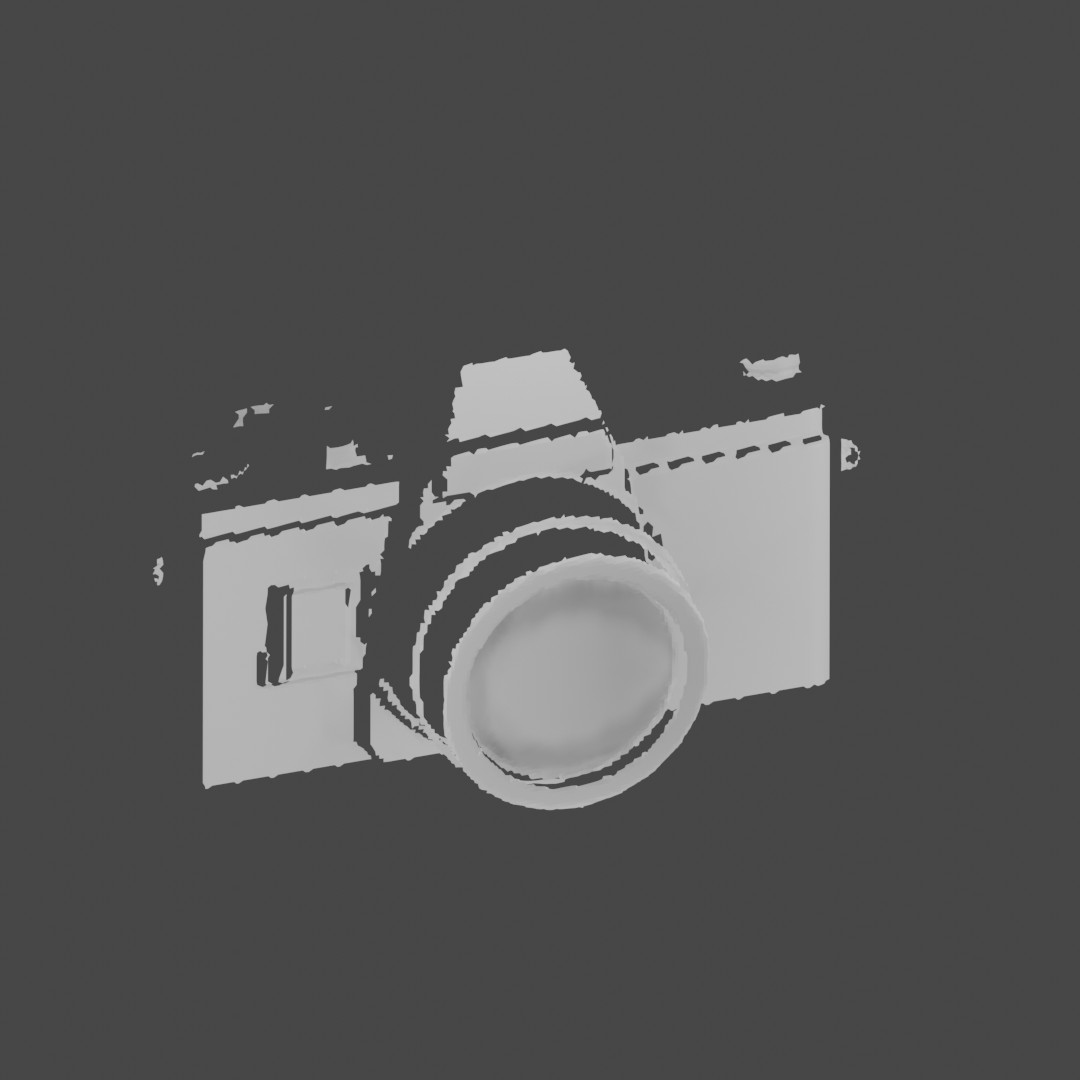}%
	\includegraphics[trim=11mm 8mm 5mm 3mm,clip,width = 0.13\columnwidth]{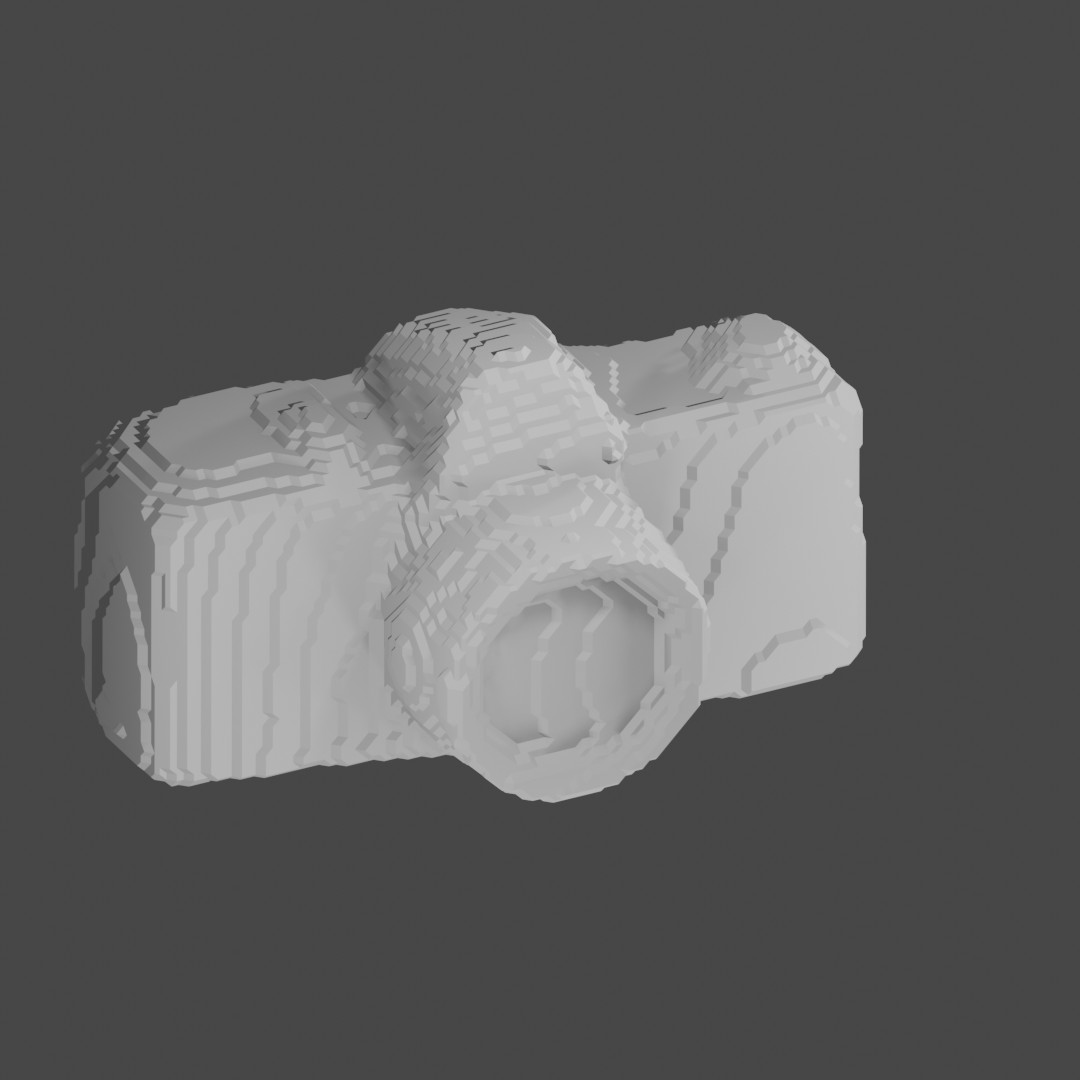}\hspace{6mm}
	\includegraphics[trim=11mm 8mm 5mm 3mm,clip,width = 0.13\columnwidth]{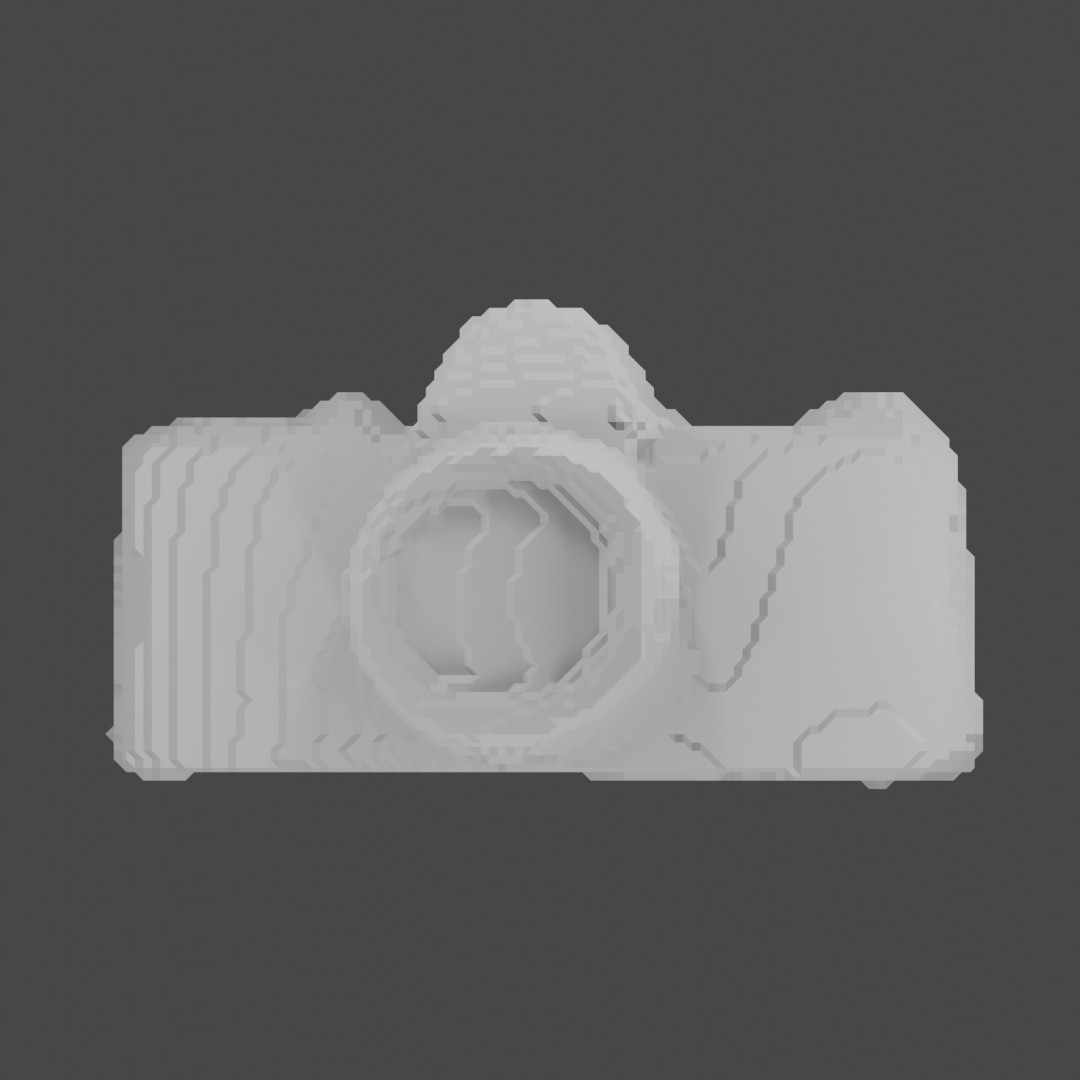} %
	\includegraphics[trim=11mm 8mm 5mm 3mm,clip,width = 0.13\columnwidth]{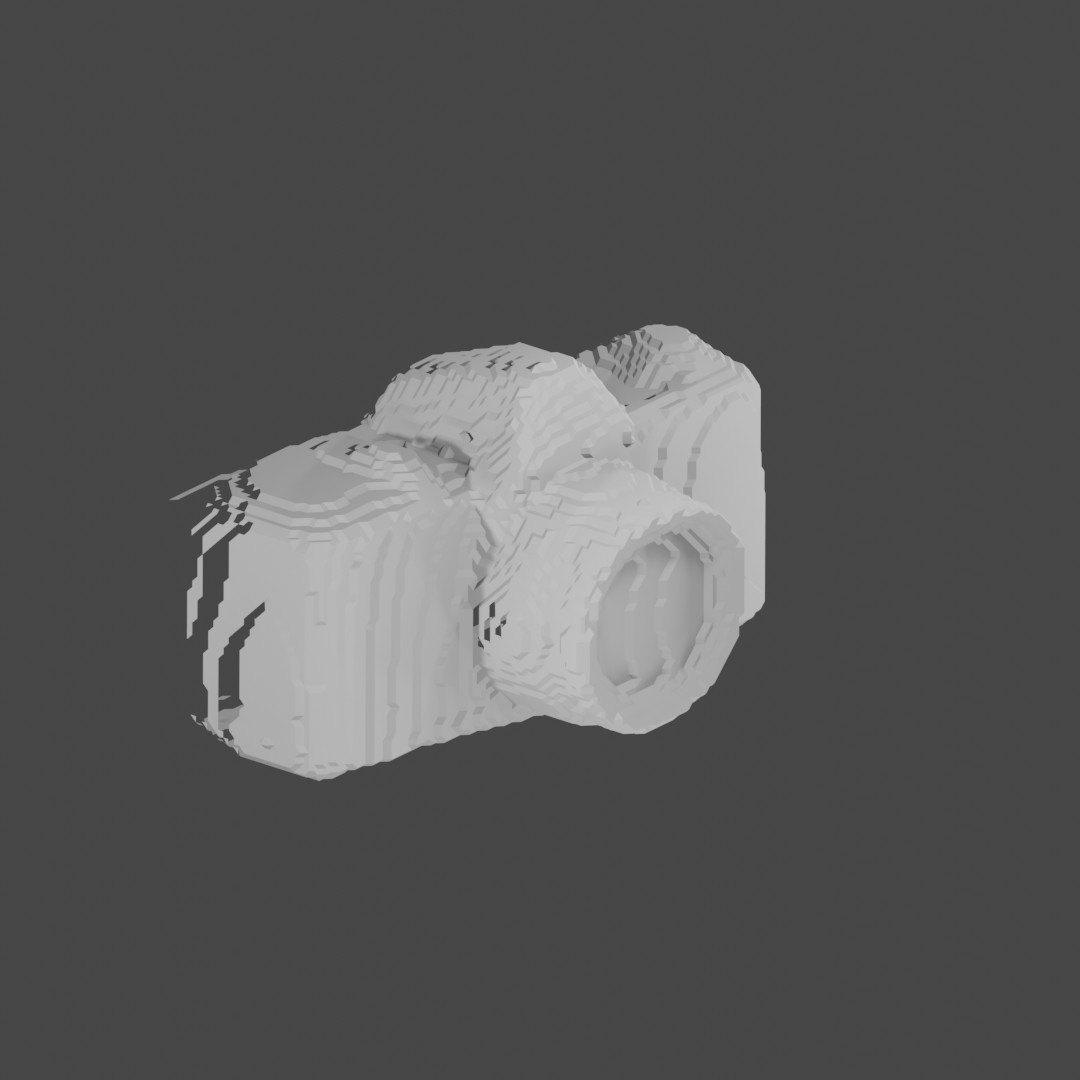} %
	\includegraphics[trim=11mm 8mm 5mm 3mm,clip,width = 0.13\columnwidth]{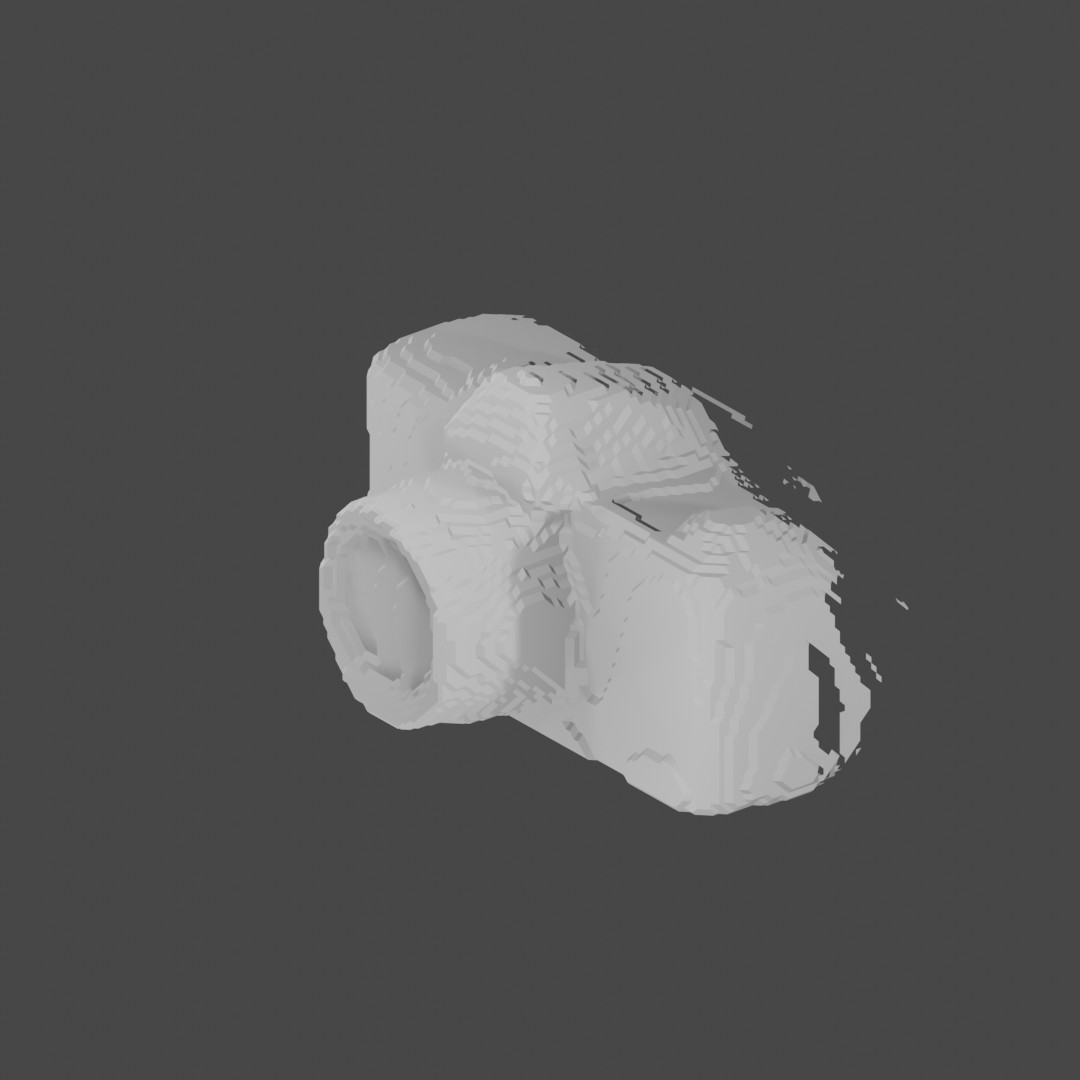} %
	\includegraphics[trim=11mm 8mm 5mm 3mm,clip,width = 0.13\columnwidth]{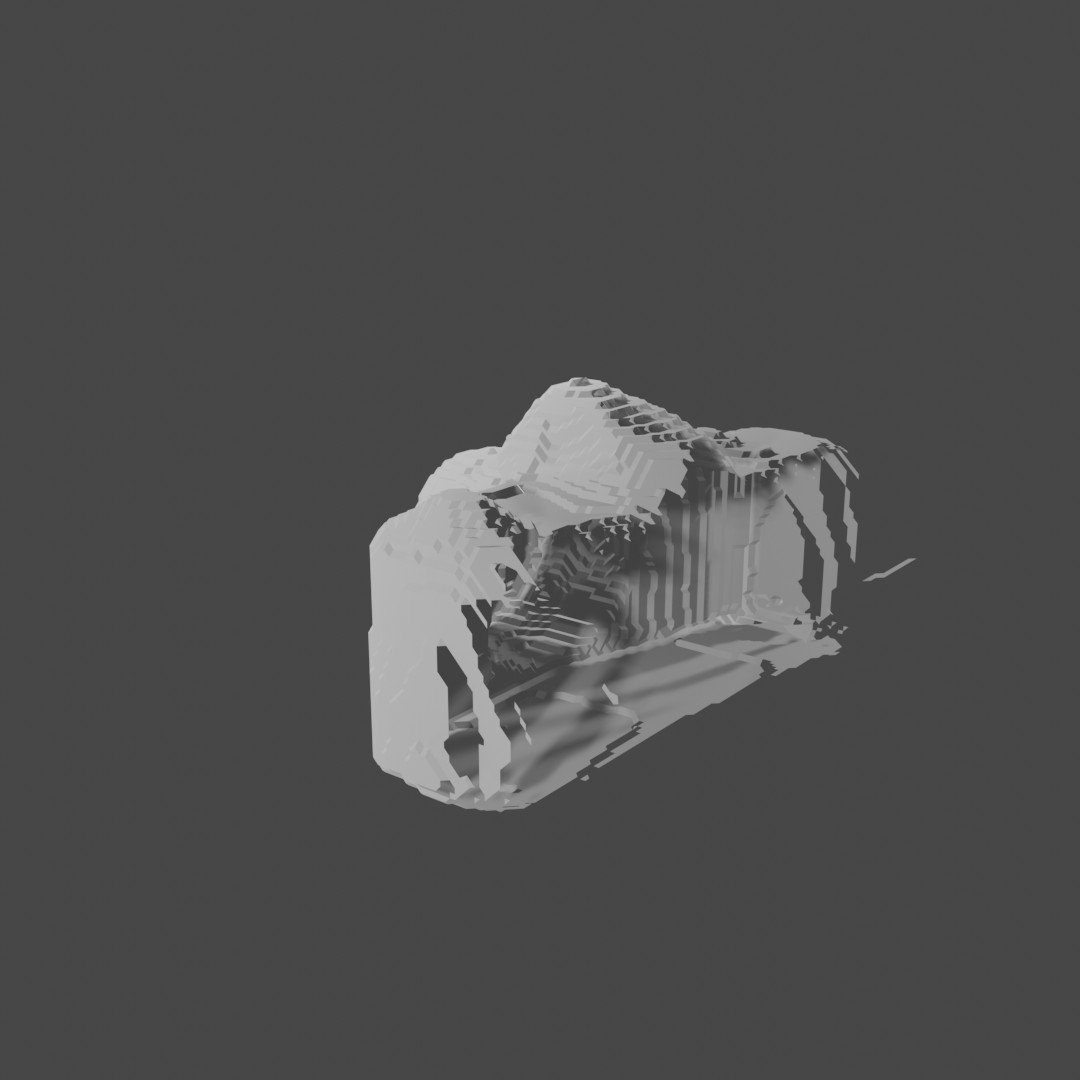}\!\!\!\\[0.2mm]%
	%
	\vspace{-2mm}
	\caption{Predictions on scenes containing considerable non-Fermat structure. \textbf{Left:} Comparison of our trained model with best Fermat case prediction. Our representation can leverage priors present in the measurement to recover primitives that would not be recovered by physically-based methods. \textbf{Right:} Multiple views of the reconstructed scenes.}
	\label{fig:results_beyond_Fermat_1+2}
\end{figure}

Figure~\ref{fig:results_beyond_Fermat_1+2} shows results over ShapeNet categories that exhibit considerable non-Fermat structures with respect to the scanned area (mugs, couches, cameras). For comparison, we tried the Fermat code by Xin \etal~\cite{xin2019theory} but we did not obtain competitive results for the considered resolutions. Instead, we compare our results to a hypothetical model that predicts meshes with perfect accuracy but obeys the specular Fermat recovery property. To do this, we simply filter triangles of the ground truth mesh that do not obey the specular Fermat criteria, also observed by \cite{liu2019analysis}, and refer to this model as \textit{best Fermat case} (BFC). The results illustrate how even the best Fermat prediction would result in sparse shapes, which in some cases makes predictions indistinguishable from the real object (couch). In contrast, the model trained using our representation can reconstruct more complete shapes by exploiting non-Fermat photons present in the measurement.

\begin{figure}[t]
	\centering\vspace{5mm}
	\rotatebox{90}{\textcolor{white}{Gp}} %
	\begin{minipage}[c]{.20\columnwidth}
		\centering \small GT 
	\end{minipage}	
	\begin{minipage}[c]{.20\columnwidth}
		\centering \small Ours (view 1) ~~
	\end{minipage}
	\begin{minipage}[c]{.20\columnwidth} 
		\centering \small Ours (view 2) ~~
	\end{minipage}
	\begin{minipage}[c]{.20\columnwidth}
		\centering \small Ours (view 3)~~~~~
	\end{minipage}\!\!\!\\
	\vspace{0.5mm}

	\includegraphics[width = 0.19\columnwidth]{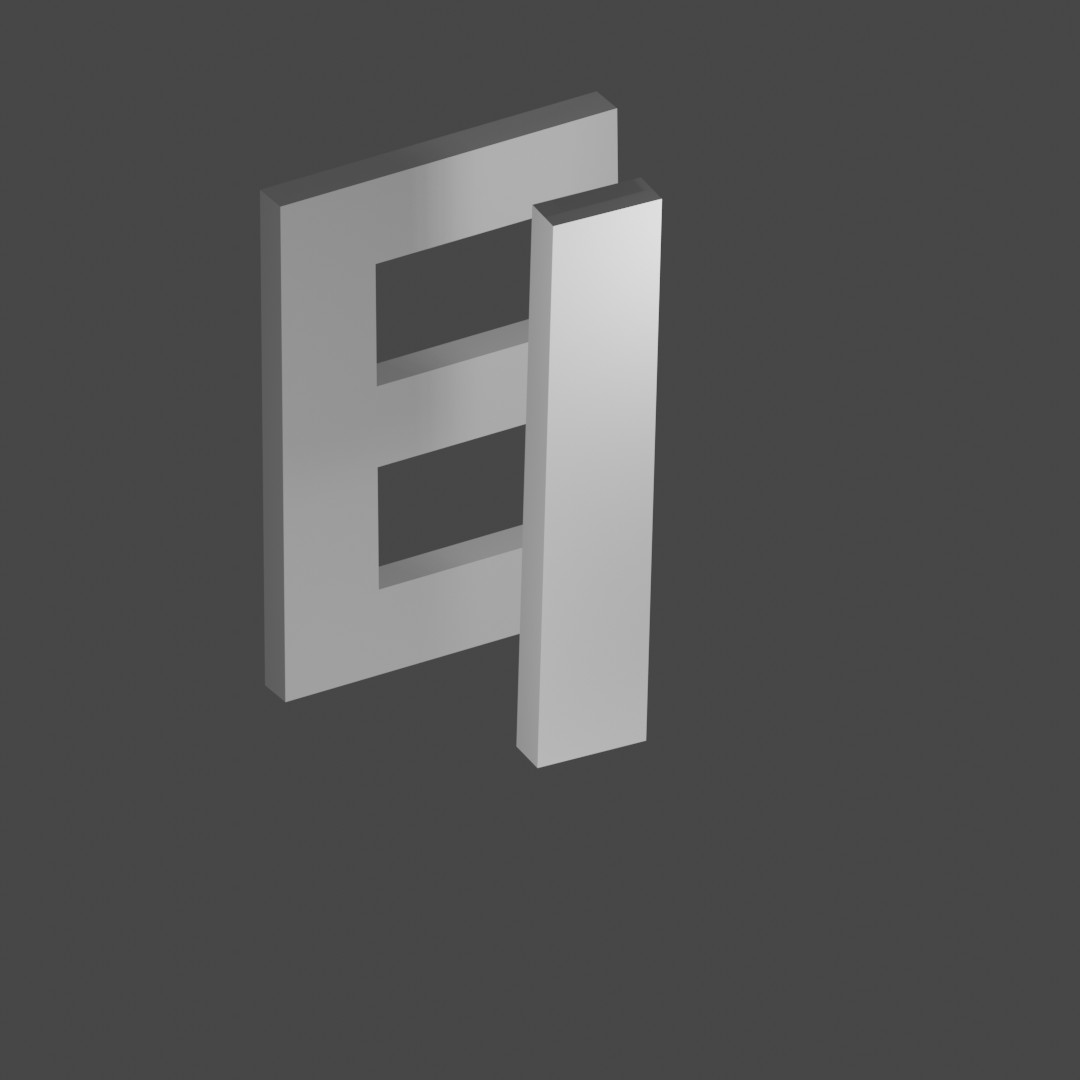}
	\includegraphics[width = 0.19\columnwidth]{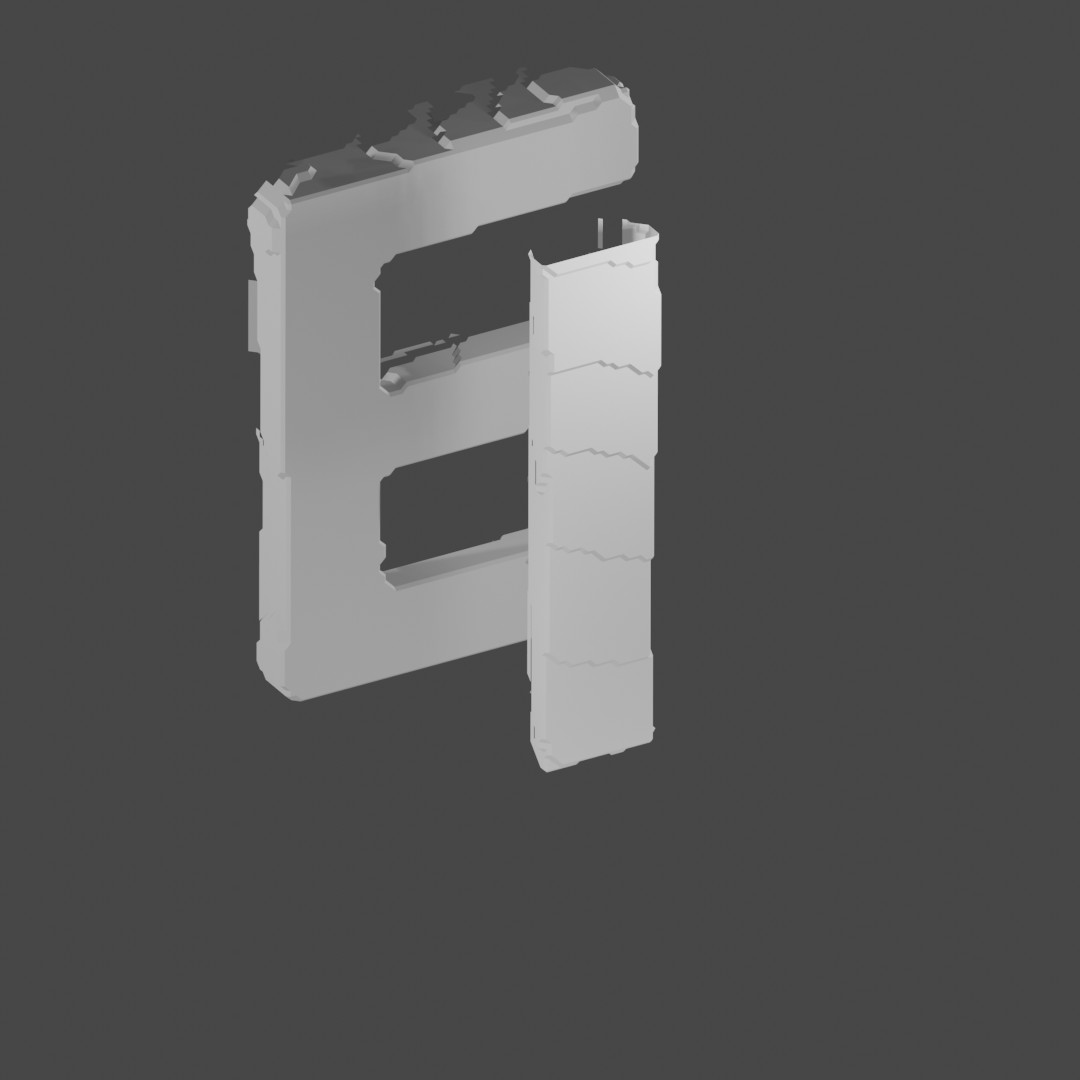}
	\includegraphics[width = 0.19\columnwidth]{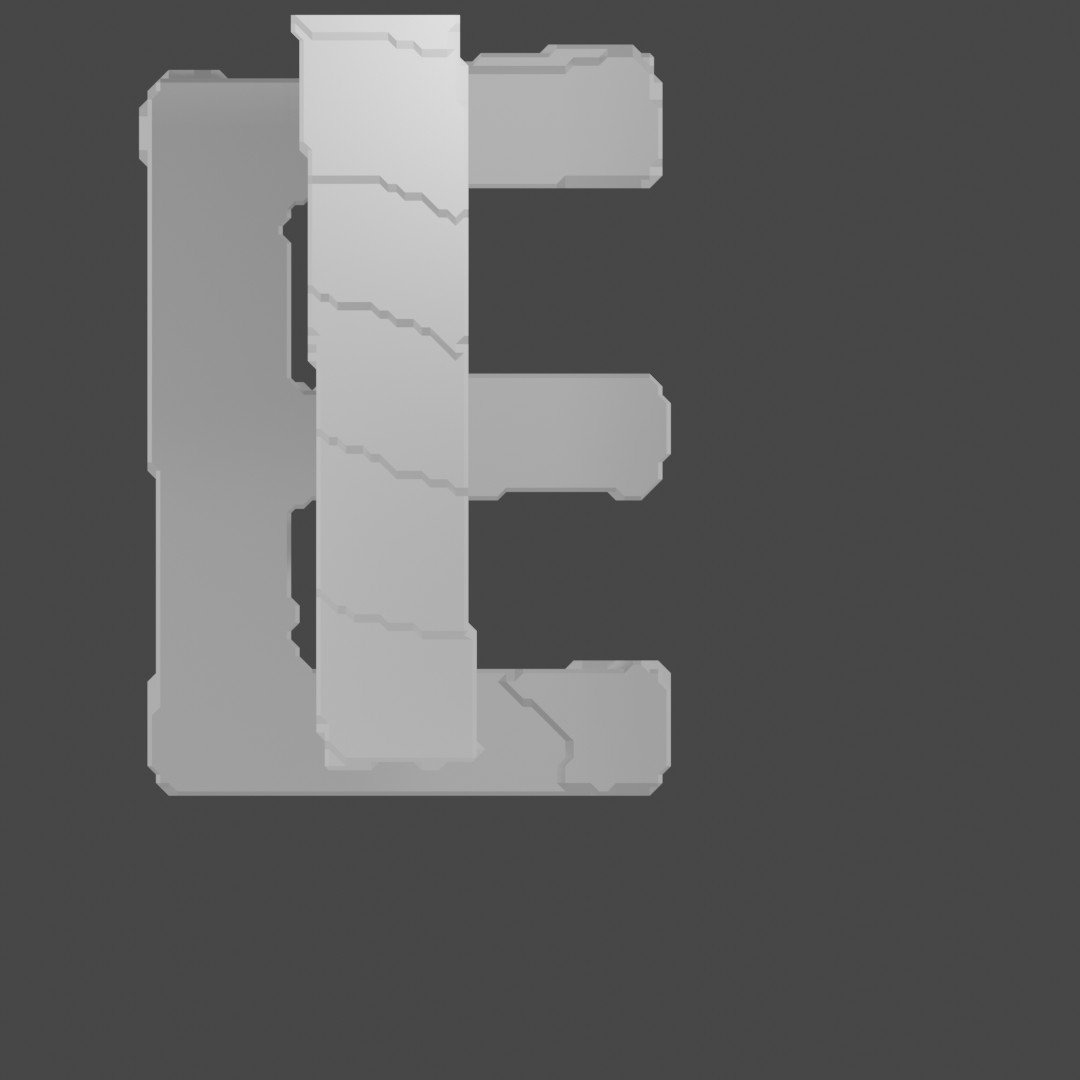} %
	\includegraphics[width = 0.19\columnwidth]{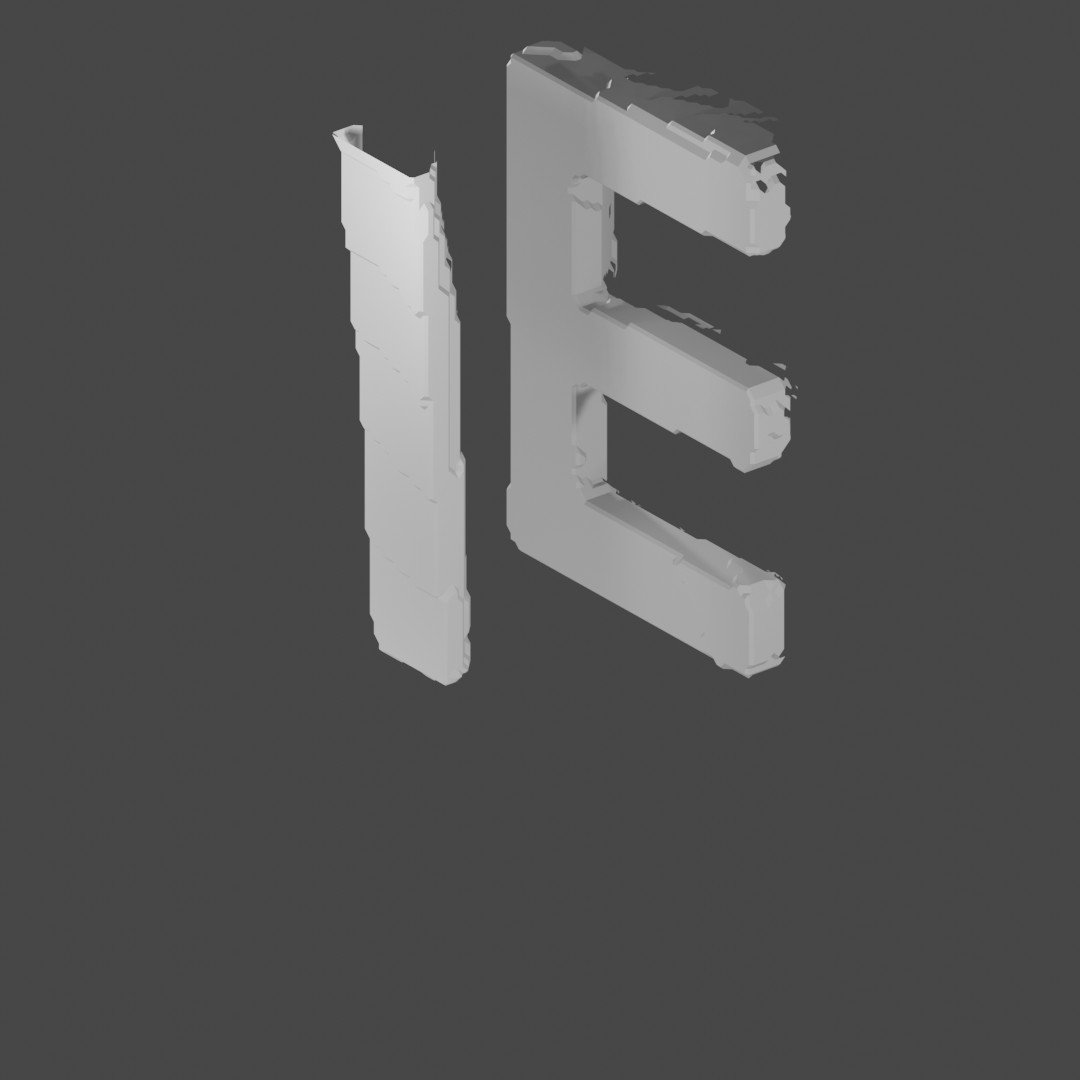}\!\!\!\\[1mm]%
	
	\includegraphics[width = 0.19\columnwidth]{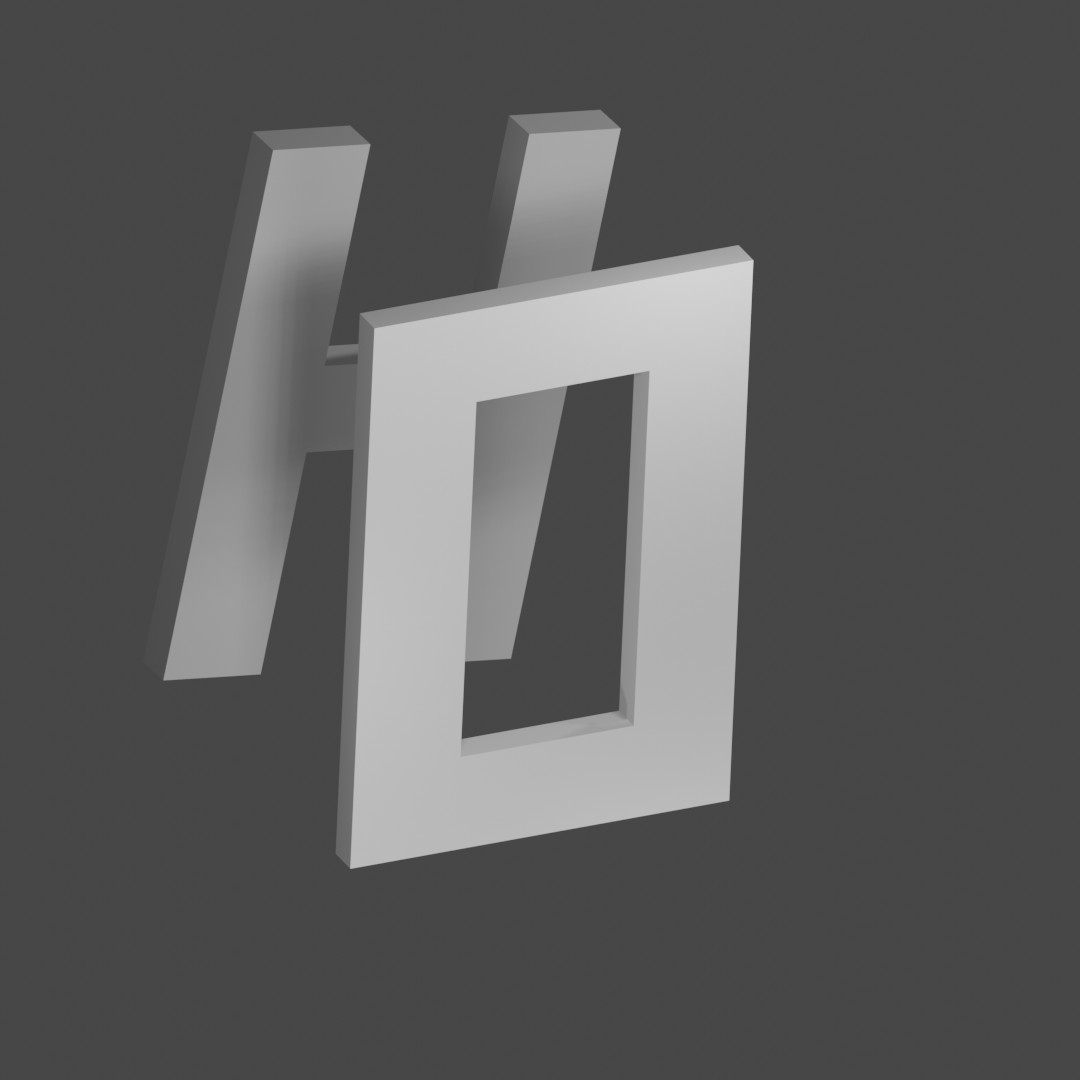}
	\includegraphics[width = 0.19\columnwidth]{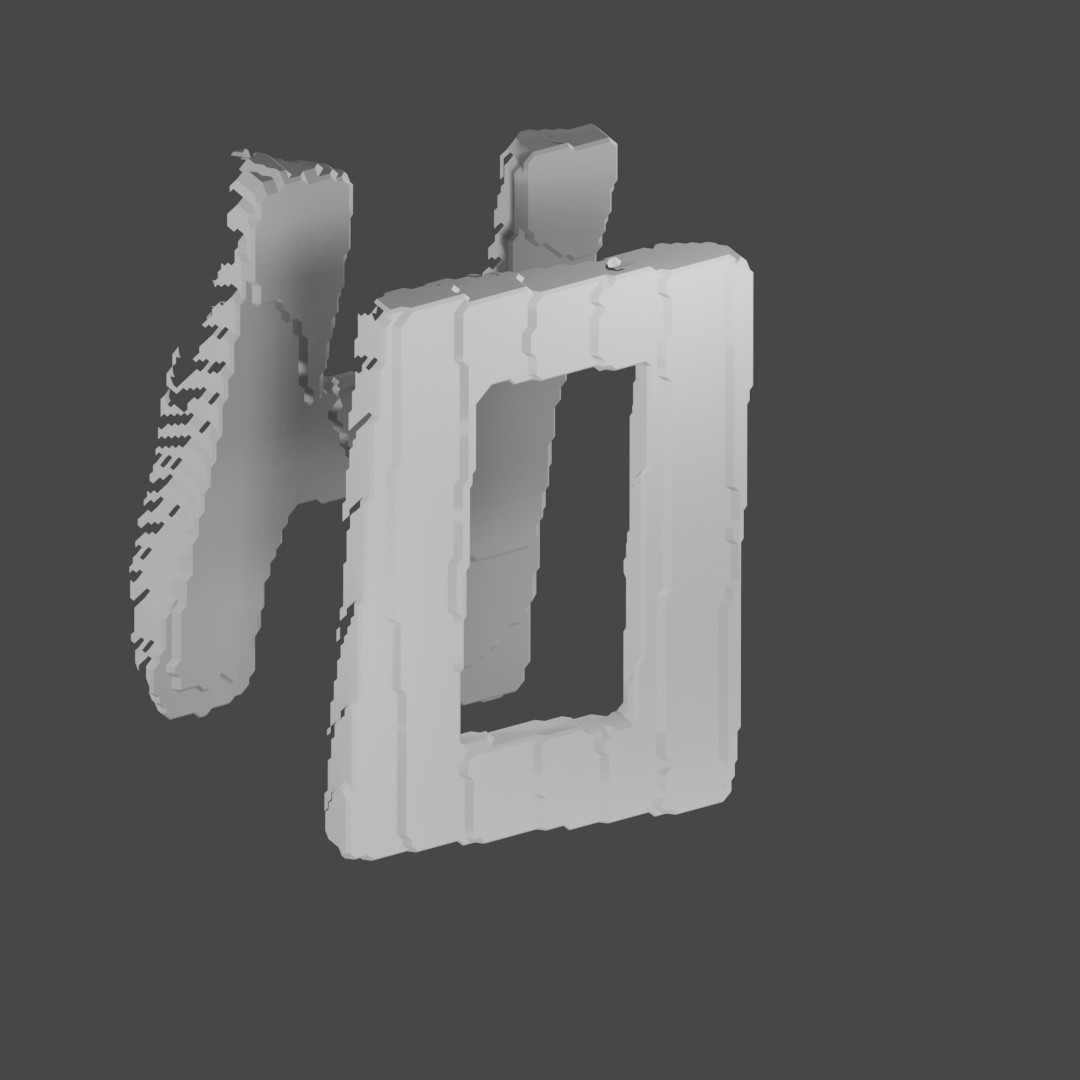} %
	\includegraphics[width = 0.19\columnwidth]{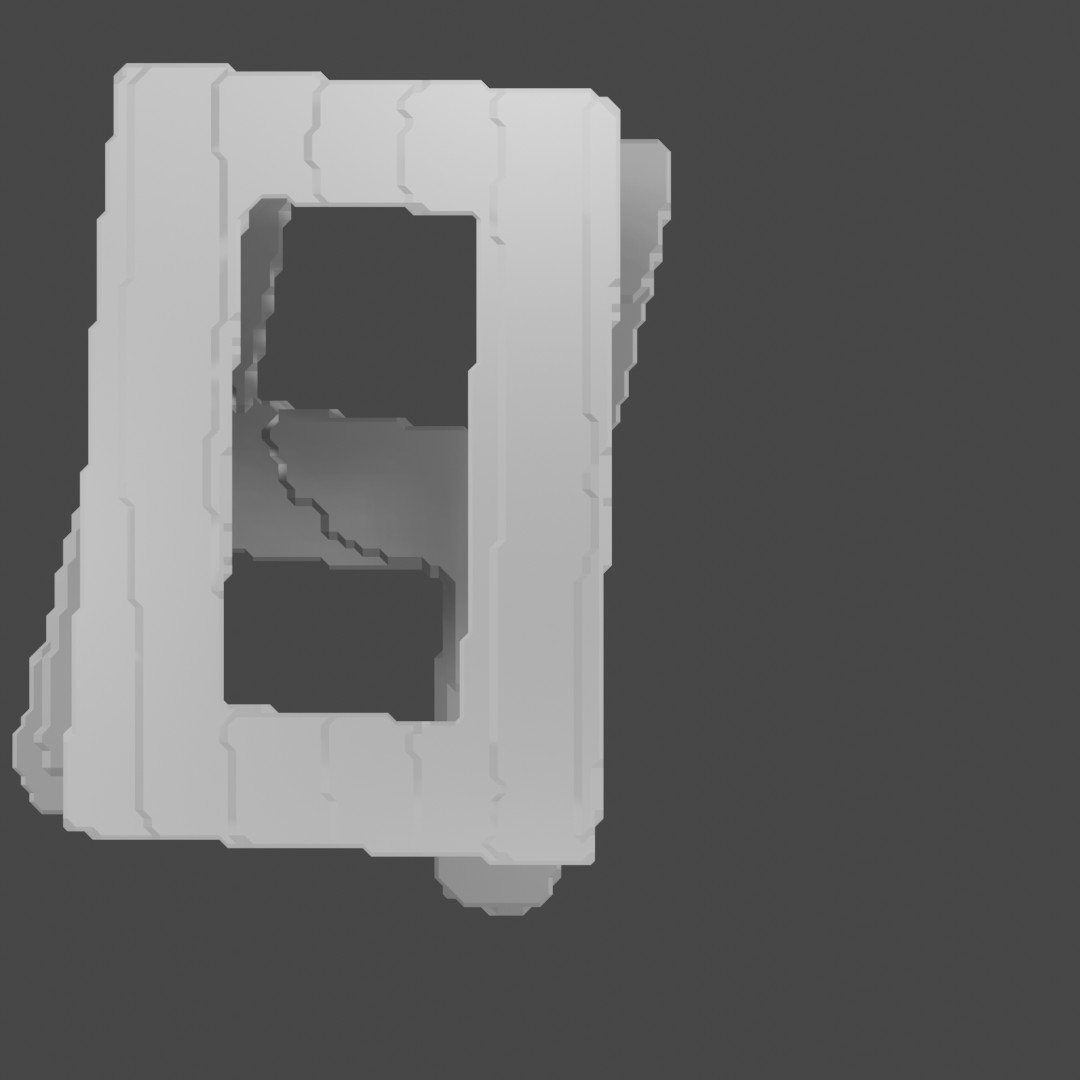} %
	\includegraphics[width = 0.19\columnwidth]{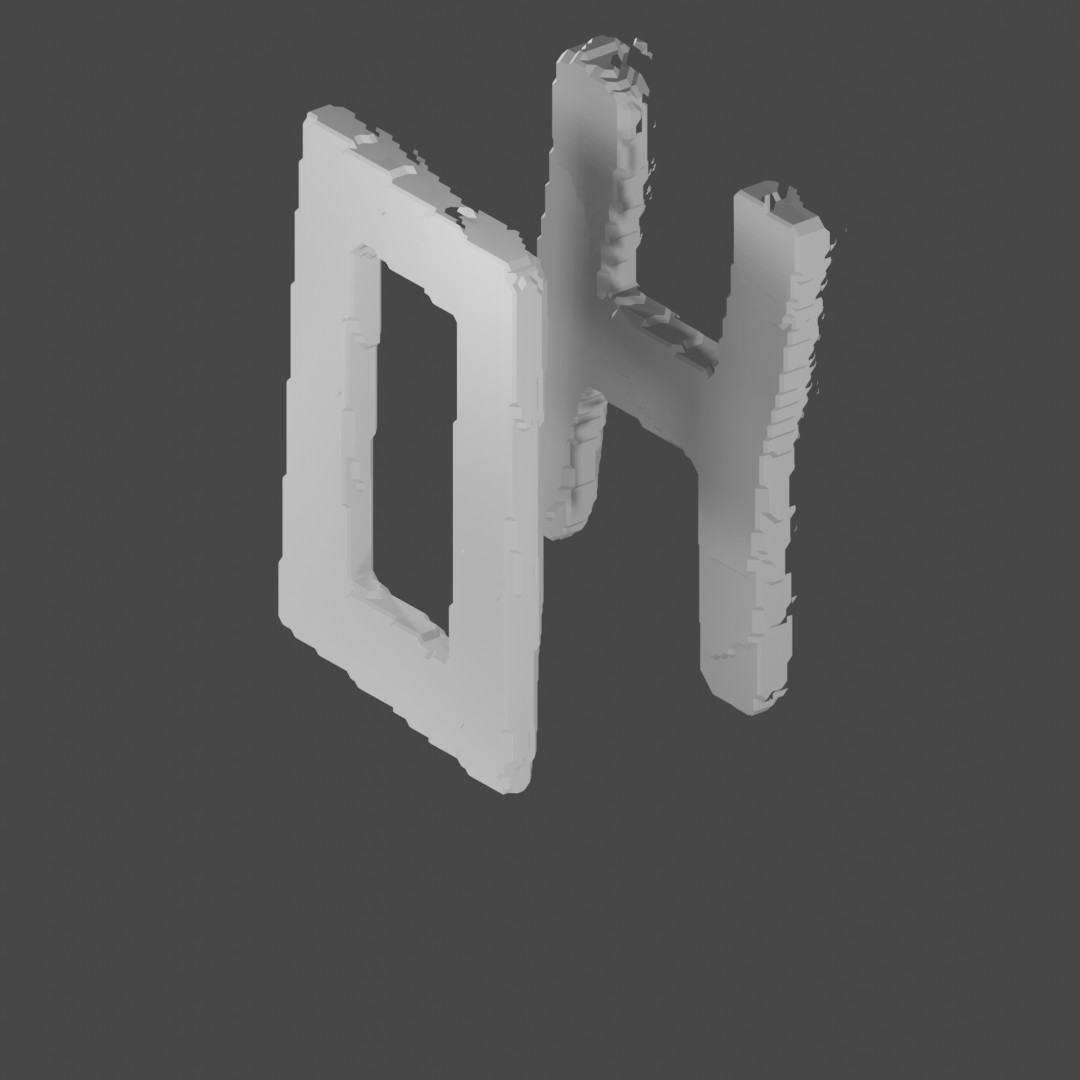}\!\!\!\\[1mm]%
	
	\includegraphics[width = 0.19\columnwidth]{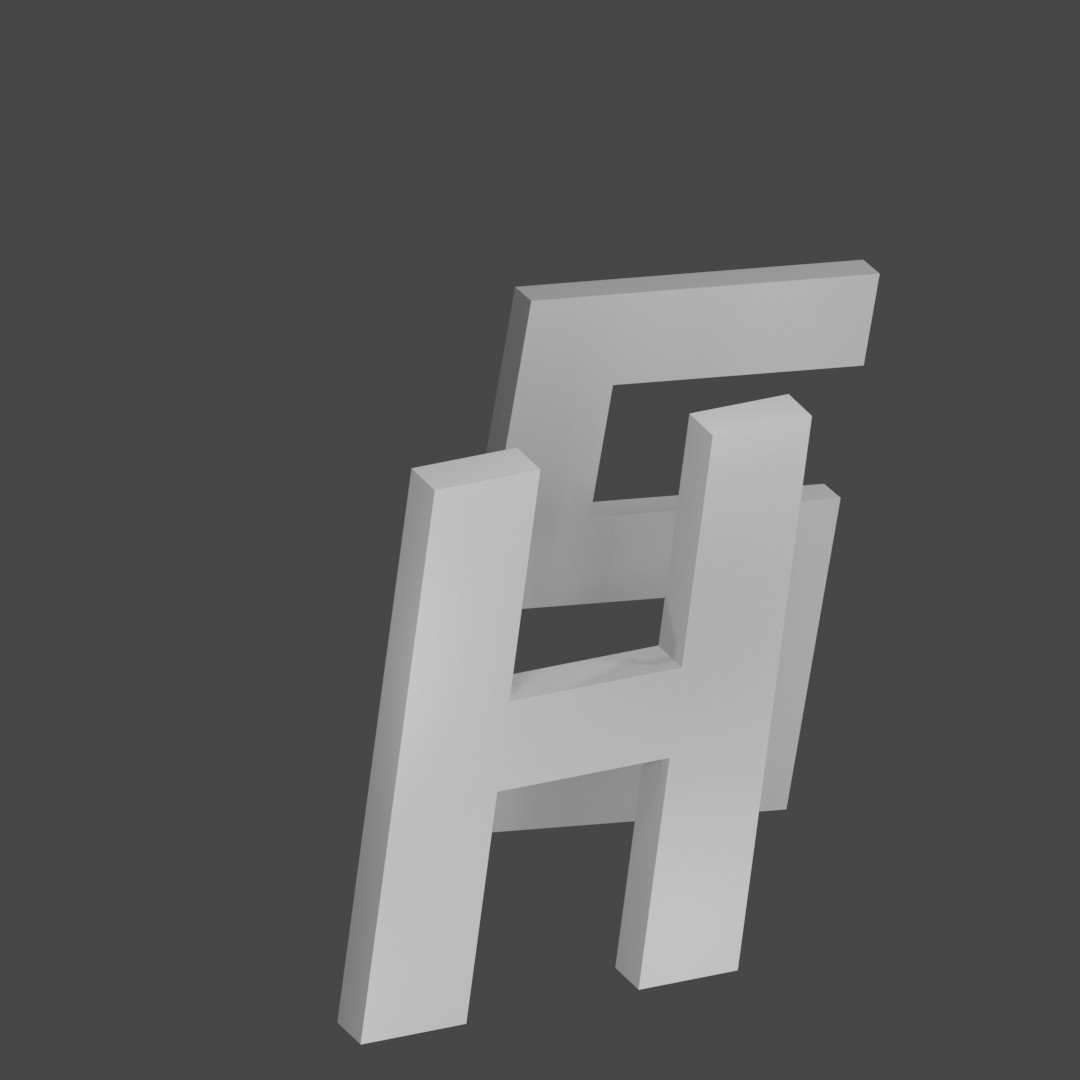}
	\includegraphics[width = 0.19\columnwidth]{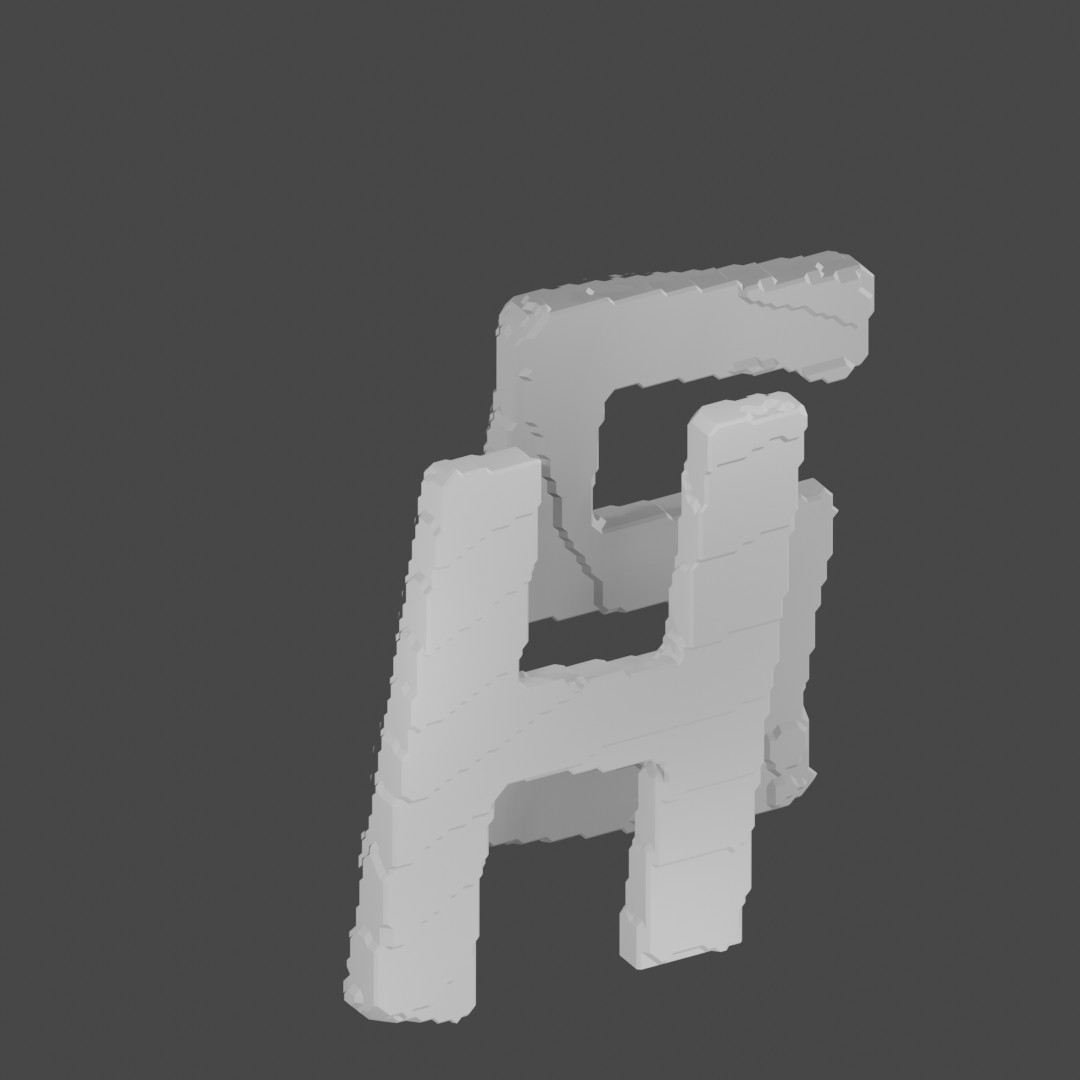} %
	\includegraphics[width = 0.19\columnwidth]{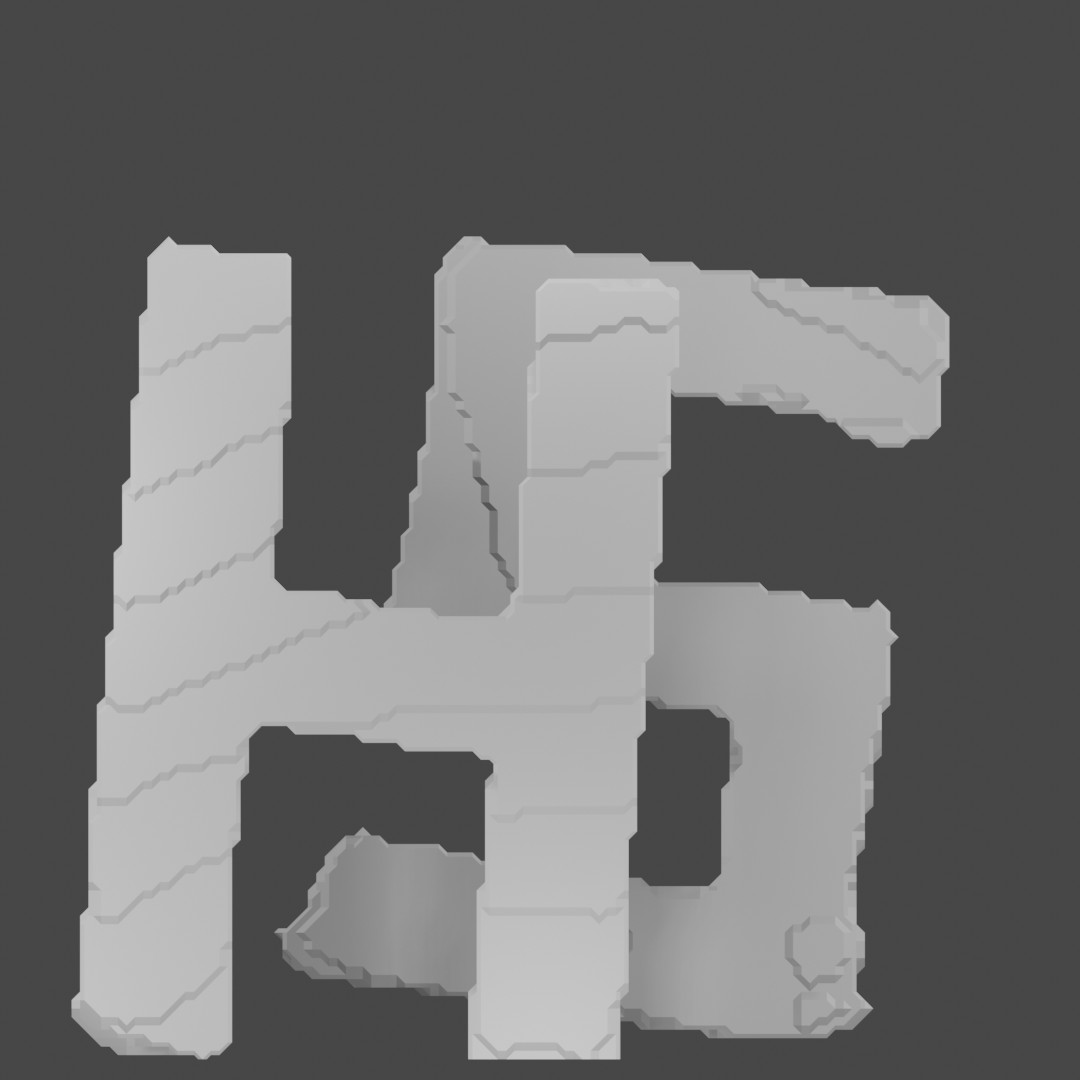} %
	\includegraphics[width = 0.19\columnwidth]{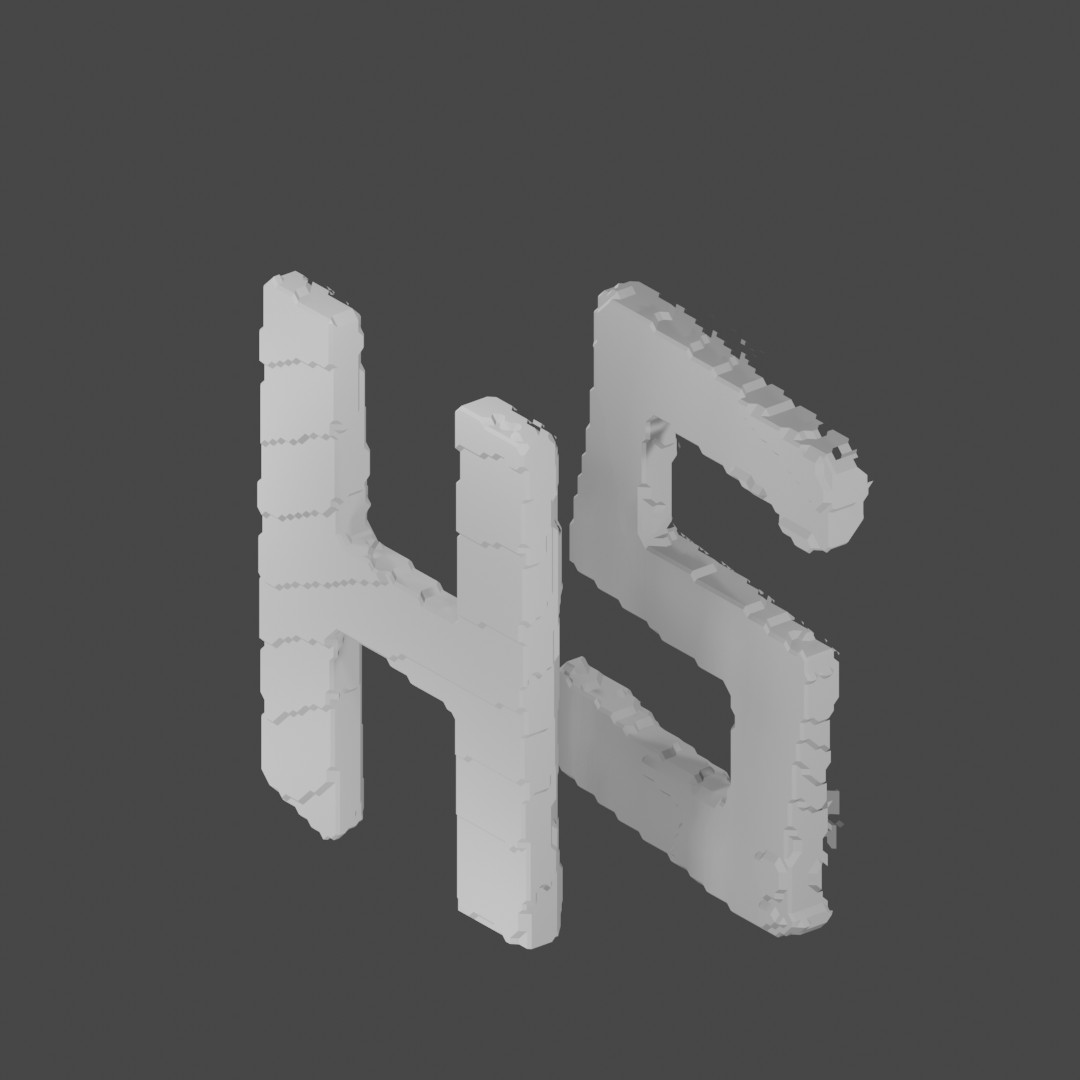}\!\!\!\\[1mm]%
	\vspace{-2mm}
	\caption{Predictions on scenes with strong self-occlusion. Our trained model is able to gain self-occlusion invariance over the entire space with different degrees of occlusion.  (Lighting+shading applied for illustration purposes).}
	\label{fig:results_self-occlusion}
\end{figure}

We also trained our network on cases of strong self-occlusion. For this experiment, we use only letters and flat geometric figures, and generate $60,000$ scenes with two objects each. Figure \ref{fig:results_self-occlusion} shows how our learned representation allows for recovering severely occluded geometry in test scenes while predicting shapes with high fidelity.

\subsection{Domain Adaptation and Real-World Predictions}
In order to conduct predictions for real-world datasets, we consider adding noise to the rendered transients with existing approaches. The study of realistic noise models lies beyond the scope of our work and thus we claim no contributions in this regard.

\begin{figure}[h!]
	\centering
	\includegraphics[trim=11mm 8mm 5mm 3mm,clip,width = 0.21\columnwidth]{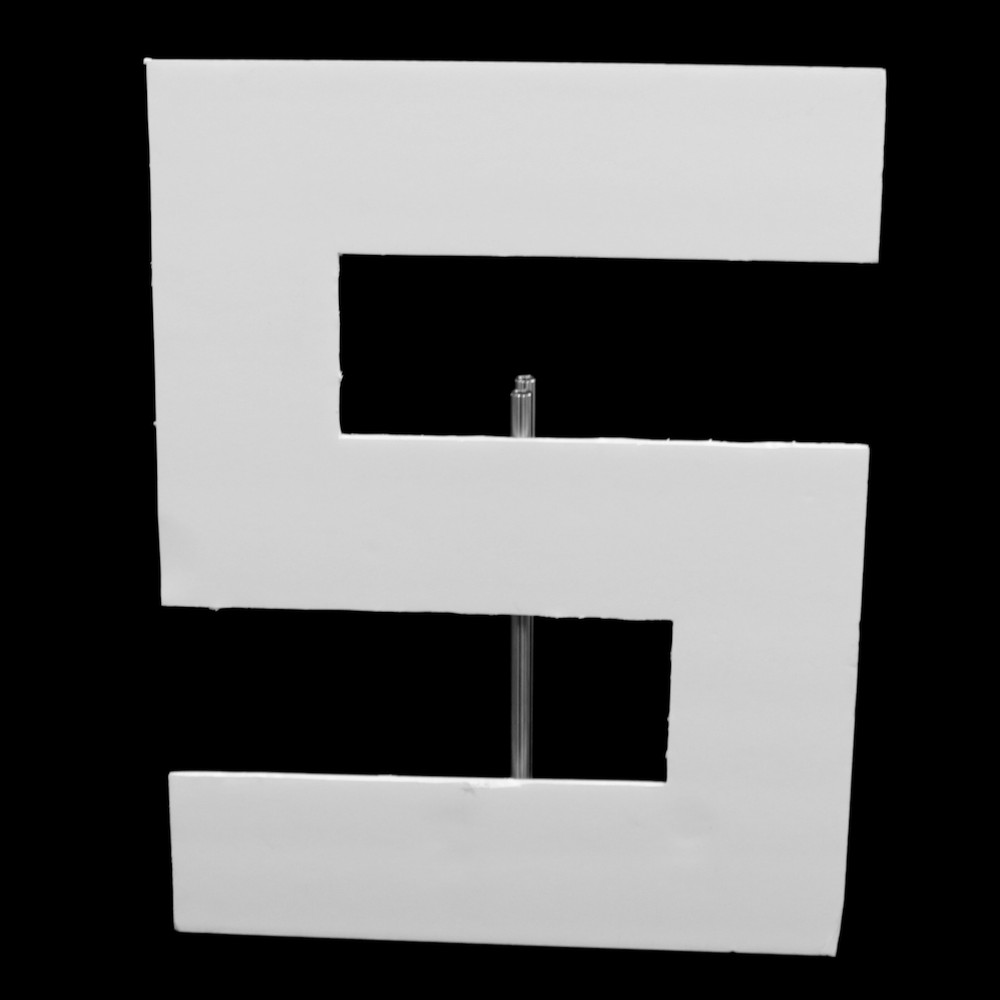}\vspace{0.25mm}
	\includegraphics[trim=11mm 8mm 5mm 3mm,clip,width = 0.21\columnwidth]{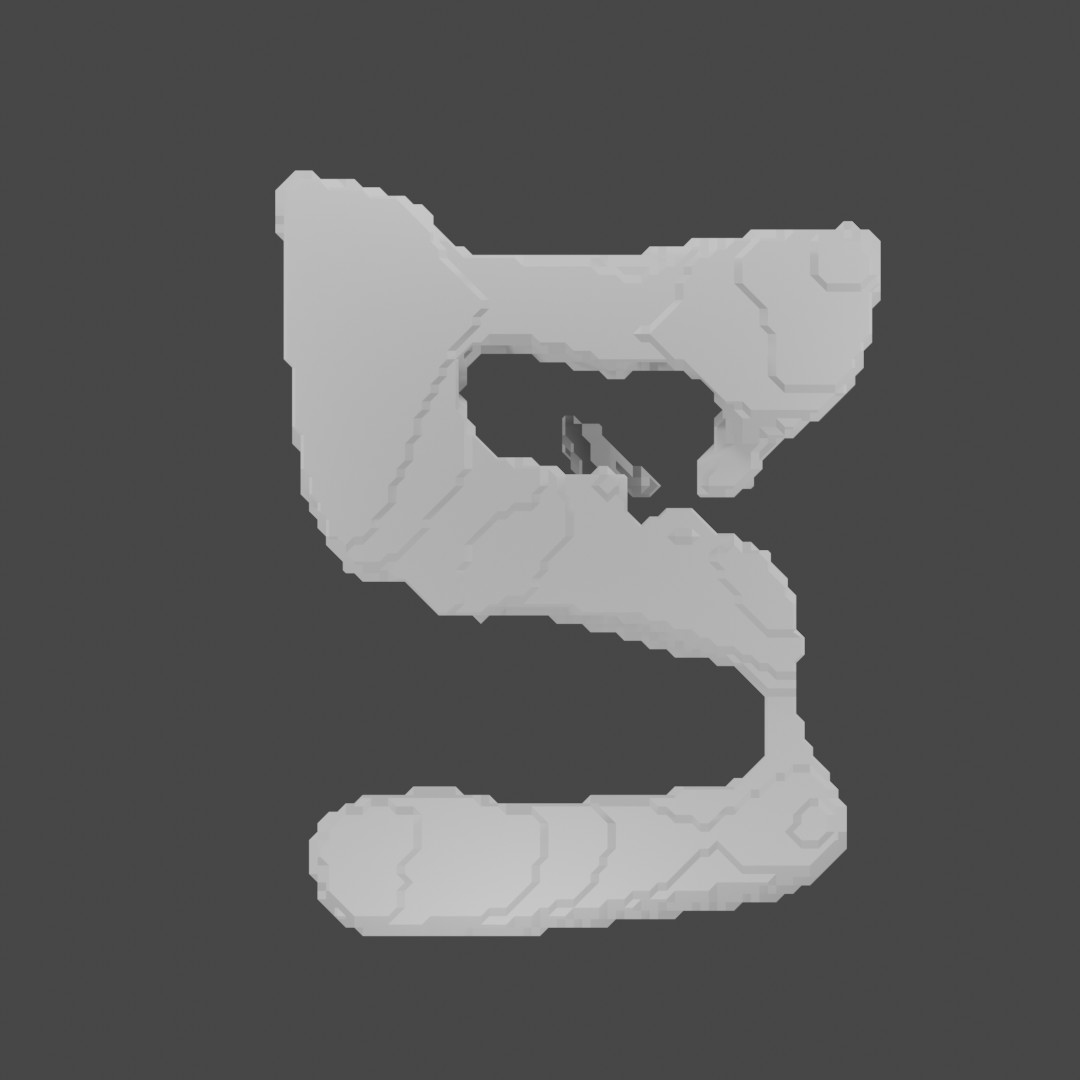}\hspace{5mm}
	\includegraphics[trim=11mm 8mm 5mm 3mm,clip,width = 0.21\columnwidth]{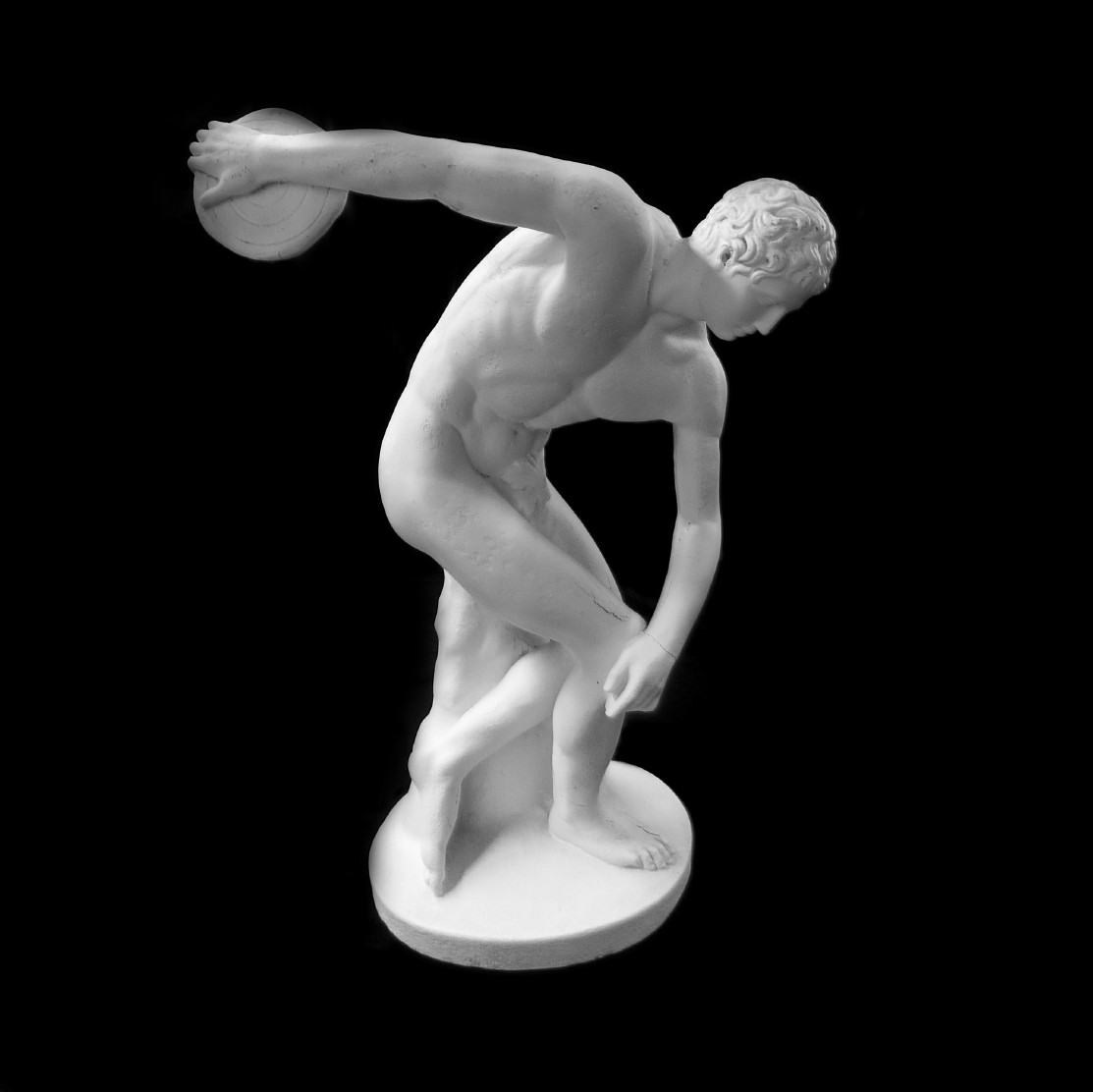}\vspace{0.25mm}
	\includegraphics[trim=11mm 8mm 5mm 3mm,clip,width = 0.21\columnwidth]{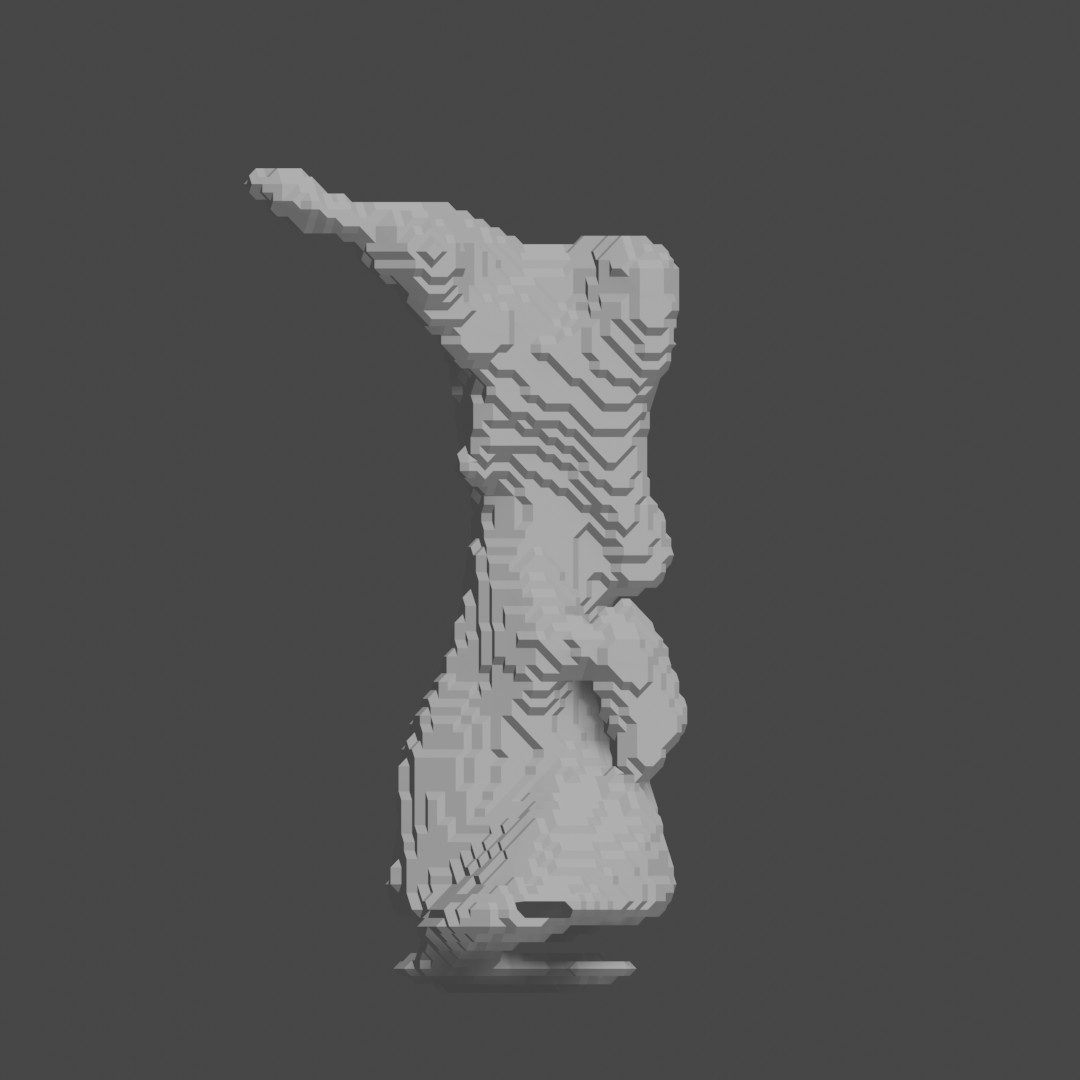}\\[0.1mm]
	%
	\caption{Predictions on experimental data using domain adaptation.}
	\label{fig:results_5}
	\vspace{-4mm}
\end{figure}

We tried using the code by Hernandez \etal~\cite{hernandez2017computational} to simulate realistic SPAD responses on our rendered transients, but we found this strategy unfeasible for training as generating measurements might take from seconds to minutes. Therefore, we implement a similar approximation as Chen \etal~\cite{chen2020learned} and Grau \etal~\cite{grau2020deep}:

\begin{equation}
m_{spad}(\vec{s_i}, \tau) = \mathcal{P}(C*m(\vec{s_i}, \tau) + B),  \ \ B \sim uniform[a,b] \label{eq:spad_model}
\end{equation}

where $\mathcal{P}$ takes samples from a Poisson distribution, $C$ is a scale constant of the scene and $B$ is a global base noise value. To ensure that our pipeline and representation are robust to noise, we trained and evaluated our models using this augmentation strategy. In general, we observe that generalization scores degrade, but the predictions on noisy inputs remain competitive (see supplemental material). We used these models to predict on the experimental datasets acquired by \cite{otoole2018confocal} and \cite{lindell2019wave}, shown in Figure~\ref{fig:results_5}. We hypothesize that our results on real data may be rooted on the physical inaccuracy between the approximated augmentation model and that of the real capture.

\section{Discussion}
In this paper we introduced a new geometric representation of non-line-of-sight scenes that allows for computing more complete meshes of the targets without suffering from self-occlusion artifacts. Follow-ups of our work include the computation of detailed surfaces, which could be achieved by considering SDF functions over the occlusion volume and high-frequency point encodings. Additionally, we believe that our representation could be further leveraged to perform NLoS scene-understanding tasks by adding 3D segmentation terms to the occlusion loss. However, as we remark in the latter sections, these advances will require as well the investigation of accurate and efficient noise models in order to capture real-world capabilities.

\section{Acknowledgements}
This work was supported by the European Research Council under ERC Starting Grant ``ECHO''.


%
%
\bibliographystyle{splncs04}
\bibliography{egbib}


\section{Appendix}
In the following sections we provide further explanations regarding our results and implementations.

\subsection{Dataset Generation Parameters}
In order to train our networks, we generate $3D$ scenes by randomly sampling meshes with affine transformations. Specifically, we chose transformation parameters that: 1) do not generate too small meshes, 2) apply moderate rotations of the objects, 3) do not place the object or its parts outside of the non-line-of-sight scene (the unit cube). Our datasets were generated with the following ranges:

\begin{itemize}
	\item Scaling:  $s\in [0.6,0.85]$
	\item Rotation: $[\theta_x,\theta_z]\in[0^\circ,10^\circ]$ and $\theta_y\in[0^\circ,20^\circ]$
	\item Translation: $[T_x,T_y]\in[-0.30,0.30]$ and $T_z\in[-0.40,0.40]$
\end{itemize}

where the $z^+$ axis corresponds to the normal of the wall and  $y^+$ is the object's up-vector.

\subsection{Training Configurations}
Throughout our work we considered the architectures by Peng et. al \cite{peng2020convolutional} and Mescheder et. al. \cite{2019occupancy} to evaluate our representation. Predictions shown in the main paper and this manuscript correspond to the former architecture, as it resulted in much faster convergence during training. Trainings were performed on a NVIDIA GeForce RTX 2080, which achieves convergence in 4 days for the convolutional model (9 million parameters) and 8 days for the CBN-cell architecture (57 million parameters). Only for the case of the statues and sculptures datasets we found necessary to increase the machine capacity of the convolutional model to 16 million parameters, as this showed to better resolve thin features (arms, legs, etc). Specifics of the architecture configurations will be provided along with training code and data upon acceptance of the submission.

\subsection{Occlusion Fields Runtime Performance}

\begin{figure}[t]
	\centering
	\includegraphics[width = \columnwidth]{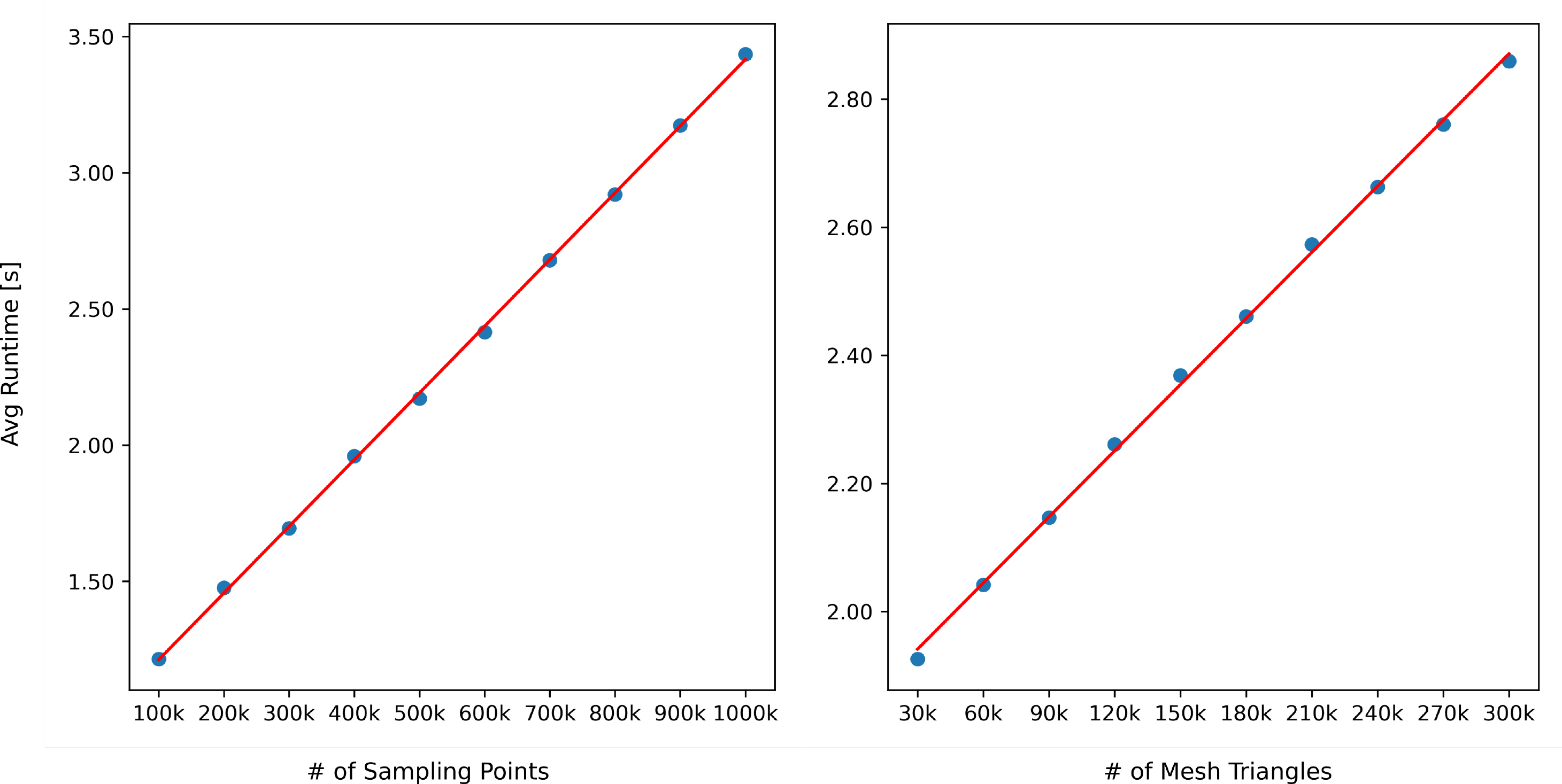}
	\vspace{-5mm}
	\caption{Runtime performance of our implementation of the occlusion/visibility test w.r.t the number of sampling points and scene triangles. When experimenting with the number of sampling points (left) we used a standard scene tessellation of 30,000 triangles. In both cases, our implementation achieves linear behavior.}
	\label{fig:occlusion_runtimes_1}
\end{figure}

We use Nvidia’s OptiX framework to implement our occlusion/visibility test. We create buffers for $n$ sampling points in order to compute $m \times n$ binary occlusion values, where $m$ denotes the number of virtual wall sensors. Figure \ref{fig:occlusion_runtimes_1} shows average runtime performances of our implementation with respect to the number of sampling points and triangles present in the scene. For the particular case of our datasets, we processed 50k-60k of the generated $3D$ scenes by sampling 400k points and $5\times 5$ sensing positions, resulting in approximately 27h and 33h of computation, respectively.
 
\subsection{Occlusion Fields vs. Volumetric Methods}
\begin{figure*}
	\centering
	\rotatebox{90}{\small~~~~GT\textcolor{white}p} 
	\includegraphics[trim=11mm 8mm 5mm 3mm,clip,width = 0.16\columnwidth]{./figures/results/fig_statues/gt/000001_front.jpg}%
	\includegraphics[trim=11mm 8mm 5mm 3mm,clip,width = 0.16\columnwidth]{./figures/results/fig_statues/gt/000002_front.jpg}%
	\includegraphics[trim=11mm 8mm 5mm 3mm,clip,width = 0.16\columnwidth]{./figures/results/fig_statues/gt/000044_front.jpg}%
	\includegraphics[trim=11mm 8mm 5mm 3mm,clip,width = 0.16\columnwidth]{./figures/results/fig_statues/gt/000045_front.jpg}%
	\includegraphics[trim=11mm 8mm 5mm 3mm,clip,width = 0.16\columnwidth]{./figures/results/fig_statues/gt/000046_front.jpg}
	\includegraphics[trim=11mm 8mm 5mm 3mm,clip,width = 0.16\columnwidth]{./figures/results/fig_statues/gt/000049_front.jpg}\\[1mm]
	\rotatebox{90}{\small~~~~~Ours\textcolor{white}p} 		
	\includegraphics[trim=11mm 8mm 5mm 3mm,clip,width = 0.16\columnwidth]{./figures/results/fig_statues/ours/000001_front.jpg}%
	\includegraphics[trim=11mm 8mm 5mm 3mm,clip,width = 0.16\columnwidth]{./figures/results/fig_statues/ours/000002_front.jpg}%
	\includegraphics[trim=11mm 8mm 5mm 3mm,clip,width = 0.16\columnwidth]{./figures/results/fig_statues/ours/000044_front.jpg}%
	\includegraphics[trim=11mm 8mm 5mm 3mm,clip,width = 0.16\columnwidth]{./figures/results/fig_statues/ours/000045_front.jpg}%
	\includegraphics[trim=11mm 8mm 5mm 3mm,clip,width = 0.16\columnwidth]{./figures/results/fig_statues/ours/000046_front.jpg}
	\includegraphics[trim=11mm 8mm 5mm 3mm,clip,width = 0.16\columnwidth]{./figures/results/fig_statues/ours/000049_front.jpg}\\[1mm]
	\rotatebox{90}{\small~~~~~LCT\textcolor{white}p} 		
	\includegraphics[trim=11mm 8mm 5mm 3mm,clip,width = 0.16\columnwidth]{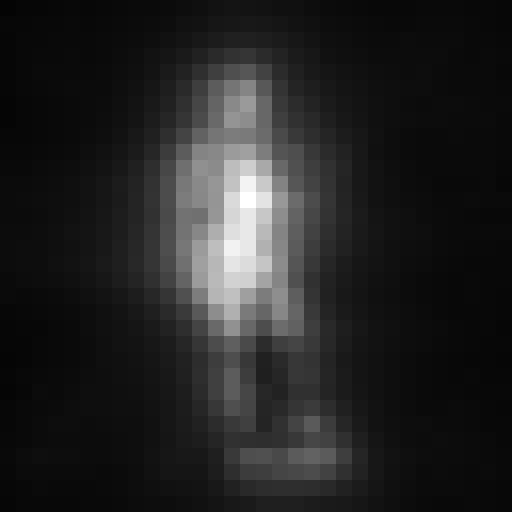}%
	\includegraphics[trim=11mm 8mm 5mm 3mm,clip,width = 0.16\columnwidth]{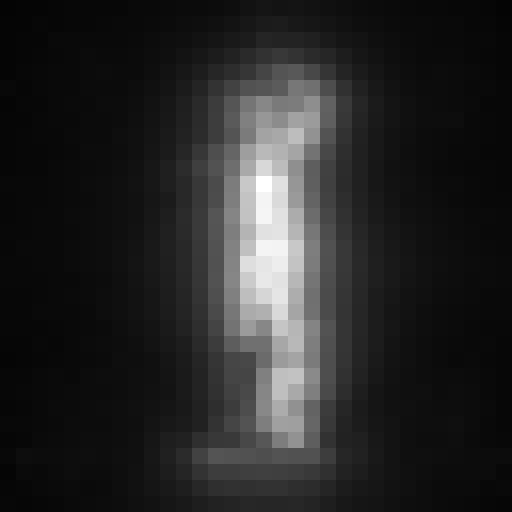}%
	\includegraphics[trim=11mm 8mm 5mm 3mm,clip,width = 0.16\columnwidth]{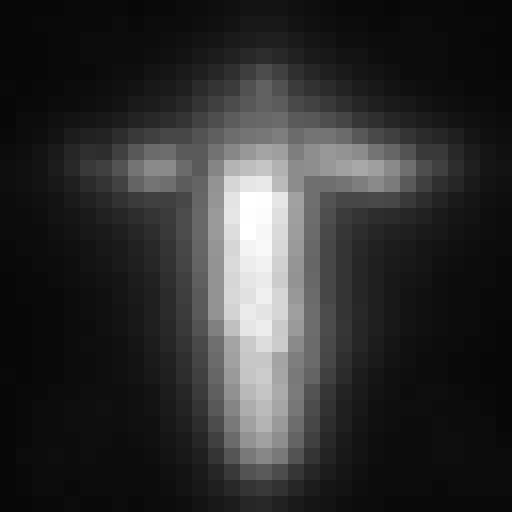}%
	\includegraphics[trim=11mm 8mm 5mm 3mm,clip,width = 0.16\columnwidth]{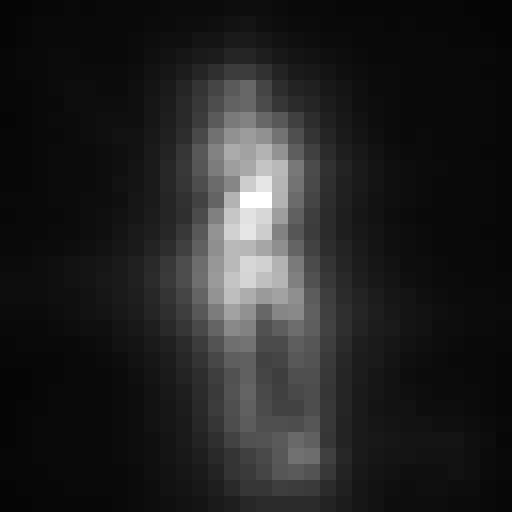}%
	\includegraphics[trim=11mm 8mm 5mm 3mm,clip,width = 0.16\columnwidth]{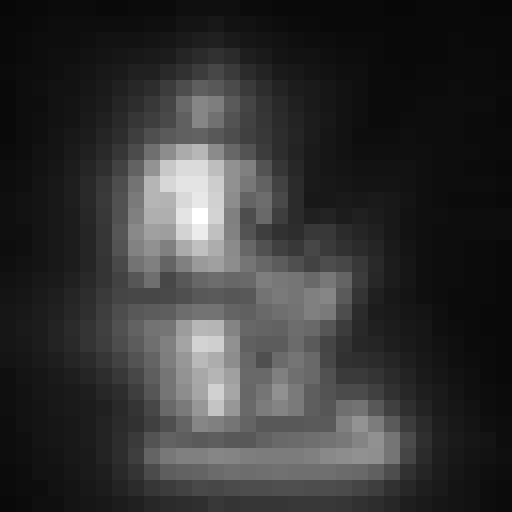}
	\includegraphics[trim=11mm 8mm 5mm 3mm,clip,width = 0.16\columnwidth]{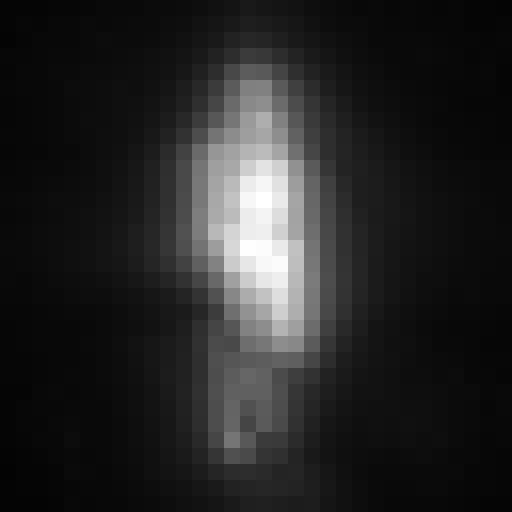}\\[1mm]
	%
	\vspace{-2mm}
	\caption{Illustration of the predictive capabilities of our trained model in comparison with a physically-based volumetric approach (LCT) on scenes with complex geometry. Some of the thin and sharp structures are difficult to reconstruct for the volumetric model at the considered resolution ($32\times 32\times 256$)}
	\label{fig:results_1_statues_suppmat}
\end{figure*}

For illustration, we show the result of our models next to predictions made by volumetric methods. We found the light cone transform algorithm \cite{otoole2018confocal} (LCT) to yield better predictions on our scenes than fk-migration \cite{lindell2019wave} and phasor fields \cite{liu2019Phasor}. The target scenes have diffuse surface and were captured with $32\times 32\times 256$ resolution. Figure \ref{fig:results_1_statues_suppmat} shows reconstructions on synthetic scenes with complex features, sampled within a scan area of 2\,m$\times$2\,m and a 64\,ps-resolved sensor.

\begin{figure}[t!]
	\centering
	\begin{minipage}[c]{.18\columnwidth}
		\centering \small GT 
	\end{minipage}	
	\begin{minipage}[c]{.18\columnwidth}
		\centering \small Ours ($32^3$)
	\end{minipage}
	\begin{minipage}[c]{.18\columnwidth} 
		\centering \small Ours ($64^3$)
	\end{minipage}
	\begin{minipage}[c]{.18\columnwidth} 
	\centering \small Ours ($128^3$)
	\end{minipage}
	\begin{minipage}[c]{.18\columnwidth} 
	\centering \small LCT
	\end{minipage}
	\vspace{0.1mm}
	
	\centering
	\includegraphics[trim=11mm 8mm 5mm 3mm,clip,width = 0.18\columnwidth]{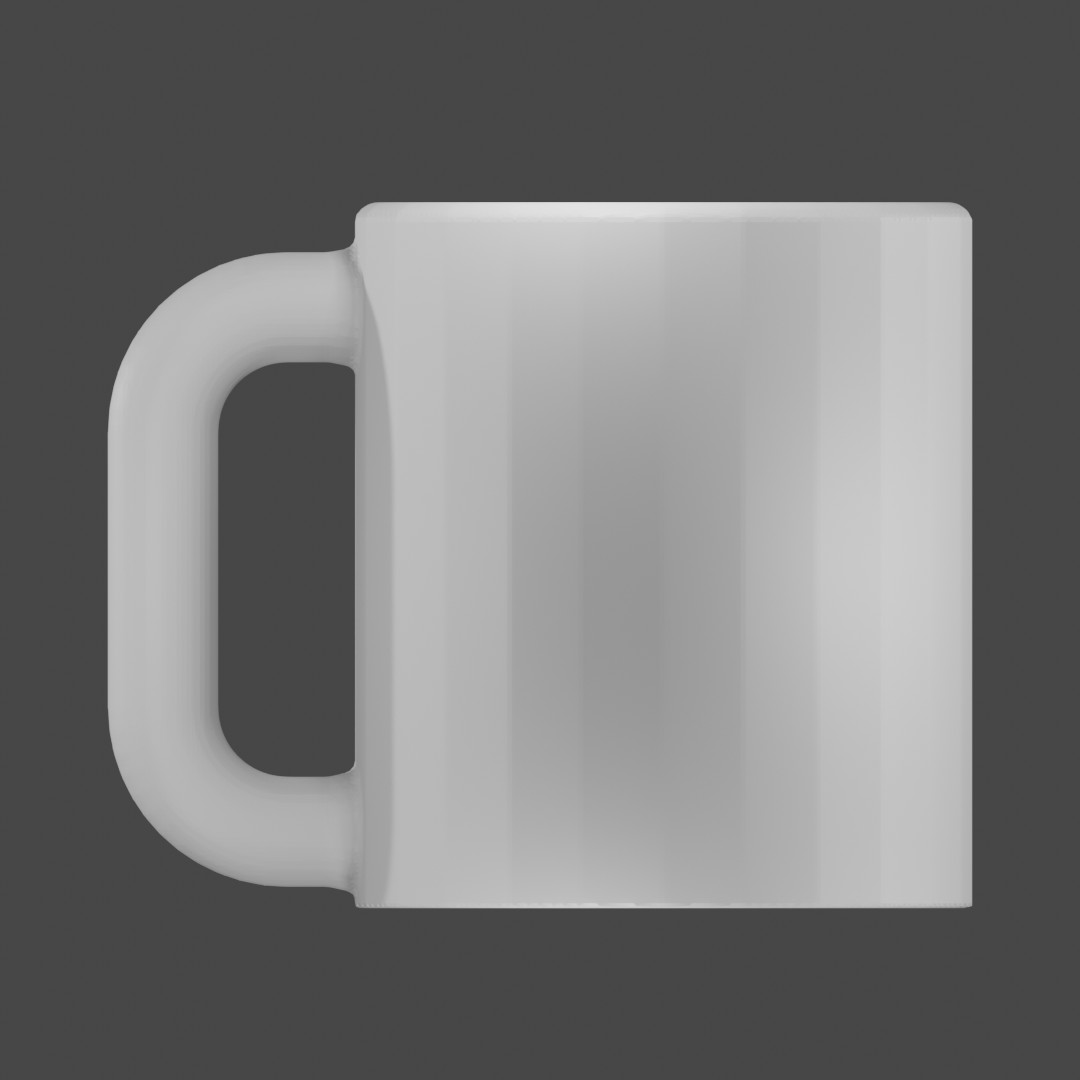}
	\includegraphics[trim=11mm 8mm 5mm 3mm,clip,width = 0.18\columnwidth]{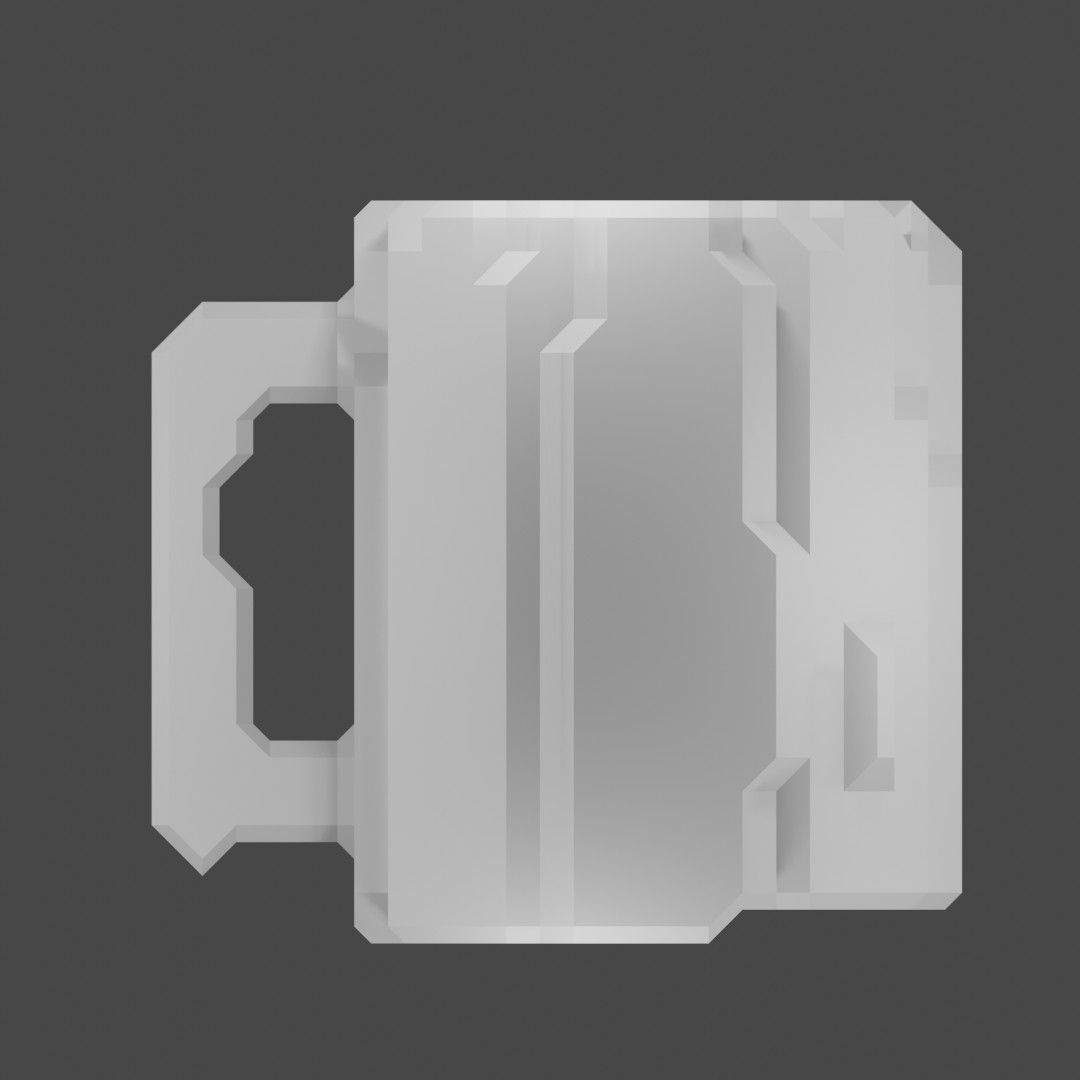}
	\includegraphics[trim=11mm 8mm 5mm 3mm,clip,width = 0.18\columnwidth]{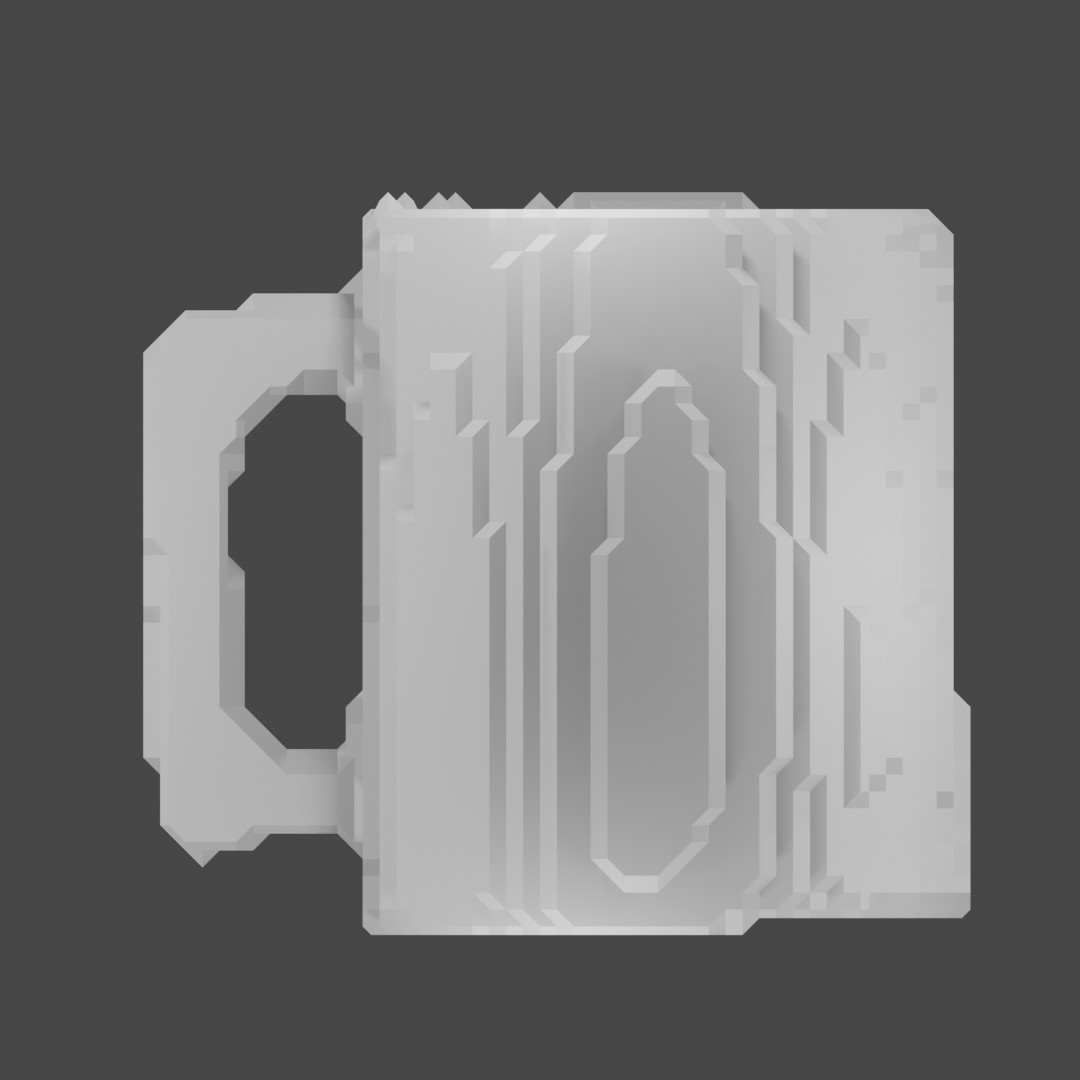}
	\includegraphics[trim=11mm 8mm 5mm 3mm,clip,width = 0.18\columnwidth]{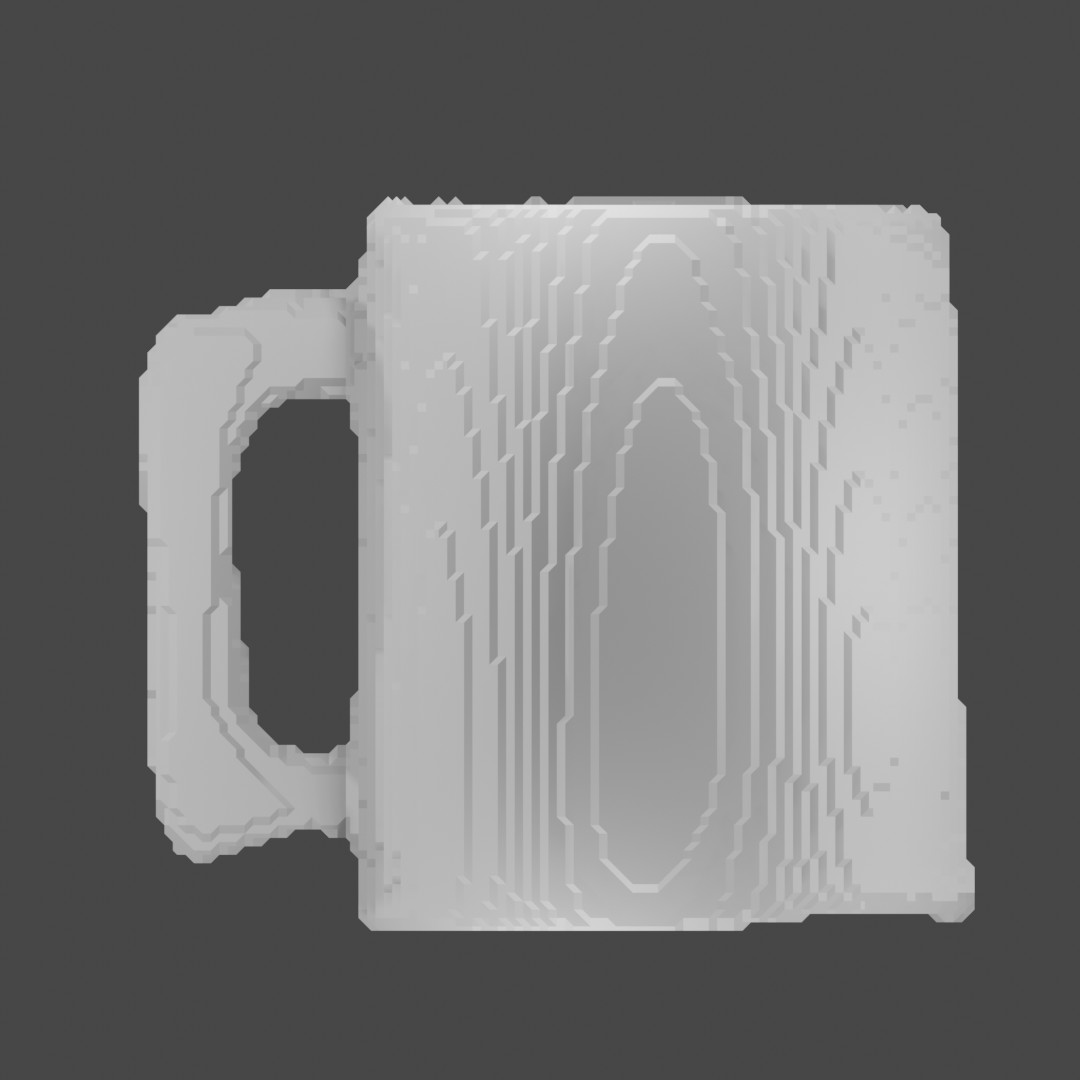}
	\includegraphics[trim=11mm 8mm 5mm 3mm,clip,width = 0.18\columnwidth]{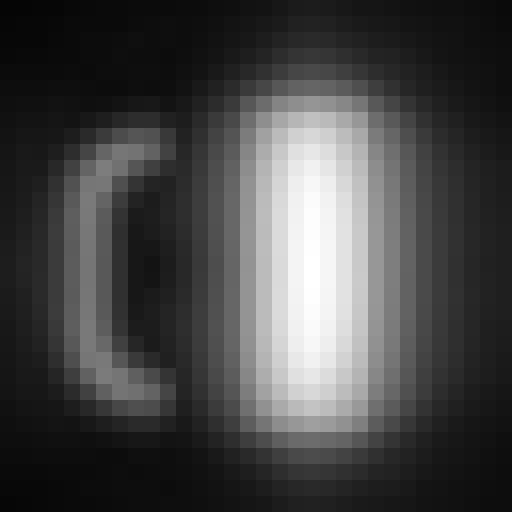}\\[0.1mm]
	\includegraphics[trim=11mm 8mm 5mm 3mm,clip,width = 0.18\columnwidth]{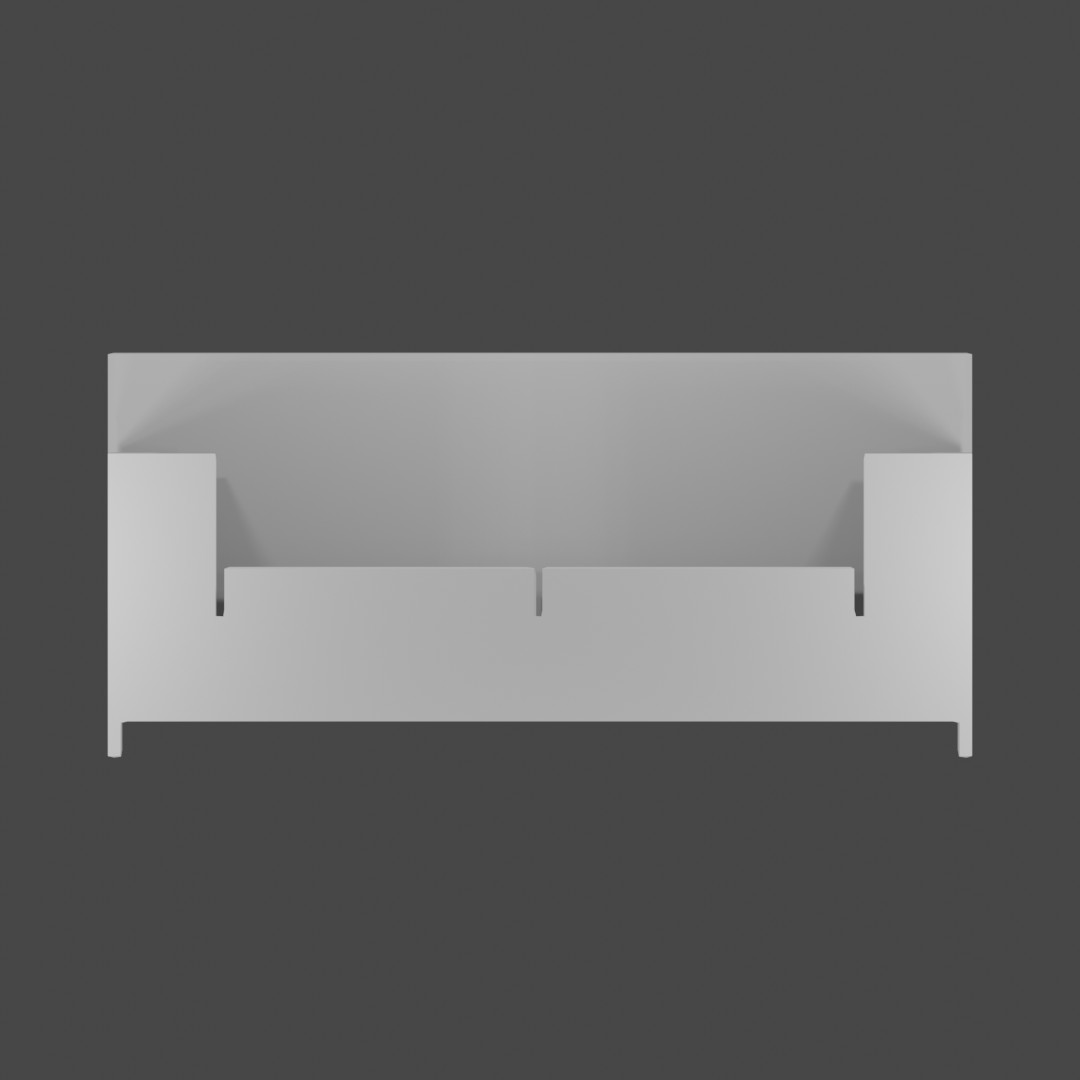}
	\includegraphics[trim=11mm 8mm 5mm 3mm,clip,width = 0.18\columnwidth]{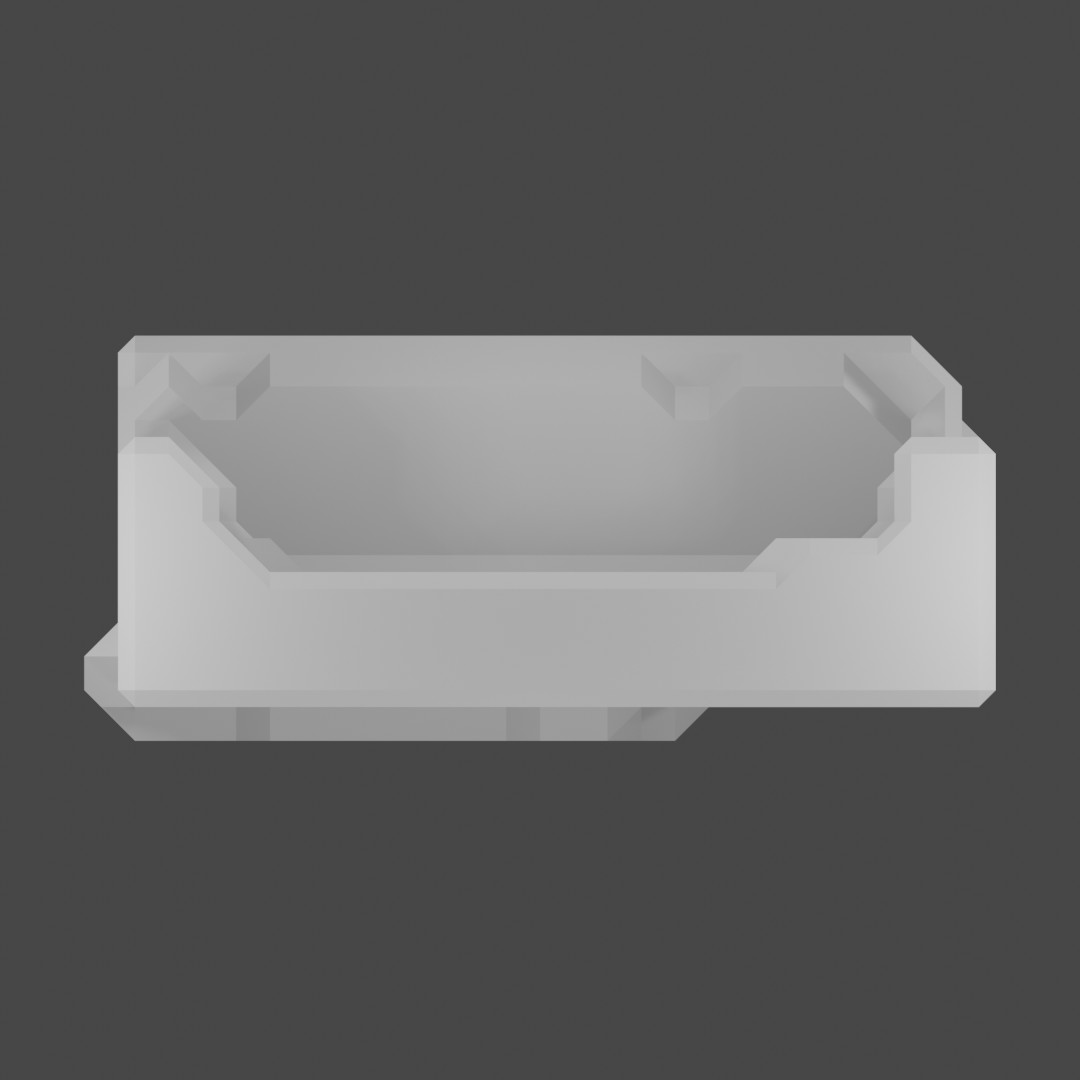}
	\includegraphics[trim=11mm 8mm 5mm 3mm,clip,width = 0.18\columnwidth]{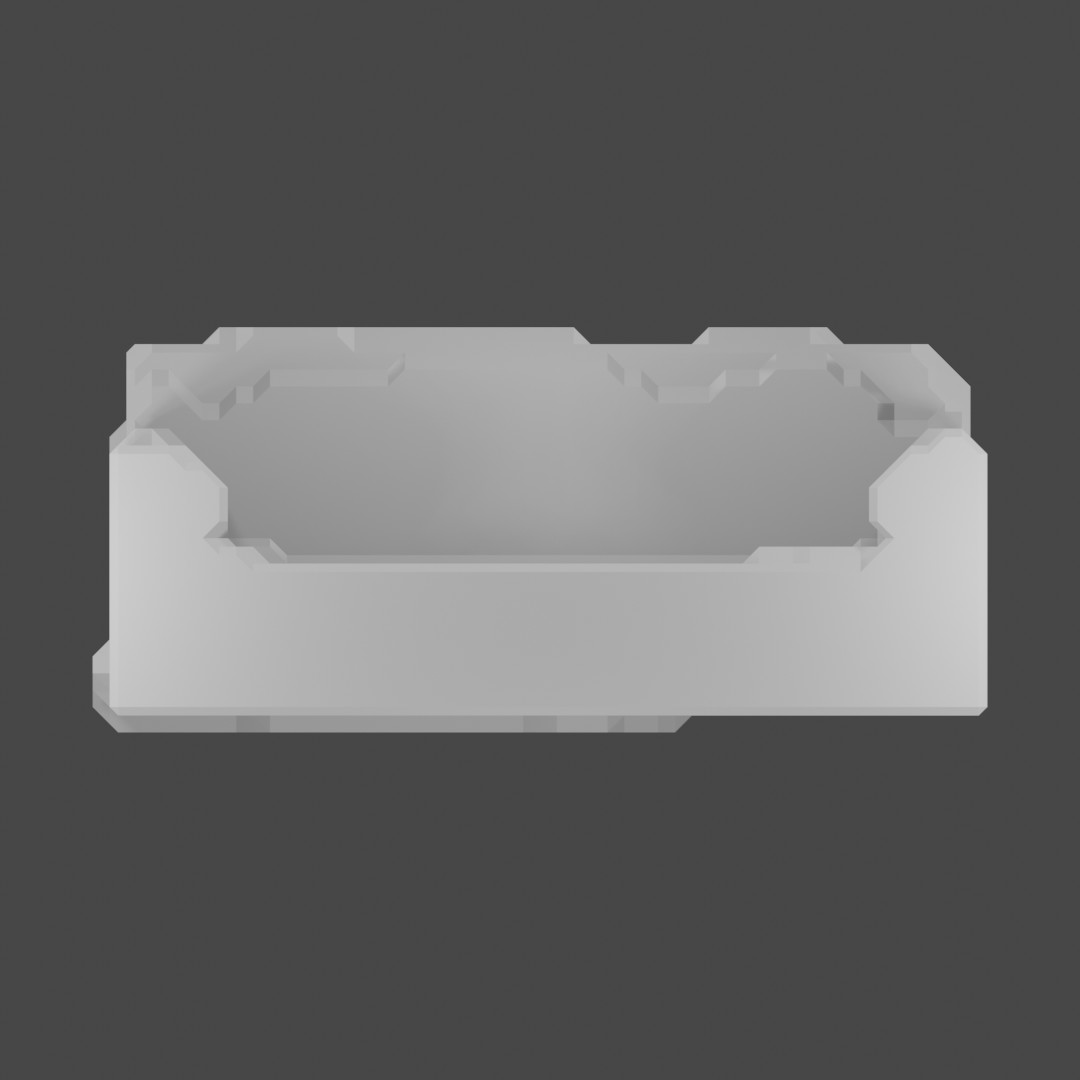}
	\includegraphics[trim=11mm 8mm 5mm 3mm,clip,width = 0.18\columnwidth]{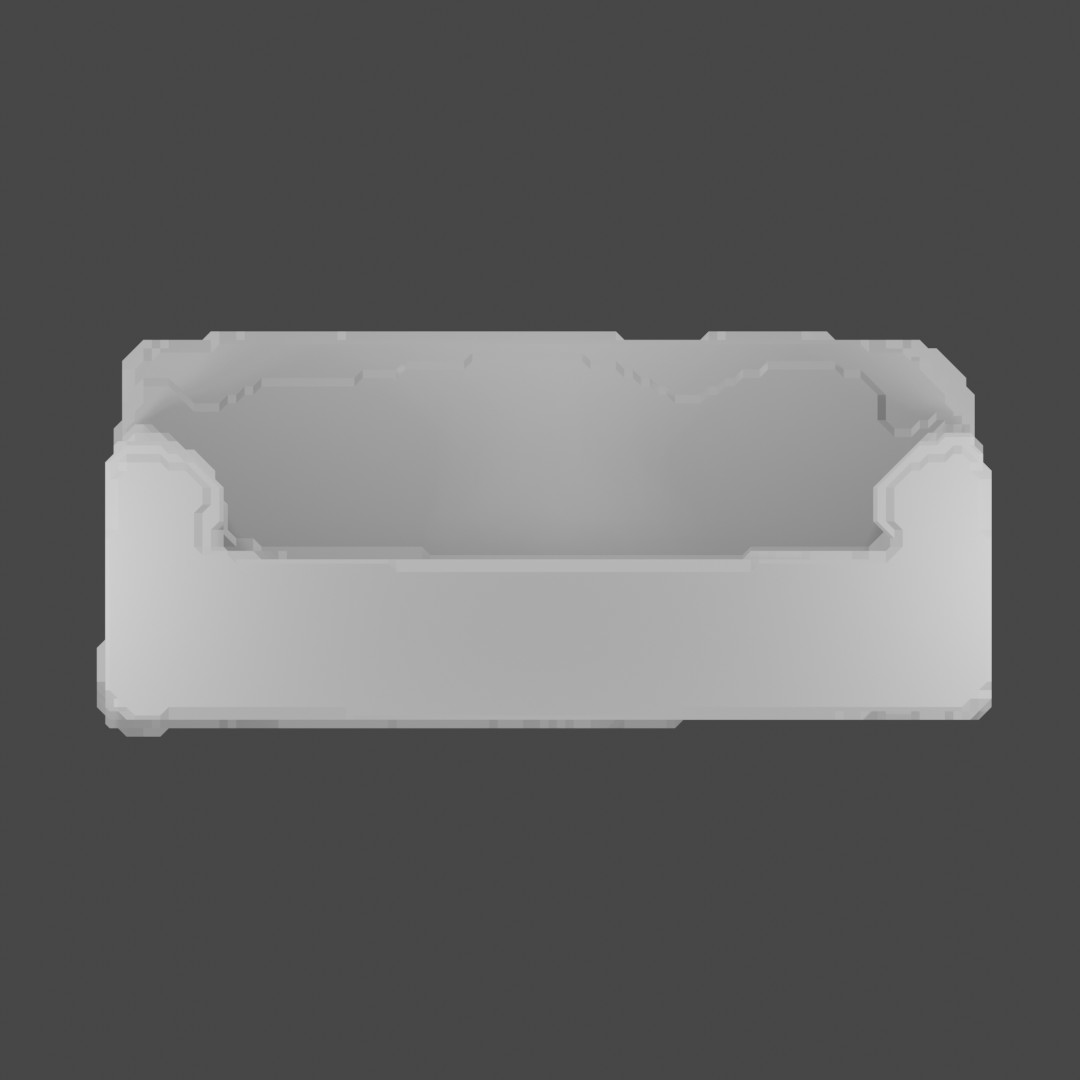}
	\includegraphics[trim=11mm 8mm 5mm 3mm,clip,width = 0.18\columnwidth]{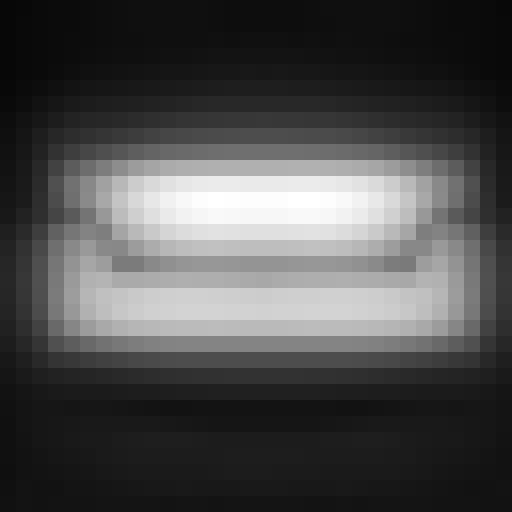}\\[0.1mm]
	\includegraphics[trim=11mm 8mm 5mm 3mm,clip,width = 0.18\columnwidth]{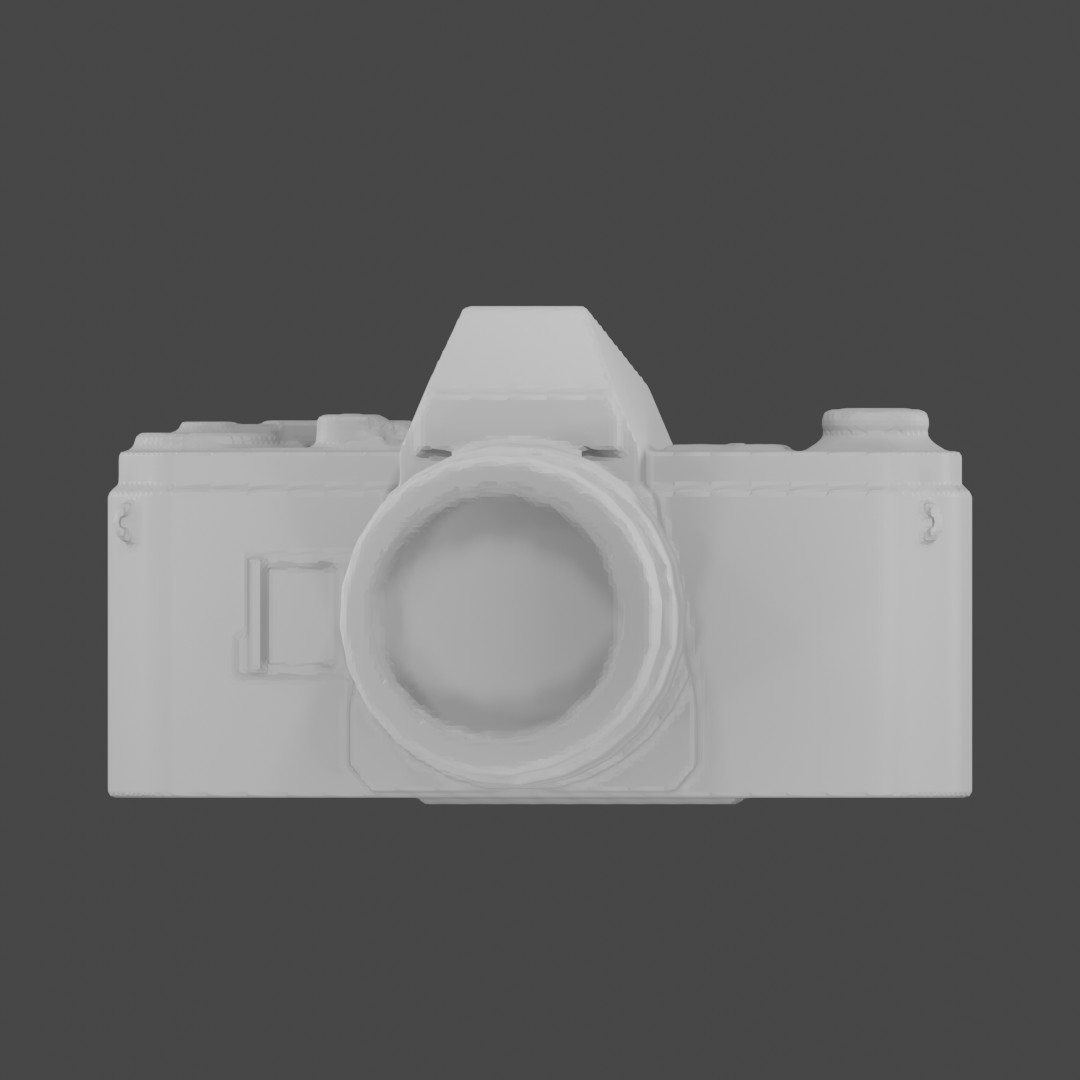}
	\includegraphics[trim=11mm 8mm 5mm 3mm,clip,width = 0.18\columnwidth]{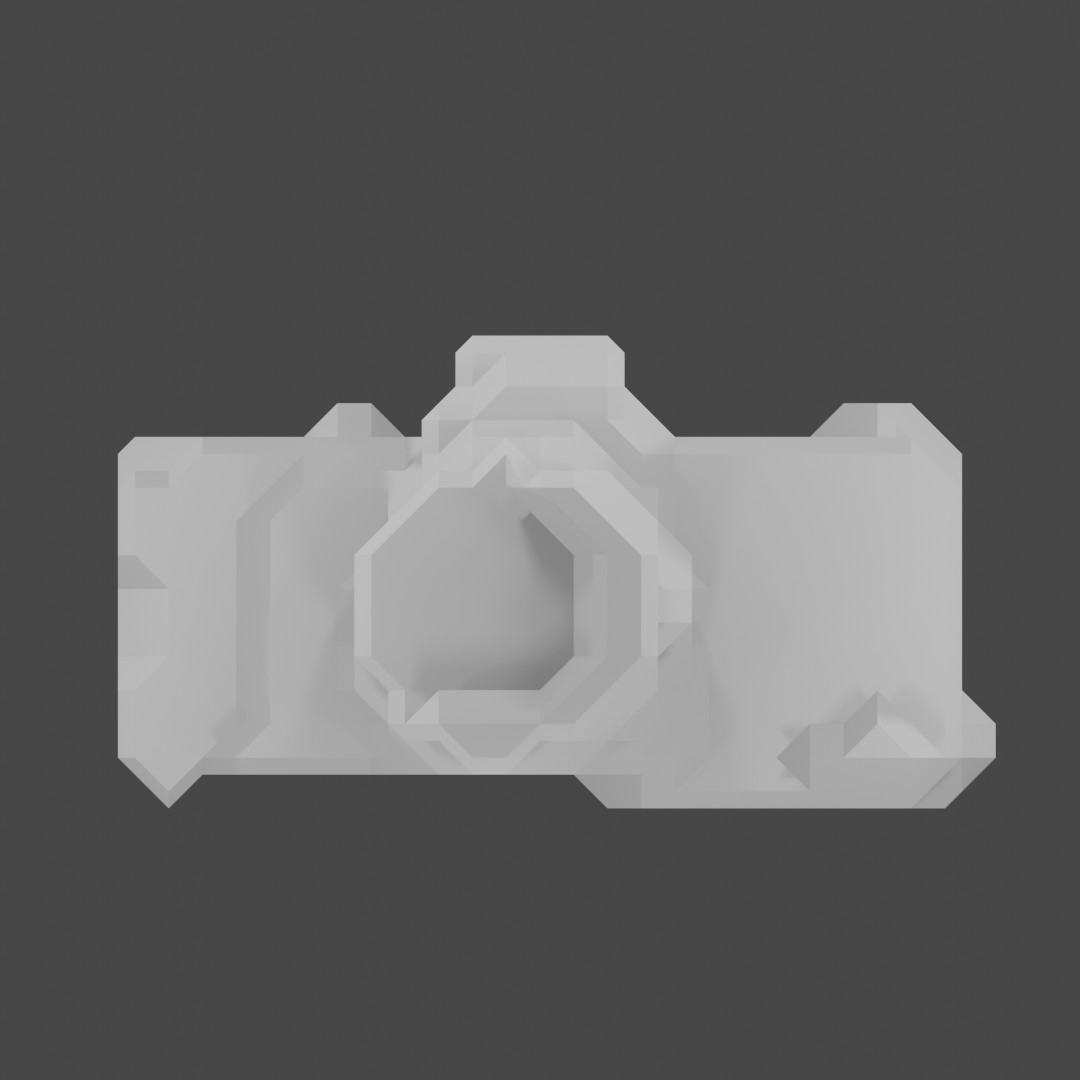}
	\includegraphics[trim=11mm 8mm 5mm 3mm,clip,width = 0.18\columnwidth]{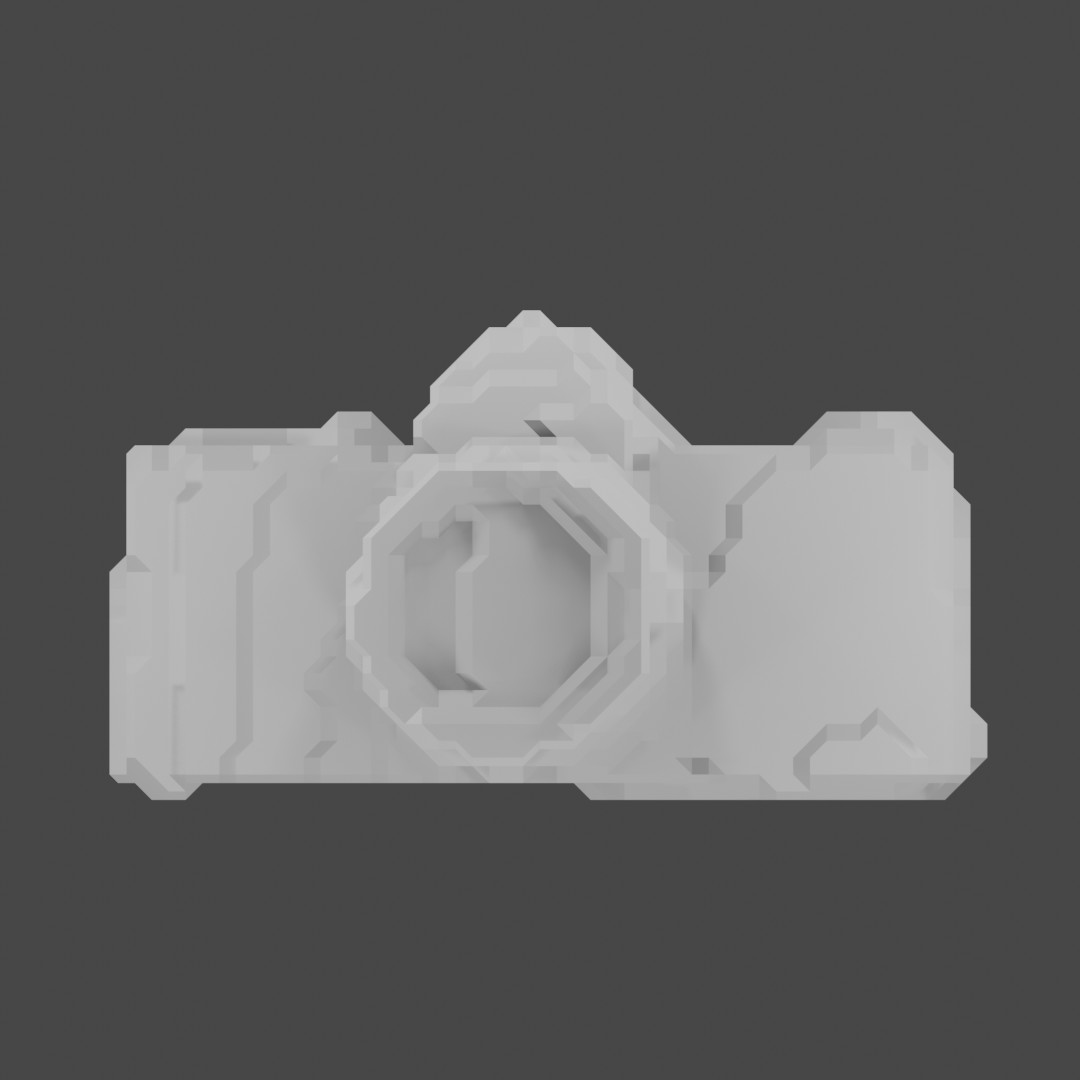}
	\includegraphics[trim=11mm 8mm 5mm 3mm,clip,width = 0.18\columnwidth]{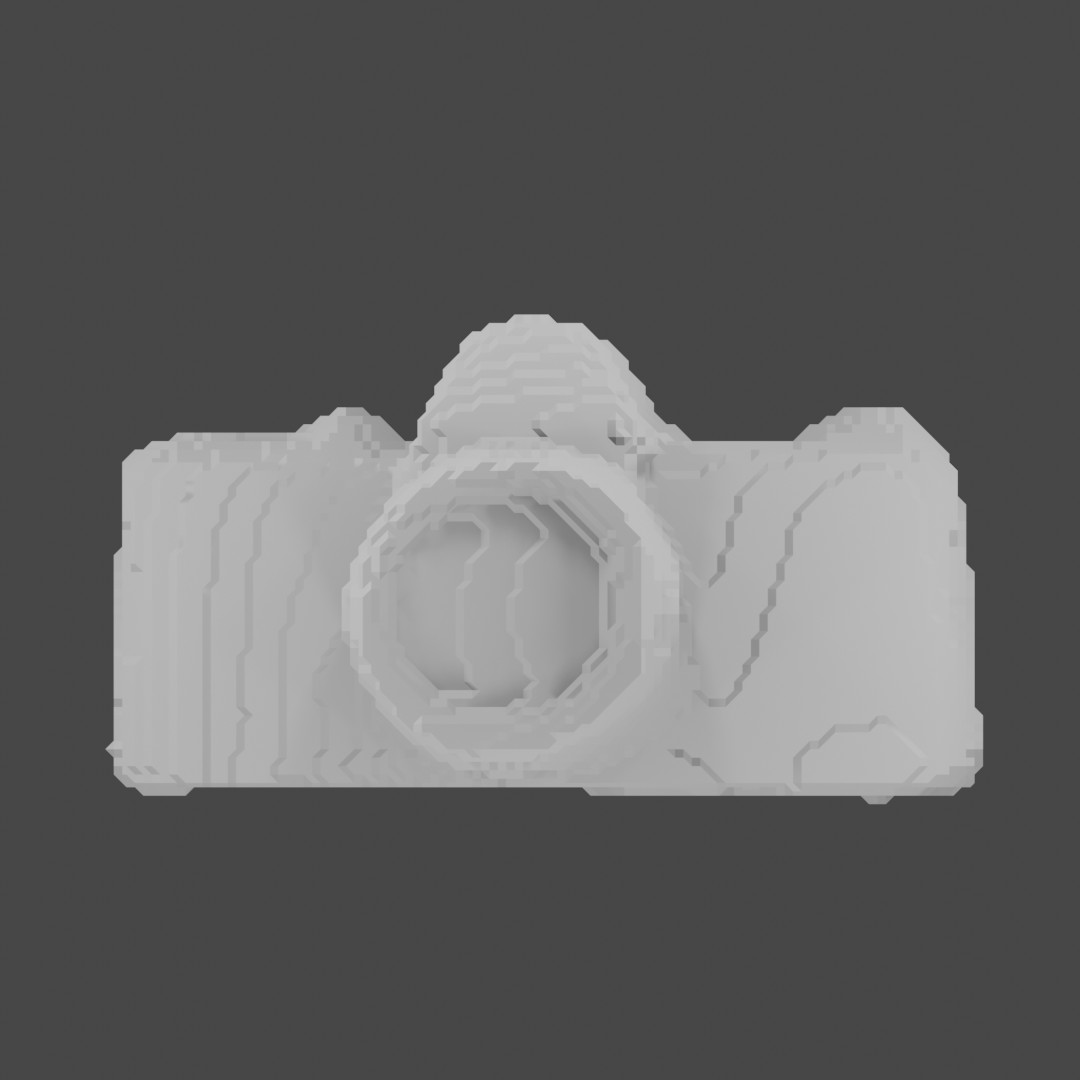}
	\includegraphics[trim=11mm 8mm 5mm 3mm,clip,width = 0.18\columnwidth]{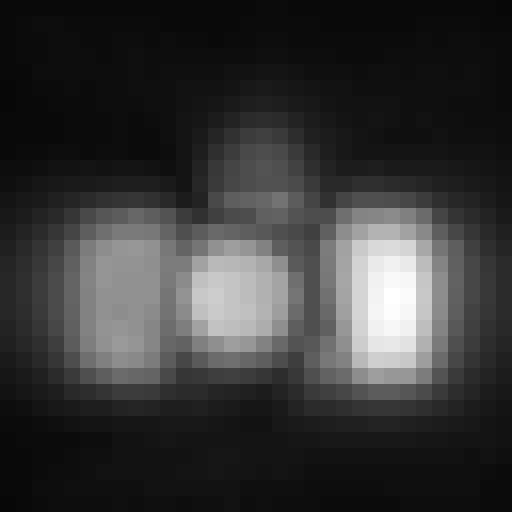}
	
	\caption{Surface adaptivity of our model after training. \textbf{First column:} ground truth scenes. \textbf{Second-fourth columns:} By sampling points close to the surface during training, it is possible to compute the occlusion function at arbitrary resolutions around the target object during inference. \textbf{Fifth column:} For illustration we show the front-projected $32\times 32$ reconstructions by LCT. The input measurements have dimensions $32\times 32\times 256$ sampled on a 70\,cm$\times$70\,cm wall area with 32\,ps of temporal resolution.}
	\label{fig:results_beyond_Fermat_1_suppmat}
\end{figure}

Figure \ref{fig:results_beyond_Fermat_1_suppmat} shows the adaptability of our trained model to compute different resolutions at inference. In order to achieve this, we sample 400,000 points which we partition as 70/30, corresponding to close to the object surface and uniformly distributed over space, respectively \cite{park2019deepsdf}. For the points close to the surface we used variances within the range $[0.001, 0.005]$, where the scene is defined inside the unit cube. Then, at prediction time we set different voxelizations of the marching cubes step in order to obtain arbitrary resolutions of the recovered target. For illustration, we also show results of the volumetric method LCT in its native output resolution. Note also how the latter suffers from the specular Fermat constraint (i.e surface elements pointing outside the wall are not easily recovered).

\subsection{Evaluation with Noisy Inputs}

Tables \ref{tab:clean_vs_noise} shows IoU and F-score for noisy vs clean predictions for the convolutional architecture \cite{peng2020convolutional}. In general, we observe that adding noise mildly degrades the evaluation scores, hence highlighting the robustness of our system to the considered noise model.

\begin{table}[H]
	\makebox[\textwidth][c]{
	\centering
	\begin{tabular}{cc|ccc|ccc|ccc}
	\toprule
	&       & \multicolumn{3}{c|}{5-Categories} & \multicolumn{3}{c|}{Self-Occlusion} & \multicolumn{3}{c}{Statues+Sculptures} \\
				\midrule
				Input Type	&	& ~F-Score 	& IoU 	& & ~F-Score 	& IoU 	& & ~~~ F-Score 	& ~IoU 	 	\\ \hline
				Clean  		&	& ~0.87  	& 0.77 	& & ~0.88		& 0.79 	& & ~~~ 0.77 		& ~0.63 		\\
				Noisy  		&	& ~0.85  	& 0.75 	& & ~0.87 		& 0.77 	& & ~~~ 0.74		& ~0.60 		\\
				\bottomrule
			\end{tabular}%
		}
		\vspace{1mm}
		\caption{Performance of the convolutional architecture for clean and noisy measurements. For all the datasets considered, the degradation in the evaluation scores after adding noise is mild.}
		\label{tab:clean_vs_noise}%
	\end{table}%

%
%

\end{document}